\documentclass[10pt,twocolumn,letterpaper]{article}

\usepackage{iccv}
\usepackage{times}
\usepackage{epsfig}
\usepackage{graphicx}
\usepackage{amsmath}
\usepackage{amssymb}
\usepackage{url}   
\usepackage{xcolor}
\usepackage{color}
\usepackage[skip=0.5ex]{caption}
\usepackage{microtype}
\usepackage{booktabs}       % professional-quality tables
\usepackage{multirow}
\usepackage{wrapfig}
\usepackage{stmaryrd} 
\usepackage[normalem]{ulem}
\usepackage{soul}
\usepackage{comment}
\usepackage{flushend}
\usepackage{dsfont}
\usepackage{bm}
\usepackage{caption}
\usepackage{subcaption}

\usepackage{graphbox}
\usepackage[export]{adjustbox}
\usepackage{widetext}

\usepackage{paralist}
\usepackage{array}
\usepackage{placeins}
\usepackage{epstopdf}
\usepackage[titletoc,title]{appendix}
\usepackage[T1]{fontenc}
\usepackage{diagbox}

\definecolor{myteal}{rgb}{0.25,0.5,0.5}
\newcommand{\relative}[1]{\textcolor{myteal}{#1}}

\usepackage{xr}
\makeatletter
\newcommand*{\addFileDependency}[1]{% argument=file name and extension
  \typeout{(#1)}
  \@addtofilelist{#1}
  \IfFileExists{#1}{}{\typeout{No file #1.}}
}
\makeatother

\newcommand*{\myexternaldocument}[1]{%
    \externaldocument{#1}%
    \addFileDependency{#1.tex}%
    \addFileDependency{#1.aux}%
}

\myexternaldocument{supplementary}
\usepackage[pagebackref=true,breaklinks=true,letterpaper=true,colorlinks,bookmarks=false]{hyperref}

\iccvfinalcopy % *** Uncomment this line for the final submission

\def\iccvPaperID{3490} % *** Enter the ICCV Paper ID here

\definecolor{orange}{rgb}{0.99,0.29,0.07}
\newcommand\emi{\textcolor{black}}

\newcommand\Gianni{\textcolor{black}}
  \newcommand\Isa{\textcolor{black}}

\newcommand{\ab}[1]{\textcolor{black}{#1}}
\newcommand{\abc}[1]{\textcolor{violet}{[AB: \em #1]}}

\newcommand{\real}{\mathbb{R}}

\newcommand{\vpsi}{\boldsymbol{\psi}}
\newcommand{\vmu}{\boldsymbol{\mathbf{\mu}}}
\newcommand{\vsigma}{\boldsymbol{\mathbf{\sigma}}}

\newcommand{\vW}{\mathbf{W}}

\newcommand{\vz}{\mathbf{z}}

\newcommand{\s}{\mathbf{s}}
\newcommand{\vr}{\mathbf{r}}

\newcommand{\vx}{\mathbf{x}}

\newcommand{\gdec}{g^{\scriptstyle \text{dec}}_{\scriptstyle \psi}}
\newcommand{\genc}{g^{\scriptstyle \text{enc}}_{\scriptstyle \phi}}
\newcommand{\ThetaLP}{\Theta^{\scriptscriptstyle \text{LP-BNN}}}

\newcommand{\losslpbnn}{\mathcal{L}_{\scriptscriptstyle \text{LP-BNN}}}
\newcommand{\lossbnn}{\mathcal{L}_{\scriptscriptstyle \text{BNN}}}

\newcommand{\parag}[1]{\smallskip\noindent\textbf{#1}~~}

\newcommand{\method}{LP-BNN\xspace}
\newcommand{\be}{BatchEnsemble\xspace}

\begin{document}

%%%%%%%%% TITLE
 \title{Encoding the latent posterior of Bayesian Neural Networks for uncertainty quantification}

\author{
	Gianni Franchi\textsuperscript{1} \ \ \ 
	Andrei Bursuc\textsuperscript{2} \ \ \ 
	Emanuel Aldea\textsuperscript{2} \ \ \ 
	S\'{e}verine Dubuisson\textsuperscript{4} \ \ \ 
	Isabelle Bloch\textsuperscript{1} \\
	\small \textsuperscript{1} Institut Polytechnique de Paris 
	\ \ \ \ \ \textsuperscript{2}valeo.ai
	\ \ \ \ \ \textsuperscript{3}Universit\'{e} Paris-Saclay
    \ \ \ \ \ \textsuperscript{4}Aix Marseille University
}

\maketitle
% Remove page # from the first page of camera-ready.
\ificcvfinal\thispagestyle{empty}\fi

%%%%%%%%% ABSTRACT
\begin{abstract}
Bayesian Neural Networks (BNNs) have been long considered an ideal, yet unscalable solution for improving the robustness and the predictive uncertainty of deep neural networks. While they 
could capture more accurately the posterior distribution of the network parameters, most BNN approaches are either limited to small networks or rely on constraining assumptions,
e.g., parameter independence. These drawbacks have enabled prominence of simple, but computationally heavy approaches such as Deep Ensembles, whose training and testing costs increase linearly with the number of networks. In this work we aim for efficient deep BNNs amenable to complex computer vision architectures, e.g., ResNet50 DeepLabV3+, and tasks, e.g., semantic segmentation, with fewer assumptions on the parameters. We achieve this by leveraging variational autoencoders (VAEs) to learn the interaction and the latent distribution of the parameters at each network layer. Our approach, Latent-Posterior BNN (LP-BNN), is compatible with the recent BatchEnsemble method, leading to highly efficient ({in terms of computation and} memory during both training and testing) ensembles. LP-BNNs attain competitive results across multiple metrics in several challenging benchmarks for image classification, semantic segmentation and out-of-distribution detection.
\end{abstract}

%%%%%%%%% BODY TEXT
\section{Introduction}

% \abc{Alternative names to VAE-BNN: Latent Posterior BNN (LP-BNN), }
%\abc{Alternative titles: Encoding the latent posterior of BNNs for uncertainty quantification } \Isa{c'est plus précis, donc mieux je pense ?}

\begin{figure}[t!]
  \renewcommand{\figurename}{Figure}
  \renewcommand{\captionfont}{\small}
  \centering
  \includegraphics[width=0.95\linewidth]{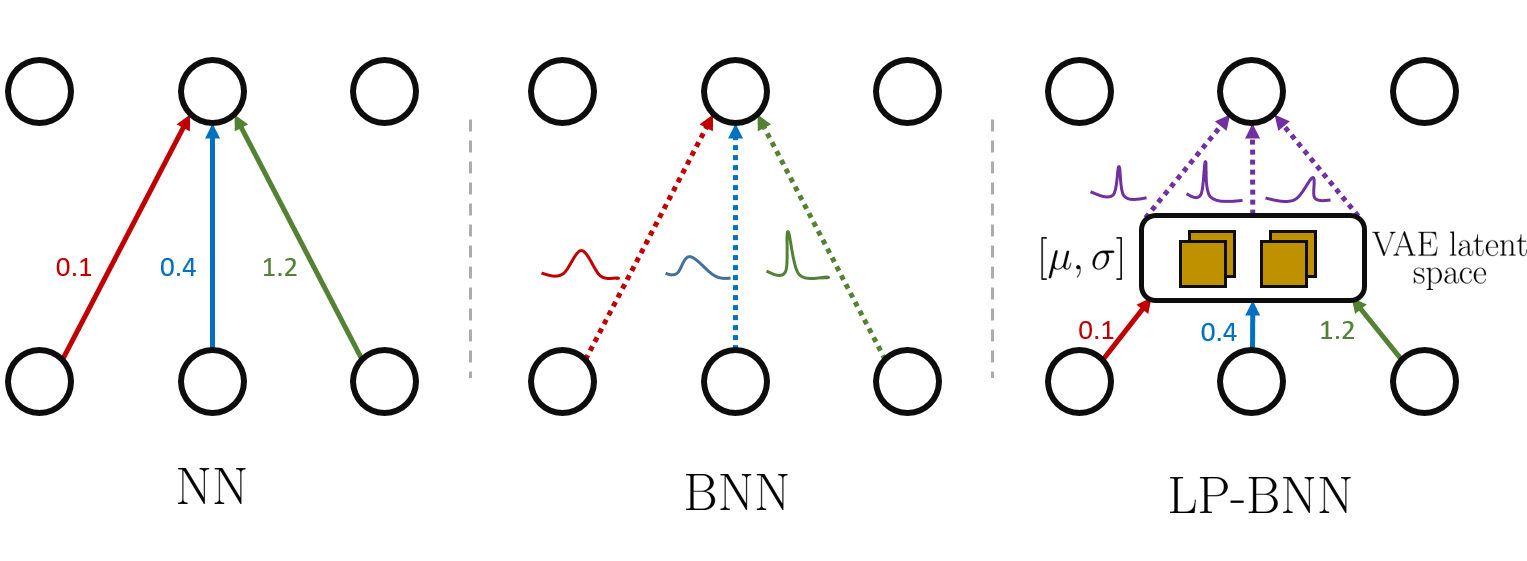}
  \caption{
%   (a) A standard DNN only trains weight point estimates for prediction; (b) A standard BNN \cite{blundell2015weight} considers that all weights are mutually independent and follow Gaussian distributions: every weight is factorized by a Gaussian distribution; (c) Our \method~ considers that {the weights at every layer} follow a multivariate Gaussian distribution, with a latent layer composed of mutually independent Gaussian distributions.  This assertion allows training a distribution on a smaller dimensional space. 
In a standard NN each weight has a fixed value. In 
% a 
\ab{most} BNN\ab{s}
% ~\cite{blundell2015weight} 
all weights follow Gaussian distributions and are assumed to be mutually independent: each weight is factorized by a Gaussian distribution. \ab{For \method  in each layer, weights follow a multivariate Gaussian distribution with a latent weight space composed of independent Gaussian distributions.} This enables computing expressive weight distributions in a lower dimensional space.
}
  \label{fig:teaser}
\vspace{-4mm}
\end{figure}

% The recent studies focusing on estimating the uncertainty of Deep Neural Networks (DNNs) are essentially based on ensemble techniques \cite{lakshminarayanan2017simple} which define the current state of the art. However their notable drawback in terms of computational cost make them often prohibitive. The study of the DNN uncertainty is difficult to formalize also due to multiple factors such as the shallow definition of in- and out-of-distribution samples, or to what one may consider as expected uncertainty. However, estimating uncertainty has become of crucial importance for computer vision tasks, in addition to the traditional goal of reaching a high accuracy.

% \ab{New intro}
\ab{
Most top-performing approaches for predictive uncertainty estimation with Deep Neural Networks (DNNs)~\cite{lakshminarayanan2017simple, ashukha2020pitfalls,maddox2019simple,franchi2019tradi} are essentially based on ensembles, in particular Deep Ensembles (DE)~\cite{lakshminarayanan2017simple}, which have been shown to display many strengths: stability, mode diversity, good calibration, \etc.~\cite{fort2019deep}. In addition, through the Bayesian lens, ensembles enable a more straightforward separation and quantification of the sources and forms of uncertainty~\cite{gal2016phd, lakshminarayanan2017simple, malinin2018predictive}, which in turn allows a better communication of the decisions to humans~\cite{bhatt2020uncertainty, kompa2021second} or to connected modules in an autonomous system~\cite{mcallister2017concrete}. This is crucial for real-world decision making systems. Although originally introduced as simple and scalable alternative to Bayesian Neural Networks (BNNs)~\cite{mackay1992practical, neal1995bayesian}, DE still have notable drawbacks in terms of computational cost for both training and testing often make them prohibitive in practical applications. 
}
% Studying DNN uncertainty is difficult to formalize
% The study of DNN uncertainty is particularly challenging 
% also due to the potentially vague definition of in- and out-of-distribution samples, or to the difficulty in identifying the type of uncertainty behind wrong predictions. \abc{The previous sentence could be removed at final trimming, not sure it helps.} 
% However, in the context of potential widespread adoption of DNNs for practical applications, uncertainty estimation is essential for computer vision tasks, on par with the traditional goal of reaching high accuracy.

\ab{In this work we address uncertainty estimation with BNNs, the departure point of DE. BNNs propose an intuitive and elegant formalism suited for this task by estimating the posterior distribution over the parameters of a network conditioned on training data. Performing exact inference BNNs is intractable and most approaches require approximations. The most common one is the mean-field assumption~\cite{jordan1999introduction}, i.e., the weights are assumed to be independent of each other and factorized by their own distribution, usually Gaussian~\cite{hinton1993keeping, graves2011practical, blundell2015weight, hernandez2015probabilistic, gal2016dropout, mishkin2018slang}.
However, this approximation can be damaging~\cite{mackay1992practical, foong2020expressiveness} as a more complex organization can emerge within network layers, and that higher level correlations contribute to better performance and generalization~\cite{bengio2017deep,salimans2016weight,srivastava2014dropout}. Yet, even under such settings, BNNs are challenging to train at scale on modern DNN architectures~\cite{ovadia2019can, dusenberry2020efficient}.
In response, researchers have looked into structured-covariance approximations~\cite{louizos2016structured, sun2017learning, zhang2018noisy, mishkin2018slang}, however they further increase memory and time complexity over the original mean-field approximation.
% In response, 
% \ab{several more efficient approaches for estimating the posterior distribution have emerged~\cite{franchi2019tradi,maddox2019simple},} yet 
% \ab{they also build upon} similar assumptions about the training process and the complete independence of the weights of the DNN. 
}

\ab{Here, we revisit BNNs in a pragmatic manner. We propose an approach to estimate the posterior of a BNN with layer-level inter-weight correlations, in a stable and computationally efficient manner, compatible with modern DNNs and complex computer vision tasks, e.g., semantic segmentation. 
We advance a novel deep BNN model, dubbed \emph{Latent Posterior BNN} (LP-BNN), where the posterior distribution of the weights at each
layer is encoded with a variational autoencoder (VAE)~\cite{kingma13vae} into a lower-dimensional latent space that follows a Gaussian distribution (see Figure~\ref{fig:teaser}). We switch from the inference of the posterior in the high dimensional space of the network weights to a lower dimensional space which is easier to learn and already encapsulates weight interaction information. LP-BNN is naturally compatible with the recent BatchEnsemble (BE) approach~\cite{wen2020batchensemble} that enables learning a more diverse posterior from the weights of the BE sub-networks. Their combination outperforms most of related approaches across a breadth of benchmarks and metrics. In particular, \method is competitive with DE and has significantly lower costs for training and prediction.
% This enables to efficiently learn weight correlations by projecting them in the latent space.
% Our training procedure may be viewed as a standard optimization problem involving the training of multiple VAEs, as opposed to specific BNN training strategies requiring individual weight samplings whose variance hinders training stability~\cite{dusenberry2020efficient}. 
% While VAEs are typically used to learn distributions in order to generate images, here our aim is to learn the distribution that generates DNNs to achieve a higher accuracy. We outline LP-BNN in Figure~\ref{fig:teaser}.
% \\
% So far BNNs have failed to reach competitive performance in popular computer vision benchmarks, often lagging behind Deep Ensemble\ab{s}~\cite{lakshminarayanan2017simple}. 
}

\parag{Contributions.} \ab{To summarize, the contributions of our
work are: \textbf{(1)} We introduce a scalable approach for BNNs to implicitly capture \emph{layer-level weight correlations} enabling more expressive posterior approximations, by foregoing the limiting mean-field assumption 
% of weight independence. 
LP-BNN scales to high capacity DNNs (e.g., 50+ layers and 30M parameters for DeepLabv3+), while still training on a single V100 GPU. \textbf{(2)} We propose to leverage VAEs for computing the posterior distribution of the weights by projecting them in the latent space. This improves significantly training stability while ensuring diversity of the sampled weights. \textbf{(3)} We extensively evaluate our method on a range of computer vision tasks and settings: image classification for \emph{in-domain uncertainty},
% (CIFAR-10/100)
 \emph{out-of-distribution (OOD) detection}, 
% (CIFAR-10 vs. SVHN), 
\emph{robustness to distribution shift}, 
% (CIFAR-C~\cite{hendrycks2018benchmarking}), 
and semantic segmentation ( high-resolution images, strong class imbalance) for \emph{OOD detection}.
% (Street Hazards, BDD Anomaly~\cite{hendrycks2019anomalyseg}). 
We demonstrate that LP-BNN achieves similar performances with high-performing Deep Ensembles, while being substantially more efficient computationally.  }
\section{Background}\label{section:background}
In this section, we present the chosen formalism for this work and \ab{offer} a short background on BNNs.

\subsection{Preliminaries}
We consider a training dataset $\mathcal{D} = \{ (\vx_i, y_i) \}_{i=1}^{n}$ with $n$ samples and labels, corresponding to two random variables $X \sim \mathcal{P}_X$ and $Y\sim \mathcal{P}_Y$. 
% For the sake of generality \Isa{ou without loss of generality?} 
Without loss of generality we represent $\vx_i \in \real^{d} $ as a vector, and $y_i$
as a scalar label. We process the input data $\vx_i$ with a neural network $f_{\Theta}(\cdot)$ with parameters $\Theta$, that outputs a classification or regression prediction. We view the neural network as a probabilistic model with $f_{\Theta}(\vx_i)=P(Y=y_i \mid X=\vx_i,\Theta)$. %\Sev{$\vy$ n'a pas ete introduit, c'est $y_i$ non ? et c'est $\vx_i$ ? (a verifier pour la suite)}. 
In the following, when there are no ambiguities, we discard the random variable from notations. For classification, $P(y_i \mid \vx_i,\Theta)$ %\Sev{$P(Y|\vx,\Theta)$ non ? GIANNI/ comme on supprime les RA on met tout en minuscule}
is a categorical distribution over the set of classes over {the domain of} $Y$, %\Sev{$y_i$}, 
typically corresponding to the cross-entropy loss function, while for regression  $P(y_i \mid \vx_i,\Theta)$ is a Gaussian distribution of real values  over {the domain of} $Y$ %\Sev{$y_i$}
when using the squared loss function.
%\Isa{Pourquoi mentionne-t-on la loss ici ?}. 
For simplicity 
% and consistency 
we unroll our reasoning for the classification task.
% \ab{classification.}

In supervised learning, we leverage gradient descent for learning $\Theta$ that minimizes the cross-entropy loss, which is equivalent to finding the parameters that maximize the likelihood estimation (MLE) $P(\mathcal{D} \mid \Theta)$ over the training set $\Theta_{\text{MLE}} = \arg \max_{\Theta} \sum_{(\vx_i,y_i) \in \mathcal{D}} \log P(y_i \mid \vx_i,\Theta)$, or equivalently minimize the following loss function: 
% \abc{Gianni initially started writing these equations from the weight point of view. However I would argue that expressing them as losses will bridge easier with the subsequent sections. Depending on the choice, I will adjust the wording.}: \Isa{I agree!}
% \begin{equation}\label{eq:MLE}
% \Theta_{\text{MLE}} = \arg \max_{\Theta} \sum_{i=1}^n \log P(y_i|\vx_i,\Theta).
% \end{equation}
\begin{equation}\label{eq:loss_MLE}
\mathcal{L}_{\scriptscriptstyle \text{MLE}}(\Theta) = - \sum_{(\vx_i,y_i) \in \mathcal{D}} \log P(y_i\mid \vx_i,\Theta).
\end{equation}
%\Isa{this last equation is redundant, so maybe can be suppressed... and this would avoid too many "which" in the previous sentence!}
% We should also decide whether sums are written over $i$ or over $\vx_i, y_i$ (equivalent, but this is just to be consistent...)}

%\Sev{somme sur $(\vx_i,y_i) \in \mathcal{D}$ ?}\\
The Bayesian approach enables adding prior information on the parameters $\Theta$, by placing a prior distribution $\mathcal{P}(\Theta)$ upon them. This prior represents some expert knowledge about the dataset and the model. Instead of maximizing the likelihood, we can now find the maximum a posteriori (MAP) weights for $\mathcal{P}(\Theta \mid \mathcal{D}) \propto \mathcal{P}( \mathcal{D} \mid \Theta) \mathcal{P}(\Theta)$
%decomposed into $\mathcal{P}( \mathcal{D} \mid \Theta)$ and $\mathcal{P}(\Theta)$ 
to compute $\Theta_{\text{MAP}}  = \arg \max_{\Theta} \sum_{(\vx_i,y_i) \in \mathcal{D}} \log \mathcal{P}(y_i \mid \vx_i,\Theta) + \log \mathcal{P}(\Theta)$, \ie to minimize the following loss function:
% \begin{equation}\label{eq:MAP}
% \Theta_{\text{MAP}}  = \arg \max_{\Theta} \sum_{i=1}^n \log \mathcal{P}(y_i|\vx_i,\Theta) + \log \mathcal{P}(\Theta),
% \end{equation}
\begin{equation}\label{eq:loss_MAP}
\mathcal{L}_{\scriptscriptstyle \text{MAP}}(\Theta)
  = - \sum_{(\vx_i,y_i) \in \mathcal{D}} \log P(y_i \mid \vx_i,\Theta) - \log P(\Theta),
\end{equation}
%\Sev{somme sur $(\vx_i,y_i) \in \mathcal{D}$ ?}\\
inducing a specific distribution over the functions computed by the network and a regularization of the weights. For a Gaussian prior, Eq.~\eqref{eq:loss_MAP} reads as $L_2$ regularization (weight decay).

\subsection{Bayesian Neural Networks}
 %\Isa{pourquoi in spite of ici ?}\abc{The MAP is formalised as getting the posterior distribution by maximizing the likelihood while taking the prior distribution into consideration. However we usually keep just the point estimates for the maximum.} 
In 
%spite of the Bayesian reasoning, for 
most neural networks only the $\Theta_\text{MAP}$ weights computed during training are kept for predictions. Conversely, in BNNs we aim to find the posterior distribution $P(\Theta \mid \mathcal{D})$ of the parameters given the training dataset, not only the values corresponding to the MAP. Here we can make {a prediction $y$} %\Isa{a decision ?} \Gianni{En machine learning on parle de prediction}
% \Gianni{a prediction $y$} \abc{it would be rather $\hat{y}$ or $P(y|\vx)$, but I wanted to keep it short since it's in the equation below} 
on a new sample $\vx$ by computing the expectation of the predictions from an infinite ensemble corresponding to different configurations of the weights sampled from the posterior distribution:
\begin{equation}\label{eq:marginalization}
P(y \mid \vx,\mathcal{D})  = \int P(y \mid \vx, \Theta) P(\Theta \mid \mathcal{D}) d\Theta ,
\end{equation}
%\Isa{integrale sur le domaine de $\Theta$ ?} \Gianni{oui pour confirmation c'est ecris dans la section 2.1 de https://www.cs.ox.ac.uk/people/yarin.gal/website/thesis/thesis.pdf et dans la plus part des articles du domaines}
which is also known as Bayes ensemble. The integral in Eq.~\eqref{eq:marginalization}, {which is calculated over the domain of $\Theta$}, is intractable,  and in practice it is approximated by averaging predictions from a limited set $\{\Theta_1, \ldots \Theta_J\}$ of $J$ weight configurations sampled from the posterior distribution:
\begin{eqnarray}\label{eq:approximate_likelihood}
P(y \mid \vx, \mathcal{D}) \approx \frac{1}{J} \sum_{j=1}^J  P( y \mid \vx, \Theta_j).
\end{eqnarray}

Although BNNs are elegant and easy to formulate, their inference is non-trivial and has been subject to extensive research across the years~\cite{hinton1993keeping,mackay1992practical,neal1995bayesian}. Early approaches relied on Markov chain Monte Carlo variants for inference, while progress in variational inference (VI)~\cite{jordan1999introduction} has enabled a recent revival of BNNs~\cite{graves2011practical, blundell2015weight, hernandez2015probabilistic}. VI turns posterior inference into an optimization problem. In detail, VI finds the parameters $\nu$ of a distribution $Q_{\nu}(\Theta)$ on the weights that approximates the true Bayesian posterior distribution of the weights $P(\Theta \mid \mathcal{D})$ through KL-divergence minimization. This is equivalent to minimizing the following loss function, also known as expected lower bound (ELBO) loss~\cite{blundell2015weight, kingma13vae}:
\begin{align}\label{eq:loss-BNN}
\mathcal{L}_{\scriptscriptstyle \text{BNN}}(\Theta, \nu)  = & -\sum_{(\vx_i,y_i) \in \mathcal{D}} \mathbb{E}_{\scriptstyle \ab{\Theta \sim}Q_{\nu}(\Theta)} \log  \left(P( y_i \mid \vx_i, \Theta) \right) \nonumber \\ 
 & + \mbox{KL}(Q_{\nu}(\Theta)\vert\vert P(\Theta)). 
\end{align}
%\Sev{Parle-t-on ici encore de somme sur $(\vx_i,y_i) \in \mathcal{D}$ ? Il faut probablement changer aussi dans $P(y|\vx,\Theta)$}\\
The loss function $\mathcal{L}_{\scriptscriptstyle \text{BNN}}(\Theta, \nu)$ is composed of two terms: the KL term depends on the weights and the prior $P(\Theta)$, while the likelihood term is data dependent. This function strives to simultaneously capture faithfully the complexity and diversity of the information from data $\mathcal{D}$, while 
% satisfying the 
preserving the simplicity of the prior $P(\Theta)$. To optimize this loss function, Blundell \etal~\cite{blundell2015weight} proposed leveraging the \emph{re-parameterization trick}~\cite{kingma13vae,rezende2014stochastic}, foregoing the expensive MC estimates.

\paragraph{Discussion.} BNNs are particularly appealing for uncertainty quantification thanks to the ensemble of predictions from multiple weight configurations sampled from the posterior distribution. However this brings an increased computational and memory cost. For instance, the simplest variant of BNNs with fully factorized Gaussian approximation distributions~\cite{blundell2015weight,graves2011practical}, 
% \ie approximating distribution factorized for each weight scalar, 
%  \ie each weight scalar, \Sev{sampled from} its own distribution, carry \Sev{carries?} \Isa{oui, c'est le poids} \abc{I was referring to BNNs, that they are overall twice in size than normal nets. } \Isa{but then the subject is the simplest variant..} 
{\ie each weight consists of a Gaussian mean and variance, carries a double amount of parameters.} In addition, 
% Dusenberry \etal~\cite{dusenberry2020efficient}
\ab{recent works~\cite{ ovadia2019can, dusenberry2020efficient}} point out that BNNs often underfit, and need multiple tunings to stabilize training dynamics involved by the loss function and the variance from weight samplings at each forward pass. 
Due to computational limitations, most BNN approaches assume that parameters are not correlated. This hinders their effectiveness~\cite{foong2020expressiveness}, as empirical evidence has shown that encouraging weight collaboration improves training stability and generalization~\cite{qiao2019weight, salimans2016weight,srivastava2014dropout}. 

In order to calculate a tractable weight correlation aware posterior distribution, we propose to leverage a VAE to compute compressed latent distributions {from which we can sample new weight configurations.}
% to sample from 
% \Isa{them?}. 
We rely on the recent BatchEnsemble (BE) method~\cite{wen2020batchensemble} to further improve the parameter-efficiency of BNNs. We now proceed to describe BE and then derive our approach.

\begin{figure}[!t]
\renewcommand{\captionfont}{\small}
\centering
\includegraphics[width=0.70\linewidth]{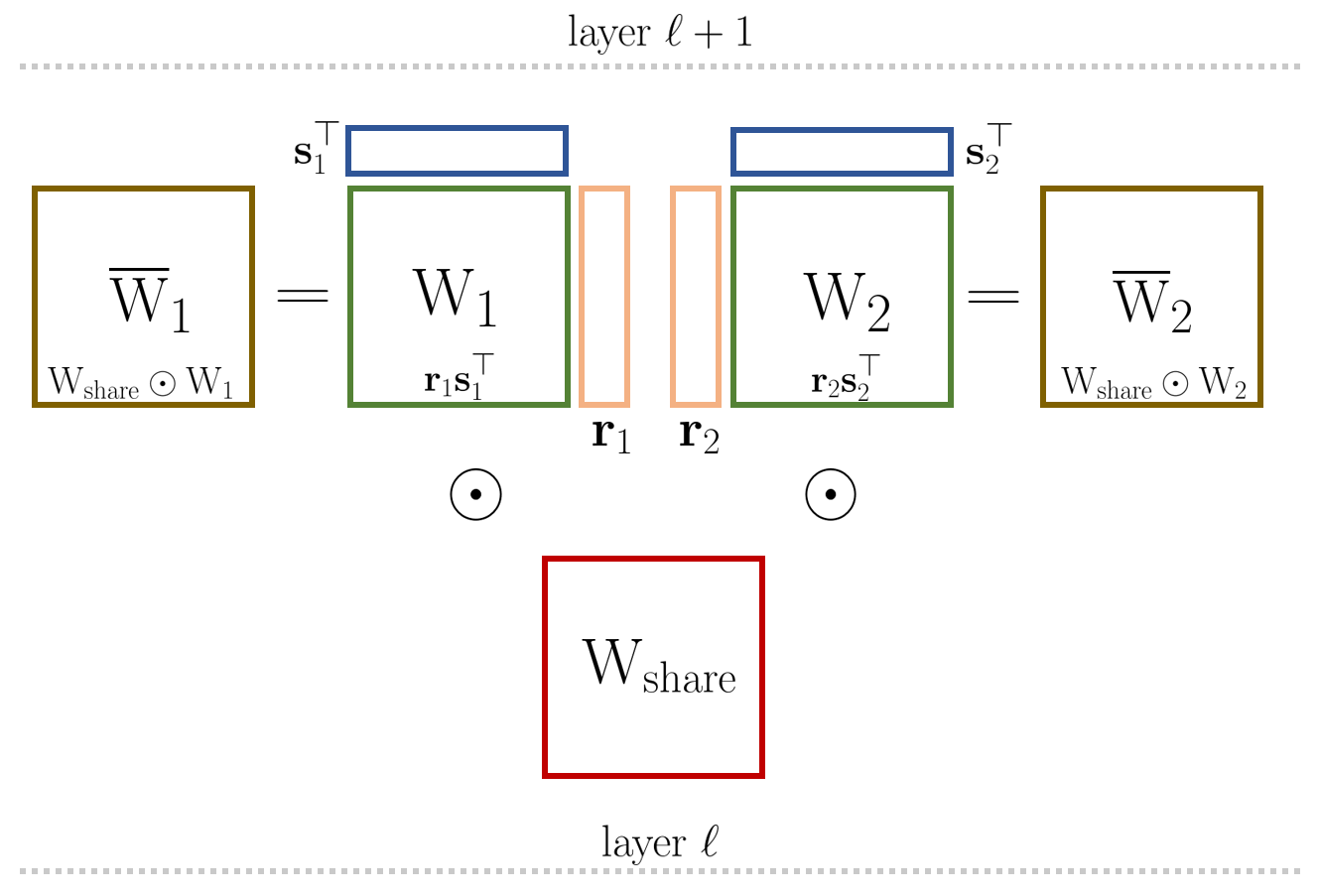}
        %  \caption{Illustration on how BatchEnsemble generates the ensemble weights for an ensemble of size $J=2$.} 
        \caption{\ab{\textbf{Diagram of a BatchEnsemble layer} that generates for an ensemble of size $J{=}2$, the ensemble weights $\overline{\vW}_j$ from shared weights $\vW_{\text{share}}$ and fast weights $\vW_j {=} \vr_j \s_j^{\top}$, with $j \in \llbracket 1, J \rrbracket $}.}
\label{fig:be-layer}
\end{figure}

\subsection{BatchEnsemble}

Deep Ensembles (DEs)~\cite{lakshminarayanan2017simple} {are a popular and pragmatic alternative to BNNs. While DEs} boast outstanding accuracy and predictive uncertainty, 
% yet \Isa{boast est-il le bon mot?} 
their training and testing cost increases linearly with the number of networks. This drawback has motivated the emergence of a recent stream of works proposing efficient ensemble methods~\cite{ashukha2020pitfalls, franchi2019tradi, maddox2019simple, mehrtash2020pep, wen2020batchensemble}. One of the most promising ones is \be~\cite{wen2020batchensemble}, which mimics in a parameter-efficient manner one of the main strengths of DE, \ie diverse  predictions~\cite{fort2019deep}. 
% We introduce BE briefly in the following. 

In a nutshell, BE builds up an ensemble from a single base network (shared among ensemble members) and a set of layer-wise weight matrices specific to each member. At each layer, the weight of each ensemble member is generated from the Hadamard product between a weight shared among all ensemble members, called \emph{``slow weight''}, and a Rank-1 matrix that varies among all members, called \emph{``fast weight''}. Formally, let $\vW_{\text{share}} \in \real^{m \times p}$ be the slow weights in a neural network layer with input dimension $m$ and with $p$ outputs. Each member $j$ from an ensemble of size $J$ owns a fast weight matrix $\vW_j \in \real^{m \times p}$. $\vW_j$ is a Rank-1 matrix computed from a tuple of trainable vectors $\vr_j \in \real^m$ and $\s_j \in \real^p$, with $\vW_j = \vr_j \s_j^{\top}$. BE generates from them a family of ensemble weights as follows: $\overline{\vW}_j = \vW_{\text{share}} \odot \vW_j$, where $\odot$ is the Hadamard product. Each $\overline{\vW}_j$ member of the ensemble is essentially a Rank-1 perturbation of the shared weights $\vW_{\text{share}}$ (see Figure~\ref{fig:be-layer}).
The sequence of operations during the forward pass reads:
\begin{equation}\label{eq:be_forward}
    h =  a \left( (W_{\text{share}}^{\top} (\vx \odot \s_j)) \odot \vr_j \right),
\end{equation}
where $a$ is an activation function and $h$ the output activations. 

The operations in BE can be efficiently vectorized, enabling each member to process in parallel the corresponding subset of samples from the mini-batch. $\vW_{\text{share}}$ is trained in a standard manner over all samples in the mini-batch. A BE network $f_{\Theta^{\scriptscriptstyle \text{BE}}}$ is parameterized by an extended set of parameters 
$\Theta^{\scriptscriptstyle \text{BE}} = \big\{ \theta^{\text{slow}}:\{\vW_{\text{share}} \}, \theta^{\text{fast}}:\{\vr_j, \s_j\}_{j=1}^J \big \}$.

With its multiple sub-networks parameterized by a reduced set of weights, BE is a practical method that can potentially improve the scalability of BNNs. We take advantage of the small size of the fast weights to capture efficiently the interactions between units and to compute a latent distribution of the weights. We detail our approach below.

\section{Efficient Bayesian Neural Networks (BNNs) }\label{section:approach}

% \ab{Attack plan:
% \begin{itemize}
%     \item the challenge of modeling the dependence between weights
%     \item quick reminder on latent variable models
%     \item description of LP-BNN
%     \item some words about backpropagation, \ie reparameterization trick
%     \item discussion on covariance w.r.t. Dusenberry \etal. Drawbacks of Rank-1 BNNs: computational inefficiency, careful initialization required.
%     \item (optional) difference between a Gaussian prior on all weights, i.e. L2 regularization, and the Gaussian prior the latent space of the weights.
%     \item utility of Rank-1 priors?
%     \item computational complexity
% \end{itemize}
% }

\subsection{Encoding the posterior weight distribution of a BNN}

Most BNN variants assume full independence between weights, both inter- and intra-layer. Modeling precisely weight correlations in modern high capacity DNNs with thousands to millions of parameters per layer~\cite{he2016deep} is however a daunting endeavor due to computational intractability. Yet, multiple strategies aiming to boost weight collaboration in one way or another, \eg Dropout~\cite{srivastava2014dropout}, WeightNorm~\cite{salimans2016weight}, Weight Standardization~\cite{qiao2019weight}, have proven to improve training speed, stability and generalization. Ignoring weight correlations might partially explain the shortcomings of BNNs in terms of underfitting~\cite{ovadia2019can, dusenberry2020efficient}. This motivates us to find a scalable way to compute the posterior distribution of the weights without discarding their correlations.
% and better exploit BNNs.

Li \etal~\cite{li2018intrinsic} have recently found that the \emph{intrinsic} dimension of DNNs can be in the order of hundreds to a few thousands. The good performances of BE, that builds on weights from a low-rank subspace, further confirm this finding. For efficiency, we leverage the Rank-1 subspace decomposition in BE and estimate here the distribution of the weights, leading to a novel form of BNNs. Formally, instead of computing the posterior distribution $P(\Theta \mid \mathcal{D})$, we aim now for $P(\theta^{\text{fast}} \mid \mathcal{D})$.

A first approach would be to compute Rank-1 weight distributions by using $\vr_j$ and $\s_j$ as variational layers, place priors on them and compute their posterior distributions in a similar manner to~\cite{blundell2015weight}. Dusenberry \etal~\cite{dusenberry2020efficient} show that these Rank-1 BNNs stabilize training by reducing the variance of the sampled weights, due to sampling only from Rank-1 variational distributions instead of full weight matrices. However this raises the memory cost significantly, as 
% inference 
\ab{training} is performed simultaneously over all $J$ sub-networks: on CIFAR-10 for ResNet-50 with $J{=}4$, the authors use $8$ TPUv2 cores with mini-batches of size $64$ per core.

%\todo{Andrei: The analysis on covariances could be made here. To avoid distracting the reader from the topic with new notations I would adapt the notations as follows: $\vW=\vr\s^{\top}$, $W(a,b)=\vW_{(u,v)}$ } \Gianni{Not essential and we have no space}

We argue that a more efficient way of computing the posterior distribution of the fast weights would be to 
% infer
\ab{learn} instead the posterior distribution of the lower dimensional latent variables of $\{\vr, \s\} \in \theta^{\text{fast}}$. This can be efficiently done with a VAE~\cite{kingma13vae} that can find a variational approximation $Q_{\phi}(\vz \mid \vr)$ to the intractable posterior distribution $P_{\psi}(\vz \mid \vr)$.  VAEs can be seen as a generative model that can deal with complicated dependencies between input dimensions via a probabilistic encoder that projects the input into a latent space following a specific prior distribution. For simplicity and clarity, from here onward we derive our formalism only for $\vr$ at a single layer and consider weights $\s$ to be deterministic. Here the input to the VAE are the weights $\vr$ and we rely on it  %\Isa{it se refere a quoi ?} \Gianni{($\vr$)}
to learn the dependencies between weights and encode them into the latent representation.

% \ab{
% \begin{equation}
% P(\theta^{\text{fast}} \vert \mathcal{D}) \approx P(\theta^{\text{fast}} \vert \mathcal{D}
% \end{equation}
% }

\begin{figure}[!t]
\renewcommand{\captionfont}{\small}
\centering
 \includegraphics[width=\linewidth]{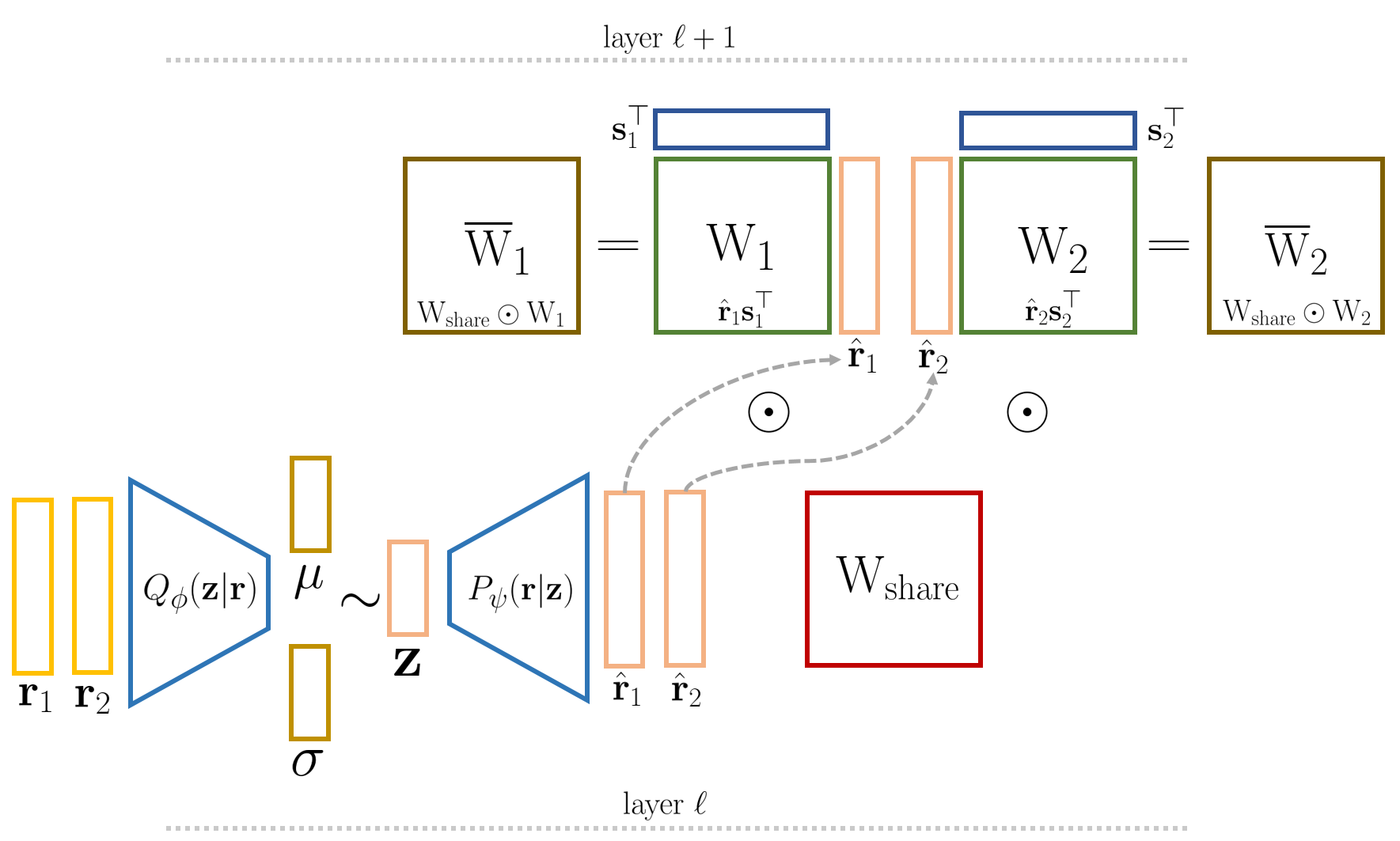}
        %  \caption{Illustration on how \method generates the ensemble weights ($J=2$) using sampled fast weights $\hat{\vr}_{\scriptscriptstyle 1}$ and $\hat{\vr}_{\scriptscriptstyle 2}$.}
         \caption{\ab{\textbf{Diagram of a \method layer} that generates for an ensemble of size $J{=}2$, ensemble weights $\overline{\vW}_j$ from shared weights $\vW_{\text{share}}$ and fast weights $\s_{\scriptscriptstyle j}$ and $\hat{\vr}_{\scriptscriptstyle j}$, the latter sampled and decoded from the corresponding latent projection $\vz_{\scriptscriptstyle j}$ of $\vr_{\scriptscriptstyle j}$, with $j \in \llbracket 1, J \rrbracket $.}} 
         
        %  that generates for an ensemble of size $J{=}2$, the ensemble weights $\overline{\vW}_j$ from shared weights $\vW_{\text{share}}$ and from fast weights $\vW_j {=} \vr_j \s_j^{\top}$, with $j \in \llbracket 1, J \rrbracket $}.
\label{fig:LPBNN}
\end{figure}

In detail, for each layer of the network $f_{\Theta}(\cdot)$ we introduce a VAE composed of a one layer encoder 
% $g^{\mbox{enc}}_{\phi}(\cdot)$ 
$\genc(\cdot)$ with variational parameters $\phi$ and a one layer decoder 
% $g^{\mbox{dec}}_{\psi}(\cdot)$
$\gdec(\cdot)$ with parameters $\psi$. Let the prior over the latent variables be a centered isotropic Gaussian distribution $P_{\psi}(\vz)= \mathcal{N}(\vz;0, \mathbf{I})$. Like common practice, we let the variational approximate posterior distribution $Q_{\phi}(\vz \mid \vr)$ be a multivariate Gaussian with diagonal covariance. The encoder takes as input a mini-batch of size $J$ (the size of the ensemble) composed of all the $\vr_j$ weights of this layer and outputs as activations $(\vmu_j, \vsigma^2_j)$. We sample a latent variable $\vz_j \sim \mathcal{N}(\vmu_j, \vsigma^2_j \mathbf{I})$ and feed it to the decoder, which in turn outputs the reconstructed weights 
$\hat{\vr}_j= \gdec(\vz_j)$. In other words, at each forward pass, we sample new fast weights $\hat{\vr}_j$ from the latent posterior distribution to be further used for generating the ensemble. The weights of each member of the ensemble $\overline{\vW}_{j} = \vW_{\scriptstyle \text{share}} \odot (\hat{\vr}_j \s_j^{\top})$ are now random variables depending on $\vW_{\scriptstyle \text{share}}$, $\s_j$ and $\vz_j$.
Note that while in practice we sample $J$ weight configurations, this approach allows us to generate larger ensembles by sampling multiple times from the same latent {distribution}. 
% The process proposed in~\method~ is illustrated in Figure \ref{fig:LPBNN}.
We illustrate an overview of an~\method~ layer in Figure~\ref{fig:LPBNN}.

The VAE modules are trained in the standard manner with the ELBO loss function~\cite{kingma13vae} jointly with the rest of the network. The final loss function is:
%$
%f(\vx) = \phi \left( (\vomega^{ft} (x\circ \hat{\vr}_j)) \circ \s_j \right)
%$. 
%Our training loss is
% \begin{multline}\label{eq:finalloss}
%     \mathcal{L}_{\text{LP-BNN}}(y_i, \Theta^{\text{LP-BNN}}) {=} - \mathbb{E}_{Q_{\phi}(\vz_j | \vr_j)}  \log \left(P( \vy_i| \vx_i, \theta^{\text{slow}},\s_j, \vz_j)\right) \\ +   \mbox{KL}(Q_{\phi}(\vz_j | \vr_j) \vert \vert P(\vz_j)) + \|\vr_j -\hat{\vr}_j \|^2_2 ,
% \end{multline}
% \begin{multline}\label{eq:final-loss-simple}
    % \mathcal{L}_{\text{LP-BNN}}(y_i, \Theta^{\text{LP-BNN}}) {=} - \mathbb{E}_{Q_{\phi}(\vz | \vr)}  \log \left(P( \vy_i| \vx_i, \theta^{\text{slow}},\s, \vz)\right) \\ +   \mbox{KL}(Q_{\phi}(\vz | \vr) \vert \vert P(\vz)) + \|\vr -g^{\mbox{dec}}_{\psi}(\vz) \|^2_2 ,
% \end{multline}
\begin{multline}\label{eq:loss-lpbnn}
    \mathcal{L}_{\scriptscriptstyle \text{LP-BNN}}(\ThetaLP) {=} -\sum_{\scriptscriptstyle (\vx_i,y_i) \in \mathcal{D}} {\mathbb{E}}_{\scriptscriptstyle \vz \sim Q_{\phi}(\vz | \vr)}  \log \left(P( {y_i} \mid \vx_i, \ThetaLP, \vz)\right) \\ 
     + 1/L\left( \mbox{KL}(Q_{\scriptstyle \phi}(\vz \mid \vr) \vert \vert P_{\scriptstyle \psi}(\vz)) + \|\vr -\hat{\vr} \|^2 \right),
\end{multline}
% \begin{equation}\label{eq:loss-lpbnn}
% \resizebox{0.95\hsize}{!}{%
%     $\mathcal{L}_{\scriptscriptstyle \text{LP-BNN}}(\ThetaLP) {=} -\sum_{\scriptscriptstyle (\vx_i,y_i) \in \mathcal{D}} {\mathbb{E}}_{\scriptscriptstyle \vz \sim Q_{\phi}(\vz | \vr)}  \log \left(P( \vy_i| \vx_i, \ThetaLP, \vz)\right) \nonumber$% 
%     }
% \end{equation}
% \begin{equation}\label{eq:loss-lpbnn}
% \resizebox{0.81\hsize}{!}{%
%     $ + \mbox{KL}(Q_{\scriptstyle \phi}(\vz | \vr) \vert \vert P_{\scriptstyle \psi}(\vz)) + \|\vr -\gdec(\vz) \|^2_2 ,$%
%     }
% \end{equation}
%\todo{Andrei: I forgot to add the regularization term. I will split the loss into multiple terms, \eg. $\mathcal{L}_{\scriptscriptstyle \text{VAE}}(\phi, \psi)$}
%\\
%\todo{Andrei: Reduce amount of notations and remove $\genc$ and $\gdec$ and use only $Q_{\phi}$ and $P_{\psi}$.}
%\\
where $\Theta^{\scriptscriptstyle \text{LP-BNN}} {=} \big\{ \theta^{\scriptstyle \text{slow}}, \theta^{\scriptstyle \text{fast}}{:}\{ \vr_j, \s_j \}_{j{=}1}^J, \theta^{\scriptstyle \text{variational}}{:} \{\phi,\psi\} \big\}$ \Gianni{and $L$ the number of layers}. %\Isa{il n'y a pas une pondération entre les termes, en particulier pour le dernier ?} -> Fait
The loss function is applied to all $J$ members of the ensemble.

At a first glance, the loss function $\losslpbnn$ bears some similarities with $\lossbnn$ (Eq.~\ref{eq:loss-BNN}). Both functions include likelihood and KL terms. The likelihood in $\lossbnn$, \ie the cross-entropy loss, depends on input data $\vx_i$ and on the parameters $\Theta$ sampled from $Q_{\nu}(\Theta)$, while $\losslpbnn$ {depends} on the latent variables $\vz_j$ sampled from $Q_{\phi}(\vz_j \mid \vr_j)$ that lead to the fast weights $\hat{\vr_j}$. It guides the weights towards useful values for the main task. The KL term in $\lossbnn$ enforces the per-weight prior, while in $\losslpbnn$ it preserves the consistency and simplicity of the common latent distribution of the weights $\vr_j$. In addition, $\losslpbnn$ has an input weight reconstruction loss (last term in Eq.~\ref{eq:loss-lpbnn}) ensuring that the generated weights $\hat{\vr}_j$ are still compatible with the rest of parameters of the network and do not cause high variance and instabilities during training, as typically occurs in standard BNNs~\cite{dusenberry2020efficient}.

At test time, we generate the LP-BNN ensemble on the fly by sampling the weights $\hat{\vr}_j$ from the encodings of $\vr_j$ to compute $\overline{\vW}_j$. For the final prediction we compute the empirical mean of the likelihoods of the ensemble:
\begin{equation}
P(y_i|\vx_i) = \frac{1}{J} \sum_{j=1}^J P(y_i \mid \vx_i, \theta^{\text{slow}},\s_j, \hat{\vr}_j)
\end{equation}
%\Isa{ et on prend y qui maximise cette moyenne je suppose ?} En fait dans un premiet 

\subsection{Discussion on \method}

%We organize the discussion as follows: \Gianni{in the supplementary materials} in section A.1 , we focus on our posterior covariance matrix. In section A.2, we develop a theory explaining the link between the number of ensemble models and the covariance matrix approximation. In section A.3, we discuss the computational complexity and proove that Lp-BNN is lighter than other techniques modeling the covariance explicitly. In section A.4, we discuss the stability of  LP-BNN. In this section, we discuss uncertainty and LP-BNN. 
% \Gianni{We organize the discussion in the supplementary material as follows: Section A.1 focuses on our posterior covariance matrix, while Section A.2 develops a theory explaining the link between the number of ensemble models and the covariance matrix approximation.  Sections A.3, A.4, and A.5. focus  on the computational complexity, the stability of LP-BNN, and the diversity of LP-BNN, respectively. In this section, we discuss uncertainty and LP-BNN. }

\ab{We discuss here the quality of the uncertainty from \method. The predictive uncertainty of a DNN stems from two main types of uncertainty~\cite{hora1996aleatory}: \emph{aleatoric uncertainty} and \emph{epistemic uncertainty}.}
The former is related to randomness, typically due to the noise in the data. The latter 
concerns 
\ab{finite size training datasets.}
%is associated with the fact that we are working with a learning dataset of finite size. 
% The epistemic uncertainty models the learning uncertainty and lack of knowledge. %\Isa{le dataset ne modelise rien du tout...}
\ab{The epistemic uncertainty captures the uncertainty in the DNN parameters and their lack of knowledge on the model that generated the training data.}

%\Isa{A citer peut-être : Hora S (1996) Aleatory and epistemic uncertainty in probability elicitation with an ex-ample from hazardous waste management. Reliability Engineering and System Safety54(2–3):217–223} \Gianni{fait}

%Classically during inference, BNN techniques allow,  by marginalizing the likelihood, to integrate the results of different DNNs weighted by the prior. So the study of uncertainty is based on the diversity that these DNNs could bring if we sample the different DNNs. Indeed let us consider we have ten realizations of a BNN. If the data is wrong, either because it is noisy or because it is too different from the training data, we want that the variance of the prediction of the 10 DNNs is huge. So when we average on these ten results, the predictions will have a low confidence score.

In BNN approaches, through likelihood marginalization over weights, the prediction is computed by integrating the outputs from different DNNs weighted by the posterior distribution \ab{(Eq.~\ref{eq:marginalization}), allowing us to conveniently capture both types of uncertainties~\cite{malinin2018predictive}.} 
% {Hence} the quality of the uncertainty estimates depends on the prediction diversity provided by the BNNs. 
\ab{The quality of the uncertainty estimates depends on the diversity of predictions and views provided by the BNN.}
% Fort \etal~\cite{fort2019deep} that DE~\cite{lakshminarayanan2017simple} have the best diversity among ensemble methods. 
\ab{DE~\cite{lakshminarayanan2017simple} achieve excellent diversity~\cite{fort2019deep} by mimicking BNN ensembles through training of multiple individual models.} 
Recently, Wilson \ab{and Izmailov}~\cite{wilson2020bayesian} proposed to 
% mix Deep Ensembles 
\ab{combine DE}
and \ab{BNNs} towards improving diversity \ab{further}. However, as 
% Deep Ensembles~  \cite{lakshminarayanan2017simple} are already very expensive
\ab{DE are already computationally demanding}, we argue \ab{that} BE is a more 
% computationally 
pragmatic choice for increasing \ab{the} diversity of our BNN\ab{, leading to better uncertainty quantification}.

Figure~\ref{fig:diversityall} shows 
% comparative results on  diversity provided by different DNNs. 
\ab{a qualitative comparison of the prediction diversity from different methods.}
% We train \method, BE and DE with WideResnet-28-10~\cite{zagoruyko2016wide} on CIFAR-10~\cite{krizhevsky2009learning}. We evaluate them on the test sets of CIFAR-10, CIFAR10-C~\cite{hendrycks2018benchmarking} and  SVHN~\cite{Netzer2011}.{The SVHN images (representing digits) are different from the training data\ab{, \ie predominant epistemic uncertainty}, while CIFAR10-C data are different from the training images due to noise corruption\ab{, \ie more aleatoric uncertainty}.
% The first row in Figure~\ref{fig:diversityall} shows the test images. The next three rows synthesize diversity results obtained with \method, BE and DE, respectively. For all methods we set the number of models to $J=4$. 
% The expected behavior is that different DNNs would not predict the same class on the out of distribution (OOD) images, reducing the confidence score of the DNN.
%The second row represents the results obtained on the test sets of CIFAR10, the third one on CIFAR10-C and the last one SVHN. \Gianni{The images of SVHN are different from the training data since it is composed of digits' images, while different kinds of noise corrupt the one of  CIFAR10-C.}
%We consider that $J$ is equal to $4$.
% The expected behavior is that different DNNs would not predict the same class on the out of distribution (OOD) images, reducing the confidence score of the DNN.
\ab{We compare \method, BE, and DE based on WRN-28-10~\cite{zagoruyko2016wide} trained on CIFAR-10~\cite{krizhevsky2009learning} and analyze predictions on CIFAR-10, CIFAR10-C~\cite{hendrycks2018benchmarking}, and SVHN~\cite{Netzer2011} test images. SVHN contains digits which have a different distribution from the training data, i.e., predominant epistemic uncertainty, while CIFAR10-C displays a distribution shift via noise corruption, i.e., more aleatoric uncertainty.
The expected behavior is that individual DNNs in an ensemble would predict different classes for OOD images and have higher entropy on the corrupted ones, reducing the confidence score of the ensemble.}
%We hope that the different DNNs will not predict the same class on the wrong images, reducing the confidence score of the DNN. 
We can see that the diversity of BE is 
% inferior
\ab{lower} for CIFAR10-C and SVHN, leading to \ab{poorer} results in Table~\ref{table:outofditribution}. 

\ab{In the supplementary we include additional discussions on our posterior covariance matrix ({\S}A.1), the link between the size of the ensemble and the covariance matrix approximation ({\S}A.2), computational complexity ({\S}A.3), training stability of \method ({\S}A.4), and diversity of \method ({\S}A.5).}

\begin{figure}[t!]
\renewcommand{\figurename}{Figure}
\renewcommand{\captionfont}{\small}
     \centering
        \begin{subfigure}[b]{0.16\linewidth}
        \caption*{{\textbf{Input}}}
        \includegraphics[width=\textwidth]{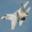}
        %\label{mnistacc}
        \end{subfigure}\;
        \begin{subfigure}[b]{0.22\linewidth}
        \caption*{{\textbf{LP-BNN}}}
        \includegraphics[width=\textwidth]{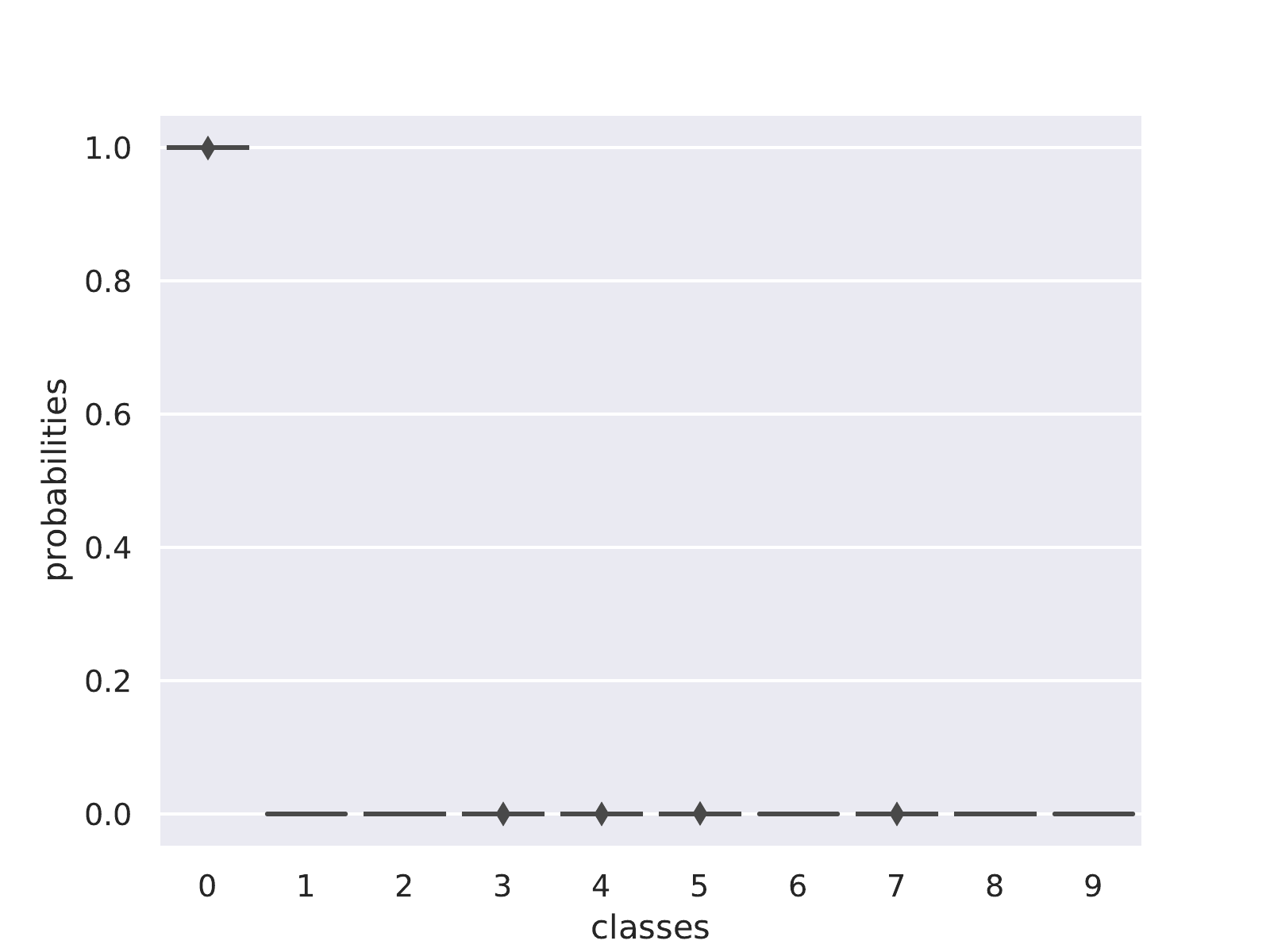}
        %\label{camvidacc}
        \end{subfigure}\;
        \begin{subfigure}[b]{0.22\linewidth}
        \caption*{{\textbf{BE}}}
        \includegraphics[width=\textwidth]{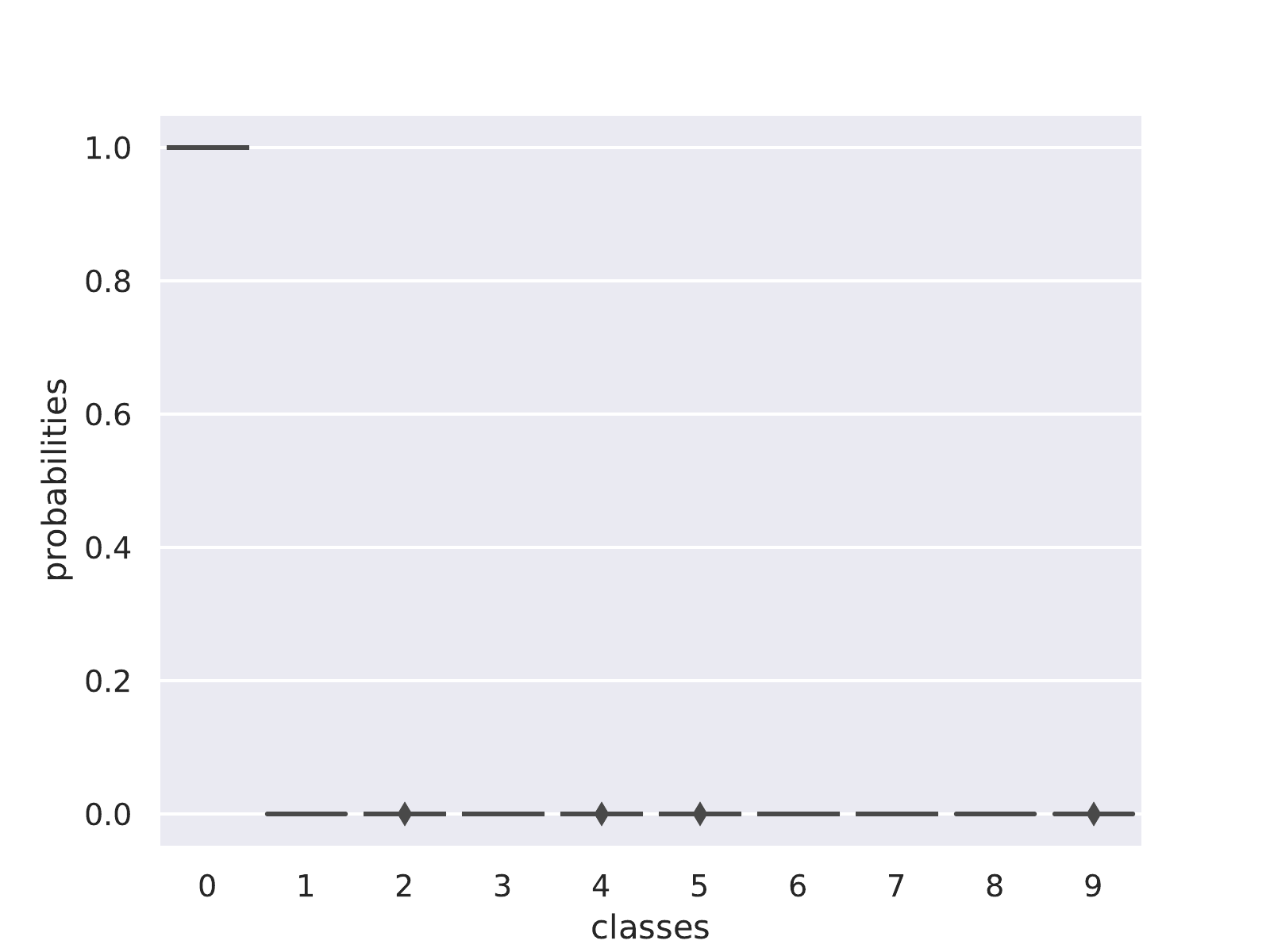}
        %\label{camvidprec}
        \end{subfigure}\;
         \begin{subfigure}[b]{0.22\linewidth}
        \caption*{{\textbf{DE }}}
        \includegraphics[width=\textwidth]{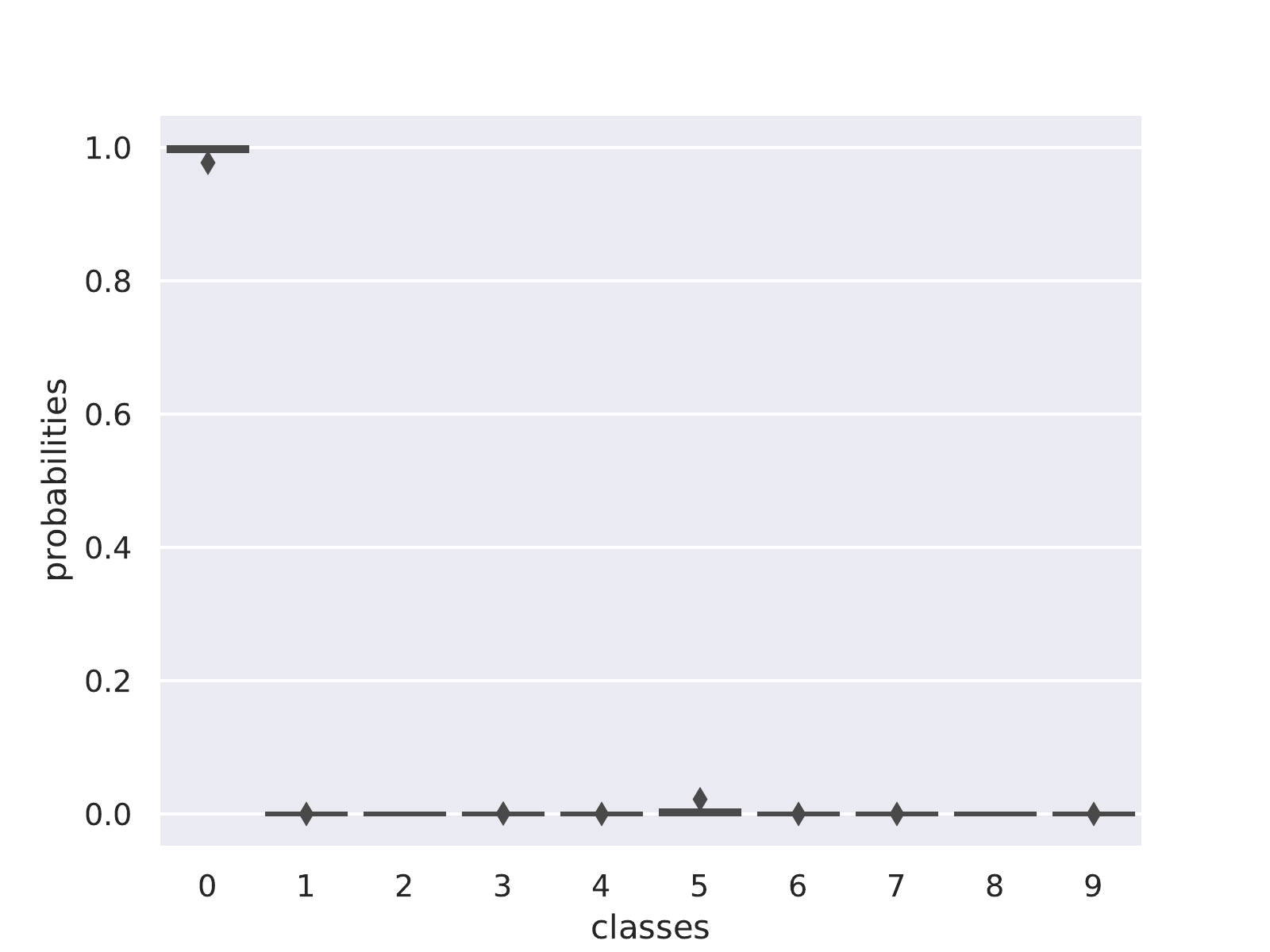}
        %\label{camvidprec}
        \end{subfigure}\;

        \begin{subfigure}[b]{0.16\linewidth}
        %\caption*{\large{\textbf{Input image}}}
        \includegraphics[width=\textwidth]{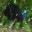}
        %\label{mnistacc}
        \end{subfigure}\;
        \begin{subfigure}[b]{0.22\linewidth}
        %\caption*{\large{\textbf{LP-BNN}}}
        \includegraphics[width=\textwidth]{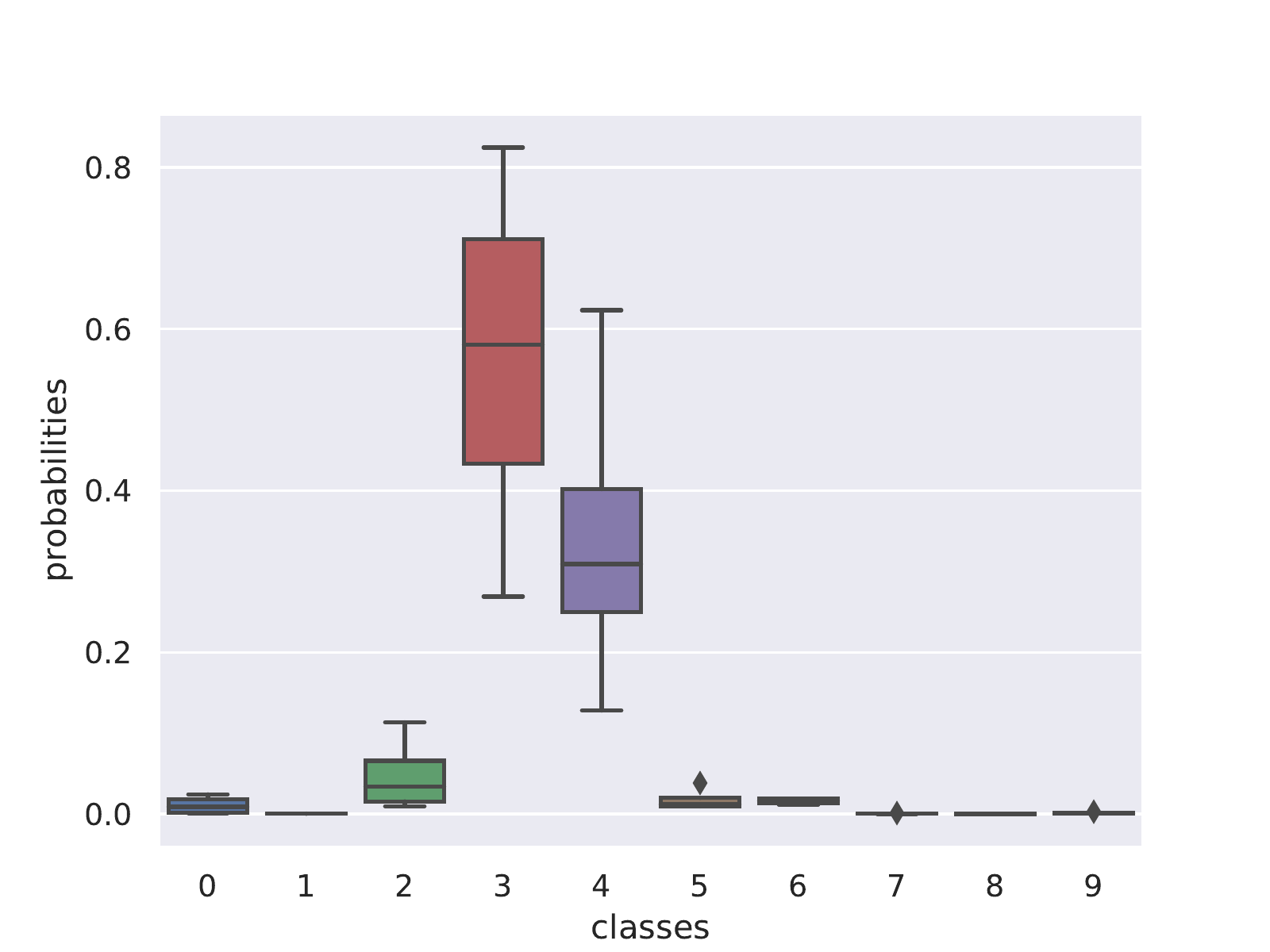}
        %\label{camvidacc}
        \end{subfigure}\;
        \begin{subfigure}[b]{0.22\linewidth}
        %\caption*{\large{\textbf{BE}}}
        \includegraphics[width=\textwidth]{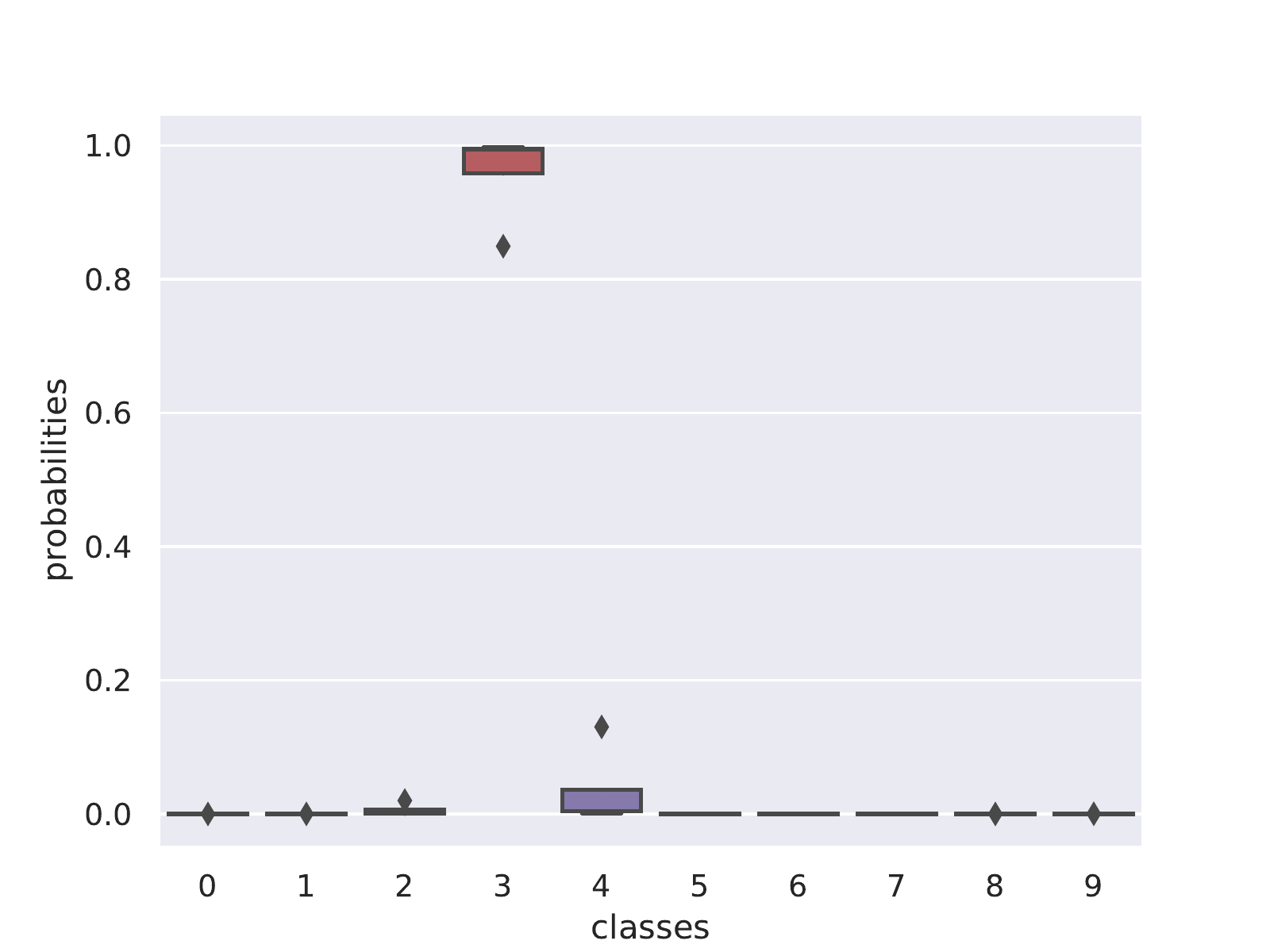}
        %\label{camvidprec}
        \end{subfigure}\;
         \begin{subfigure}[b]{0.22\linewidth}
        %\caption*{\large{\textbf{DE }}}
        \includegraphics[width=\textwidth]{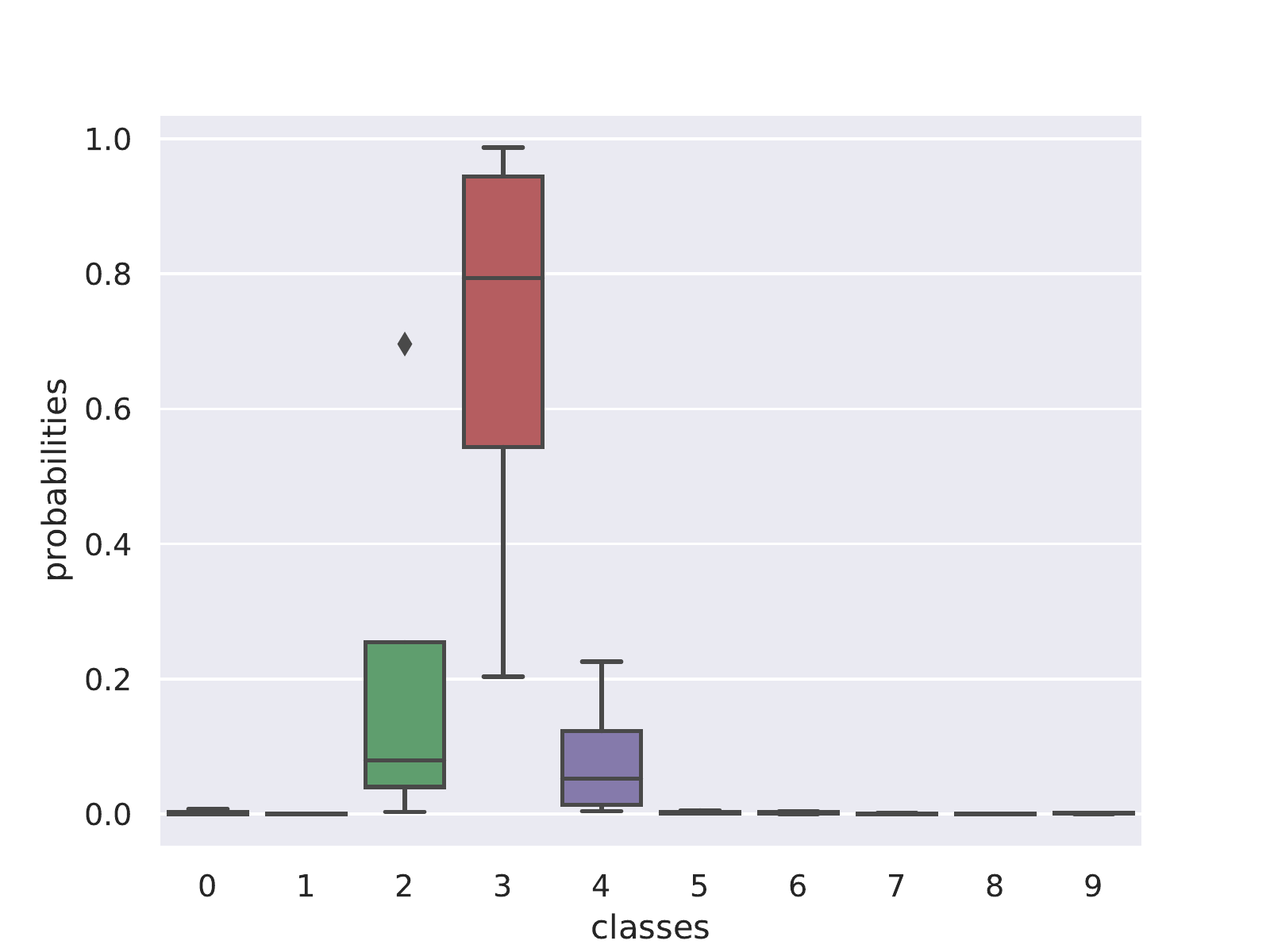}
        %\label{camvidprec}
        \end{subfigure}\;

        \begin{subfigure}[b]{0.16\linewidth}
        %\caption*{\large{\textbf{Input image}}}
        \includegraphics[width=\textwidth]{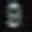}
        %\label{mnistacc}
        \end{subfigure}\;
        \begin{subfigure}[b]{0.22\linewidth}
        %\caption*{\large{\textbf{LP-BNN}}}
        \includegraphics[width=\textwidth]{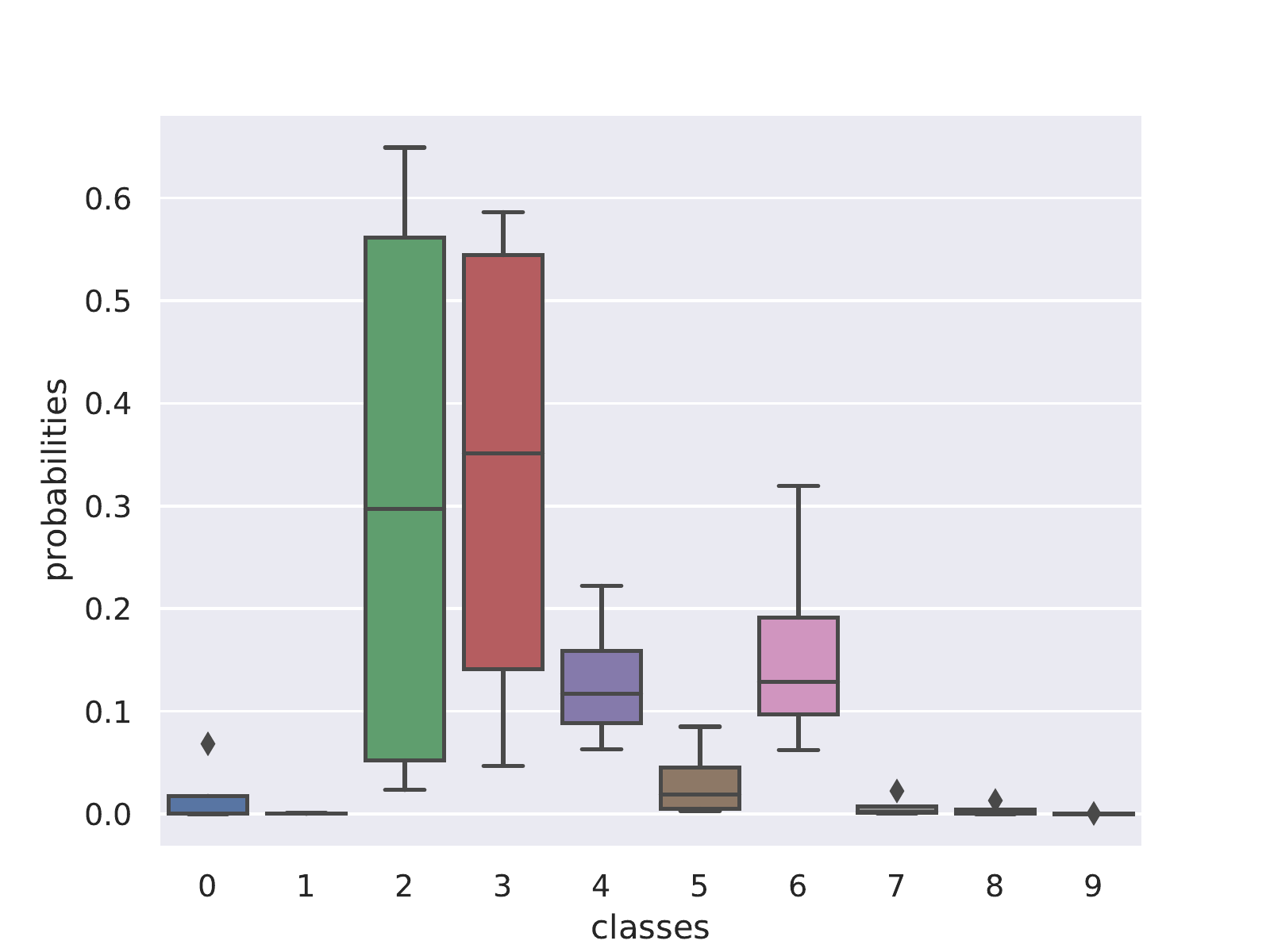}
        %\label{camvidacc}
        \end{subfigure}\;
        \begin{subfigure}[b]{0.22\linewidth}
        %\caption*{\large{\textbf{BE}}}
        \includegraphics[width=\textwidth]{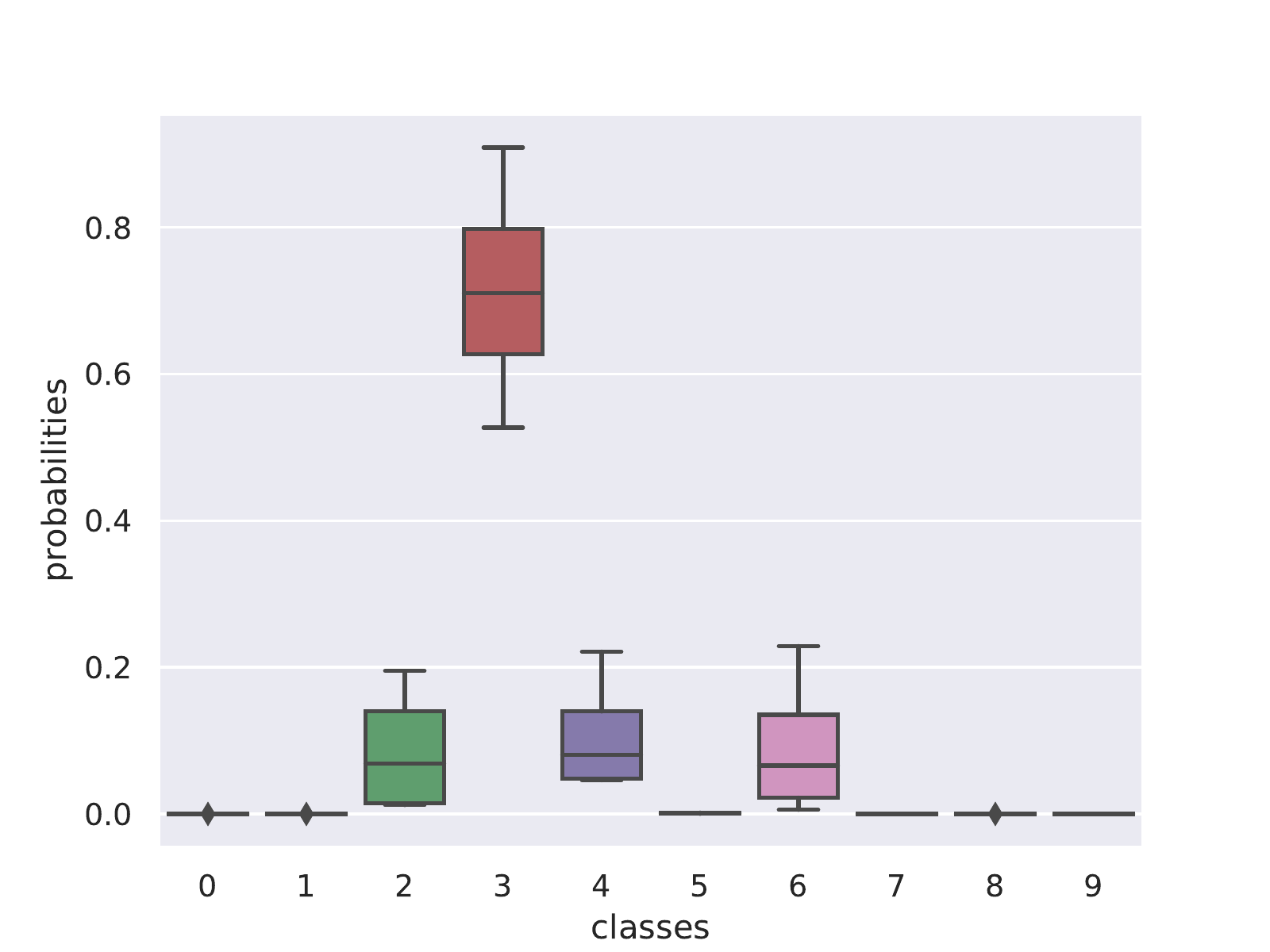}
        %\label{camvidprec}
        \end{subfigure}\;
         \begin{subfigure}[b]{0.22\linewidth}
        %\caption*{\large{\textbf{DE }}}
        \includegraphics[width=\textwidth]{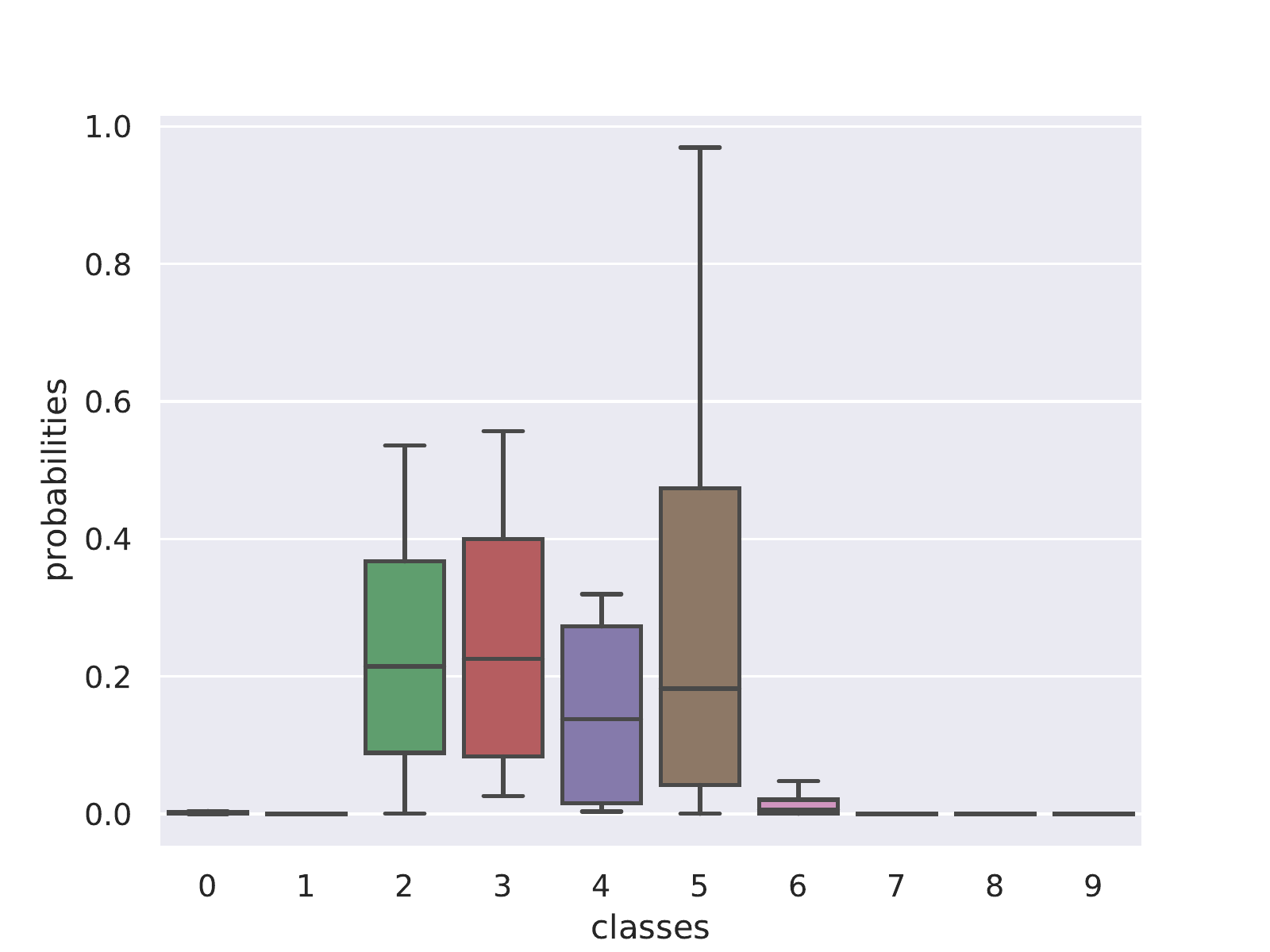}
        %\label{camvidprec}
        \end{subfigure}\;
\vspace{2mm}      
% \caption{\textbf{Illustration of the diversity provided by the different ensembles.} The first column contains in order one image from the test set of CIFAR-10, of  CIFAR-10-C  and of SVHN. The next three columns represent the corresponding outputs of the different sub models for the three ensembling algorithms being considered: \method, BatchEnsemble and Deep Ensembles. }
\caption{\ab{\textbf{Diversity of predictions of different ensemble methods.} \emph{Col.} $1$: in top-down order images from the test set of CIFAR-10,  CIFAR-10-C, and SVHN; \emph{Col.} $2{-}4$: outputs of different sub-models of the three ensemble techniques:
% considered
\method, BE~\cite{wen2020batchensemble}, and DE~\cite{lakshminarayanan2017simple}. For all methods we set the number of models to $J{=}4$. }}
\label{fig:diversityall}
\end{figure}

\section{Related work}\label{section:related}

% \abc{I would put related work right before the experiments as the introduction of the method itself is long with all the preliminaries.}\Gianni{I am ok with that}

\parag{Bayesian Deep Learning.} \ab{Bayesian approaches and neural networks have a long joint history~\cite{mackay1992bayesian,mackay1992practical, neal1995bayesian}. Early approaches relied on Markov chain Monte
Carlo variants for inference on BNNs, which was later replaced by variational inference (VI)~\cite{jordan1999introduction} in the context of deeper networks. Most of the modern approaches make use of VI with the mean-field approximation~\cite{hinton1993keeping, graves2011practical, blundell2015weight, hernandez2015probabilistic, gal2016dropout, mishkin2018slang} which conveniently makes posterior inference tractable. However this limits the expressivity of the posterior~\cite{mackay1992practical, foong2020expressiveness}. 
This drawback became subject of multiple attempts for structured-covariance approximations using matrix variate Gaussian priors~\cite{louizos2016structured, sun2017learning} or natural gradients~\cite{zhang2018noisy, mishkin2018slang}. However they further increase memory and time complexity over the original mean-field approximation. Recent methods proposed more simplistic BNNs by performing inference with structured priors only over the first and last layer~\cite{pearce2020structured} or just the last layer~\cite{riquelme2018deep, ovadia2019can}. Full covariance can be computed for shallow networks thanks to a meta-prior in a low-dimensional space where the VI can be performed~\cite{karaletsos2018probabilistic, karaletsos2020hierarchical}. 
Most BNNs are still challenging to train, underfit and are difficult to scale to big DNNs~\cite{ovadia2019can, dusenberry2020efficient}, while the issue of finding a proper prior is still open~\cite{wenzel2020good, fortuin2021bayesian}. Our approach builds upon the low dimensional fast weights from BE and on the stability of the VAEs, foregoing many of the shortcomings of BNN training.}

\parag{Ensemble Approaches.}
\ab{Ensembles 
% represent a well established technique that 
mimick and attain, to some extent, properties of BNNs~\cite{fort2019deep}. Deep Ensembles~\cite{lakshminarayanan2017simple} train multiple DNNs with different random initializations leading to excellent uncertainty quantification scores. The major inherent drawback in terms of computational and memory overhead has been subsequently addressed through multi-head networks~\cite{lee2015m}, snapshot-ensembles from intermediate training checkpoints~\cite{huang2017snapshot, garipov2018loss}, efficient computation of the posterior distribution from weight trajectories during training~\cite{maddox2019simple,franchi2019tradi}, use of multiple Dropout masks at test time~\cite{gal2016dropout}, multiple random perturbations to the weights of a pre-trained network~\cite{franchi2019tradi, mehrtash2020pep, atanov2018uncertainty}, multiple perturbation of the input image~\cite{ashukha2020pitfalls}, multiple low-rank weights tied to a backbone network~\cite{wen2020batchensemble}, simultaneous processing of multiple images by the same DNN~\cite{havasi2020training}. Most approaches still have a significant computational overhead for training or for prediction, while struggling with diversity~\cite{fort2019deep}.}

\parag{Dirichlet Networks (DNs).} 
\ab{DNs~\cite{malinin2018predictive,malinin2019reverse,sensoy2018evidential, charpentier2020posterior,tsiligkaridis2021information} bring a promising line of approaches that estimate uncertainty from a single network by parameterizing a Dirichlet distribution over its predictions. However, most of these methods~\cite{malinin2018predictive, malinin2019reverse} use OOD samples during training, which may be unrealistic in many applications~\cite{charpentier2020posterior}, or do not scale to bigger DNNs~\cite{joo2020being}.DNs have been developed only for classification tasks and extending them to regression requires further adjustments~\cite{malinin2020regression}, unlike \method that can be equally used for classification and regression.}

\section{Experiments and results}\label{section:experiments}

\subsection{Implementation details}
We {evaluate} the performance of LP-BNN in assessing  {the uncertainty of its predictions.} 
For our benchmark, we evaluate \method on different scenarios against several 
% related 
\ab{strong} baselines with different advantages in terms of performance, training or runtime: \ab{BE~\cite{wen2020batchensemble}, DE~\cite{lakshminarayanan2017simple}, Maximum Class Probability (MCP)~\cite{hendrycks2016baseline},} MC Dropout~\cite{gal2016dropout}, TRADI~ \cite{franchi2019tradi}, \Gianni{EDL \cite{sensoy2018evidential}, DUQ \cite{van2020uncertainty}, and MIMO \cite{havasi2020training}.} %  PostN \cite{charpentier2020posterior}
%for \Sev{Out of Distribution Detection (OOD)} detection  and provide a calibrated network under \Sev{on?} different scenarios. For the various experiments, we compare LPBNN to the deterministic technique, , Deep Ensemble, and Batch Ensemble [ref].

%\noindent\textbf{Metrics.}
%In the rest of this section, we will describe the experimental protocol, followed by the four experiments. 
%\Sev{Remarque remise : Est-ce qu'il ne faudrait pas un paragraphe metrics et un autre scenarios? la a la lecture c'est tout melange ca fait etrange}

First, we evaluate the predictive performance in terms of accuracy for image classification and mIoU~\cite{everingham2015pascal} for semantic segmentation, respectively.
Secondly, we evaluate the quality of the confidence scores provided by the DNNs by means of Expected Calibration Error (ECE)~\cite{guo2017calibration}. For ECE we use $M$-bin histograms of confidence scores and accuracy, and compute the average of $M$ bin-to-bin differences between the two histograms. Similarly to~\cite{guo2017calibration} we set $M=15$. 
% To evaluate the DNNs on 
% corrupted images and dataset shift, 
\ab{To evaluate the robustness to dataset shift via corrupted images, }
we first train the DNNs on CIFAR-10~\cite{krizhevsky2009learning} or CIFAR-100~\cite{krizhevsky2009learning} and then test on the corrupted versions of these datasets~\cite{hendrycks2018benchmarking}. The corruptions include different types of noise, 
blurring, and some other transformations that alter the quality of the images. 
For this scenario, similarly to~\cite{wen2020improving}, we use as evaluation measures the Corrupted Accuracy (cA) and Corrupted Expected Calibration Error (cE), {that offer} a better understanding of the behavior of our DNN when facing \ab{shift of data distribution and} aleatoric uncertainty. 
%\Isa{Verifier qu'on introduit bien plus haut la distinction entre aleatoric (statistical, referring to randomness) uncertainty vs epistemic (systematic, caused by a lack of knowledge) uncertainty}\Gianni{Fait j'ai deplace}

In order to evaluate the epistemic uncertainty, we propose to assess the {OOD} {detection} performance. This scenario typically consists in training a DNN over a dataset following a given distribution, and testing it on data coming from this distribution and data from another distribution \ab{, not seen during training}. We 
% would 
\ab{quantify} the confidence of the DNN predictions in this setting 
% by using theirs 
\ab{through their} prediction scores, i.e., output softmax values. We use the same indicators of the accuracy of detecting OOD data as in~\cite{hendrycks2016baseline}: AUC, AUPR, and the FPR-95\%-TPR.
{These indicators measure 
whether the DNN model lacks knowledge regarding some specific data and how reliable are its predictions.} %It is useful in various real-life scenarios since it can help us know if we can rely on the DNN prediction. \Isa{supprimer la derniere phrase ?}

\begin{table*}[!t]
\renewcommand{\figurename}{Table}
\renewcommand{\captionfont}{\small}
\begin{center}
\scalebox{0.65}
{
\begin{tabular}{l|ccccccc|cccc}
\toprule
 &   \multicolumn{7}{c|}{CIFAR-10}   &  \multicolumn{4}{c}{CIFAR-100}        \\ 
Method                  & Acc $\uparrow$   & AUC $\uparrow$  & AUPR $\uparrow$   & FPR-95-TPR $\downarrow$   & ECE $\downarrow$    & cA $\uparrow$   & cE     $\downarrow$  &  Acc $\uparrow$   & ECE  $\downarrow$    & cA  $\uparrow$  & cE  $\downarrow$    \\ 
\midrule
MCP + cutout~\cite{hendrycks2016baseline}            & 96.33 & 0.9600 & 0.9767 & 0.115 &0.0207 & 32.98 & 0.6167  & 80.19 & 0.1228 & 19.33 & 0.7844 \\ 
\midrule
MC dropout~\cite{gal2016dropout}              & 95.95 & 0.9126 & 0.9511 & 0.282 & 0.0172 & 32.32 & 0.6673  & 75.40 & 0.0694 & 19.33 & 0.5830 \\ 
\midrule
MC dropout +cutout~\cite{gal2016dropout}     & 96.50 & 0.9273 & 0.9603 & 0.242 & 0.0117 & 32.35 & 0.6403  & 77.92 & 0.0672 & 27.66 & 0.5909 \\ 
%\midrule
%PostN \cite{charpentier2020posterior}  & 84.73 & 0.998 & 0.998 & 0.0191 & 0.1711 & x6 & x7  & x8 & x9 & x10 & x11 \\ 
\midrule
DUQ~\cite{van2020uncertainty}$^\dagger$  & 87.48 & 0.7083 & 0.8114 & 0.698 & 0.3983 & 64.89 & 0.2542  & - & - & - & - \\
\midrule 
DUQ Resnet18~\cite{van2020uncertainty}$^\ddagger$  & 93.36 & 0.8994 & 0.9213 & 0.1964& 0.0131 & 69.01 & 0.5059  & - & - & - & - \\ 
\midrule 
EDL~\cite{sensoy2018evidential}$^\dagger$  & 85.73 & 0.9002 & 0.9198 & 0.247 & 0.0904 & 59.54 & 0.3412  & - & - & - & - \\ 
\midrule
MIMO~\cite{havasi2020training} & 94.96 & 0.9387 & 0.9648 & 0.175 & 0.0300 & \textbf{69.99} & 0.1846  & 0.7869 & 0.1018 & 0.4735 & 0.2832 \\ 
\midrule
Deep Ensembles + cutout~\cite{lakshminarayanan2017simple} & \textbf{96.74} & \textbf{0.9803} & \textbf{0.9896} & \textbf{0.071} & \textbf{0.0093} & 68.75 & 0.1414  & \textbf{83.01} & \textbf{0.0673} & 47.35 & \textbf{0.2023} \\ 
\midrule
BatchEnsembles  + cutout~\cite{wen2020batchensemble}   & 96.48 & 0.9540 & 0.9731 & 0.132 & 0.0167 & \textbf{71.67} & 0.1928  & 81.27 & 0.0912 & 47.44 & 0.2909 \\ \midrule
LP-BNN (ours) + cutout        & 95.02 & \textbf{0.9691} & \textbf{0.9836} & \textbf{0.103} & \textbf{0.0094} & \textbf{69.51} &\textbf{ 0.1197}  & 79.3 & 0.0702 & \textbf{48.40} & \textbf{0.2224} \\ 
\bottomrule
\end{tabular}
} %scalebox
\end{center}
\vspace{-2mm}
\caption{\ab{\textbf{Comparative results for image classification tasks}. We evaluate on CIFAR-10 and CIFAR-100 for the tasks: in-domain classification, out-of-distribution detection with SVHN (CIFAR-10 only), robustness to distribution shift (CIFAR-10-C, CIFAR-100-C)}. We run all methods ourselves in similar settings using publicly available code for related methods. Results are averaged over three seeds. \ab{$^\dagger$: We did not manage to scale these methods to WRN-28-10 on CIFAR-100. A similar finding for EDL was reported in ~\cite{joo2020being}. $^\ddagger$ DUQ does not scale on CIFAR-100 and it does not perfectly scale to WRN-28-10 on CIFAR-10 so we train it with Resnet 18 \cite{he2016deep} architecture like in the original paper. }}
\label{table:tab1}
\vspace{-4mm}
\end{table*}
%\Isa{les 4 chiffres après la virgule sont ils significatifs ?}  pour l'instant 

%MC Dropout~\cite{gal2016dropout}, TRADI~ \cite{franchi2019tradi}, \Gianni{EDL \cite{sensoy2018evidential}, DUQ \cite{van2020uncertainty}, PostN \cite{charpentier2020posterior}, and MIMO \cite{havasi2020training}.

\begin{figure}[!t]
\renewcommand{\captionfont}{\small}
\centering
 \includegraphics[width=0.80\linewidth]{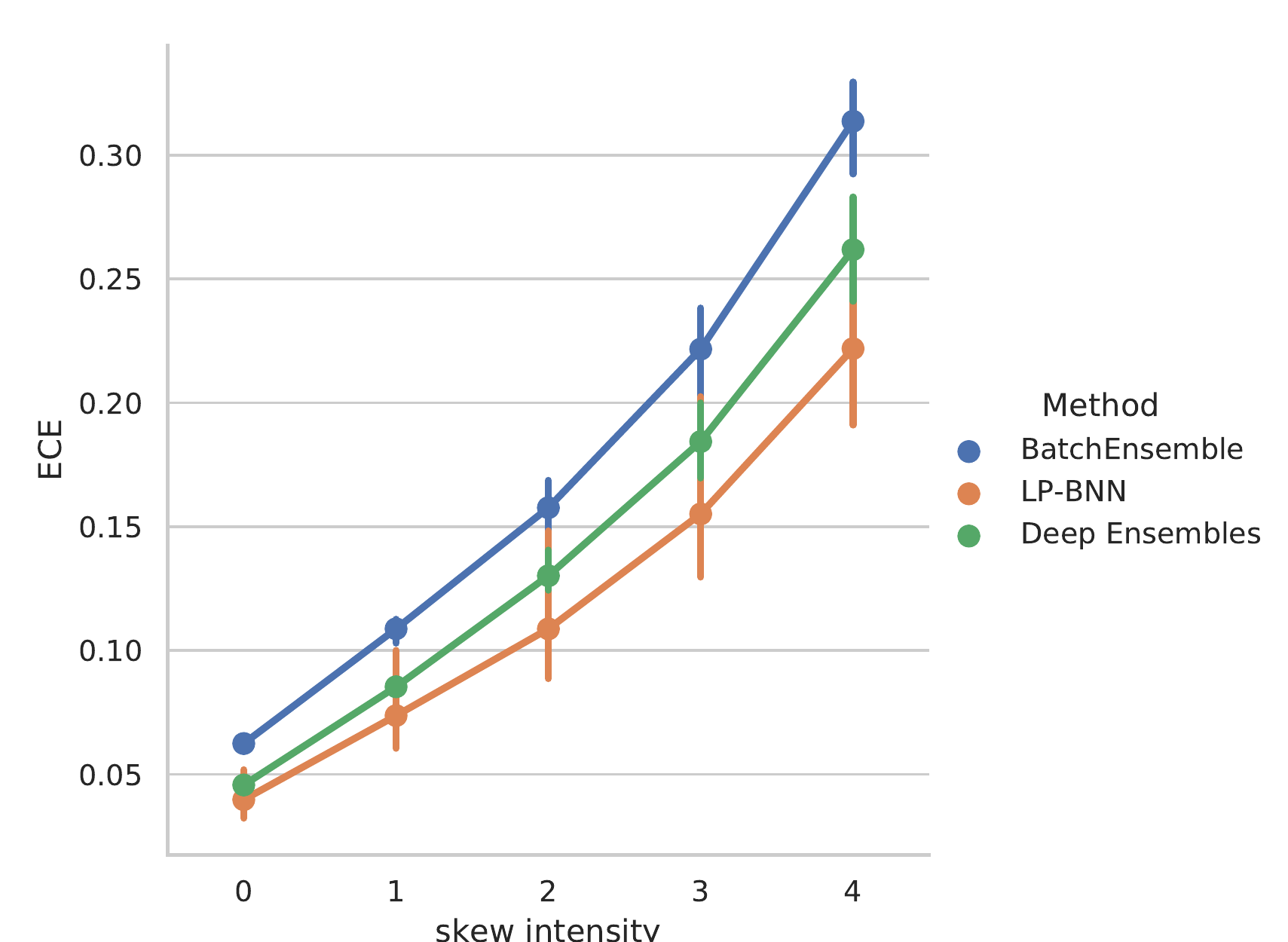}
\caption{ 
% Study of the evolution of the ECE  of \method,~Deep Ensembles and BatchEnsemble on CIFAR-10-C with different levels of corruption.
\ab{\textbf{Calibration at different levels of corruption.} We report ECE scores for \method, BE~\cite{wen2020batchensemble}, and DE~\cite{lakshminarayanan2017simple} on CIFAR-10-C.}
}
\label{fig:ECELBNN}
\end{figure}

\subsection{Image classification with CIFAR-10/100~\cite{krizhevsky2009learning}}\label{subsection:Experiments:cifar}
\noindent\textbf{Protocol.}
Here we train on CIFAR-10~\cite{krizhevsky2009learning} composed of 10 classes. % airplanes, cars, birds, cats, deer, dogs, frogs, horses, ships and trucks \Isa{ou simplement composed of 10 classes ?}. \Gianni{FAIT}
For  CIFAR-10 we consider as OOD the SVHN dataset~\cite{Netzer2011}. Since SVHN is a color image dataset of digits, it guarantees that the OOD data comes from a distribution different from those of CIFAR-10.
We use 
% WideResNet-28-10~\cite{zagoruyko2016wide}
\ab{WRN-28-10~\cite{zagoruyko2016wide}} for all methods, a popular architecture for this dataset, and evaluate on CIFAR-10-C~\cite{hendrycks2018benchmarking}. 
For CIFAR-100~\cite{krizhevsky2009learning} we use again 
% WideResNet-28-10 \cite{zagoruyko2016wide} 
\ab{WRN-28-10}
and 
% test 
\ab{evaluate} on the test sets of CIFAR-100 and 
% of 
CIFAR-100-C~\cite{hendrycks2018benchmarking}.
Note that for all DNNs, even for DE, 
% results are averaged over 3 random seeds, 
\ab{we average results over three random seeds} for statistical relevance. % \Isa{preciser ce que tu appelles seeds ici ?} \Gianni{c'est la base de la programmation}
We use cutout~\cite{devries2017improved} as data augmentation, as commonly used for these datasets. {Please find in the supplementary the hyperparameters for this experiment.}
%\Sev{Question: il y a plein d'info, y aurait-il un moyen de présenter tout cela dans un tableau synthétique qui résume tous les choix pour toutes les approches (réseau pour entrainement, caractéristiques, dataset, etc. \Gianni{Je laisse on le fait dans le supplementary}?}

\noindent\textbf{Discussion.} We illustrate results for this experiment in Table~\ref{table:outofditribution}. We notice that DE with cutout outperforms other methods on most of the metrics except ECE, cA, and cE on CIFAR-10, and cA on CIFAR-100, where \method~ achieves state of the art results. This means that \method~ is competitive for aleatoric %\Sev{aleatoric est le bon terme ? pas random ? \Gianni{oui oui j en suis sure}} 
uncertainty estimation. In fact, ECE is calculated on the test set of CIFAR-10 and CIFAR-100, so it mostly measures the reliability of the confidence score in the training distribution. cA and cE are evaluated on corrupted versions of CIFAR-10 and CIFAR-100, which amounts to quantifying the aleatoric %\Sev{aleatoric est le bon terme ? pas random ?\Gianni{Dans la communaute du DEEP learning Computer vision et Machine learning ce sont les bon noms. Dans le precedent artile Isabelle posait la meme question.}} 
uncertainty. We can see that for this kind of uncertainty, \method~ achieves state of the art performance.
 On the other hand, for epistemic uncertainty, we can see that DE always attain best results. %Yet, \method, in most cases, {performs close to} DE. Computation wise, DE takes $52$ hours to train on CIFAR-10, while our solution needs 2 times less, $26$ hours and $30$ minutes. 
Overall, our \method~ is more computationally efficient 
%less greedy in calculation time \Sev{/cost expensive ?}
while providing better results for the aleatoric %\Sev{aleatoric est le bon terme ? pas random ?} 
uncertainty.  
\ab{Computation wise, DE takes $52$ hours to train on CIFAR-10, while \method needs 2 times less, $26$ hours and $30$ minutes.}
% {As a comparison, it takes $26$ hours for BE to be trained on CIFAR-10, significantly less than \method.
%\Isa{est-ce utile ? c'est la meme duree et on ne conclut rien...} \Gianni{c'est strategique on a pas le droit de faire des experience en plus pendant la rebutal la le resultats est présent donc au pire on présente juste un résultat pas vue par un reviewer...}
In Figure~\ref{fig:ECELBNN} and Table~\ref{table:tab1}, we observe that our method exhibits 
% the best global ECE on CIFAR-10-C, as well as the best ECE for the stronger corruptions
\ab{top ECE score on CIFAR-10-C, as well as for the stronger corruptions.}

 \begin{table}[t!]
% \vspace{-8pt}
\renewcommand{\figurename}{Table}
\renewcommand{\captionfont}{\small}
\centering
 \scalebox{0.6}
 {
\begin{tabular}{l l | r r r r r}
\toprule
  &     & Vanilla & BatchEnsemble      & Deep Ensembles  & TRADI    & LP-BNN     \\
\midrule
%\multirow{4}{*}{Training}&  Time (s)  &   30,128    & 39,657    & 120,512  & 35,642  & 39,978     \\
\multirow{4}{*}{Training}&  Time (s)  &  1,506    & 1,983    & 6,026  & 1,782  & 1,999     \\
  &  for 1 epochs  &   \relative{1}   &  \relative{$\times$1.31}   &  \relative{$\times$4.0} &  \relative{$\times$1.18} &  \relative{$\times$1.33}       \\
  & Memory (MiB) &   8,848 & 9,884 & 35,392   &9,040  & 9,888  \\
  & &   \relative{1}   &  \relative{$\times$1.11}   &  \relative{$\times$4.0}  &  \relative{$\times$1.02}&  \relative{$\times$1.11}     \\
\midrule
% \multirow{4}{*}{Testing} & Time (s) &  307.9   & 846.1   & 1,231.7  & 857.1   & 1,231.7  \\
%\multirow{4}{*}{Testing} & Time (s) &  308   & 846   & 1,232 & 1,232  & 857     \\
\multirow{4}{*}{Testing} & Time (s) &  0.21   & 0.56   & 0.84 & 0.84   & 0.57     \\
  & on 1 image  &  \relative{1}  & \relative{$\times$2.67}   & \relative{$\times$4.0} & \relative{$\times$4.0} &  \relative{$\times$2.71}    \\ 
  &  Memory (MiB) &  1,884 & 4,114 & 7,536 & 7,536    & 4,114     \\
  &  & \relative{1}   &  \relative{$\times$2.18}  & \relative{$\times$4.0}  & \relative{$\times$4.0} &   \relative{$\times$2.18}     \\
\bottomrule
\end{tabular}
}
\vspace{-1mm}
\caption{\ab{\textbf{Runtime and memory analysis.} Numbers correspond to StreetHazards images processed with DeepLabv3+ ResNet-50 with PyTorch on a PC: Intel Core i9-9820X and $1\times$ GeForce RTX 2080Ti. Colored numbers are relative to vanilla approach. Mini-batch size for training is $4$ and for testing $1$.}
}
\vspace{-6mm}
\label{table:time}
\end{table}

\subsection{Semantic segmentation}

Next, we evaluate semantic segmentation, a task of interest for autonomous driving, 
% for which 
\ab{where} high capacity DNNs are used for processing high resolution images with complex urban scenery \ab{with strong class imbalance.}

\noindent\textbf{StreetHazards~\cite{hendrycks2019anomalyseg}.}
StreetHazards is a large-scale dataset that consists of different sets of synthetic images of street scenes. More precisely, this dataset is composed of $5,125$ images for training and $1,500$ test images.
The training dataset contains pixel-wise annotations for $13$ classes. The test dataset comprises $13$ training classes and $250$ OOD classes, unseen in the training set, making it possible to test the robustness of the algorithm when facing a diversity of possible scenarios. 
For this experiment, we used DeepLabv3+~\cite{chen2018encoder} with a ResNet-50 encoder~\cite{he2016deep}. Following the implementation in~\cite{hendrycks2019anomalyseg}, most papers use PSPNet~\cite{zhao2017pyramid} that aggregates predictions over multiple scales, an ensembling that can obfuscate in the evaluation the uncertainty contribution of a method. This can partially explain the excellent performance of MCP on the original settings~\cite{hendrycks2019anomalyseg}. We propose using DeepLabv3+ instead, as it enables a clearer evaluation of the predictive uncertainty.
 We propose two DeepLabv3+ variants as follows. DeepLabv3+ is composed of an encoder network and a decoder network; in the first version, we change the decoder by replacing all the convolutions with our new version of \method~convolutions and leave the encoder unchanged. In the second variant we use weight standardization~\cite{qiao2019rethinking} on the convolutional layers of the decoder, replacing batch normalization \cite{ioffe2015batch} in the decoder with group normalization \cite{wu2018group}\ab{, to better balance mini-batch size and ensemble size.} We denote the first version LP-BNN and the second one LP-BNN + GN.

\begin{table*}[htbp]
\renewcommand{\figurename}{Table}
\renewcommand{\captionfont}{\small}
 \vspace{-2mm}
 \begin{center}
 \scalebox{0.65}
 {
 \begin{tabular}{c l  c c c c c }
 \toprule
 Dataset & OOD method  & mIoU $\uparrow$ & AUC $\uparrow$  & AUPR $\uparrow$ & FPR-95-TPR $\downarrow$ & ECE $\downarrow$ \\ 
 \midrule
\multirow{6}{*}{\shortstack[c]{\textbf{StreetHazards} \\ DeepLabv3+ \\ ResNet50}}  & Baseline (MCP)~\cite{hendrycks2016baseline} & 53.90  & 0.8660& 0.0691 & 0.3574 & 0.0652 \\ 
             & TRADI      \cite{franchi2019tradi} &  52.46	& 0.8739 & 0.0693 & 0.3826 &0.0633\\
           & Deep Ensembles  \cite{lakshminarayanan2017simple}& 55.59 & 0.8794 & \textbf{0.0832} & 0.3029 & 0.0533 \\
            & MIMO  \cite{havasi2020training} & 55.44 &  0.8738 &  0.0690 & 0.3266 & \textbf{0.0557} \\ 
                  & BatchEnsemble \cite{wen2020batchensemble}  &\textbf{56.16}  & 0.8817 & 0.0759 & 0.3285 & 0.0609 \\ 
                                 & LP-BNN (ours)  &  54.50 & \textbf{0.8833} & 0.0718 & 0.3261 & \textbf{0.0520}  \\
           & LP-BNN + GN (ours)  & \textbf{56.12} & \textbf{0.8908} & 0.0742 &\textbf{0.2999} & \textbf{0.0593}  \\
\midrule 
  \multirow{6}{*}{\shortstack[c]{\textbf{BDD-Anomaly} \\ DeepLabv3+ \\ ResNet50}}                               & Baseline (MCP)~\cite{hendrycks2016baseline} &  47.63  & 0.8515  & 0.0450  & 0.2878  & 0.1768 \\ 

    & TRADI   \cite{franchi2019tradi}   & 44.26	& 0.8480	& 0.0454	& 0.3687	& 0.1661 \\ 
 & Deep Ensembles  \cite{lakshminarayanan2017simple}& \textbf{51.07}  & 0.8480  & 0.0524  & \textbf{0.2855 } &\textbf{0.1419 } \\
  & MIMO  \cite{havasi2020training}& 47.20  &  0.8438  & 0.0432  & 0.3524 &0.1633  \\

                    & BatchEnsemble  \cite{wen2020batchensemble}  &  48.09  & 0.8427  & 0.0449  & 0.3017  & 0.1690  \\ 
                                  & LP-BNN  (ours)   & {49.01}  &\textbf{0.8532 } & 0.0452  & 0.2947  & 0.1716  \\ 
                                  &LP-BNN + GN (ours)   & 47.15  & \textbf{0.8553}  & \textbf{0.0577}  & 0.2866  & 0.1623 \\ 
\bottomrule
 \end{tabular}
 } %scalebox
 \end{center}
 \vspace{-3mm}
 \caption{\ab{\textbf{Comparative results on the OOD task for semantic segmentation.}  We run all methods ourselves in similar settings using publicly available code for related methods. Results are averaged over three seeds.}\label{table:outofditribution}}
 \vspace{-2mm}
 \end{table*}

\begin{figure*}[t!]
\renewcommand{\figurename}{Figure}
\renewcommand{\captionfont}{\small}
     \centering
        \begin{subfigure}[b]{0.16\linewidth}
        \caption*{\textbf{Input image}}
        \includegraphics[width=\textwidth]{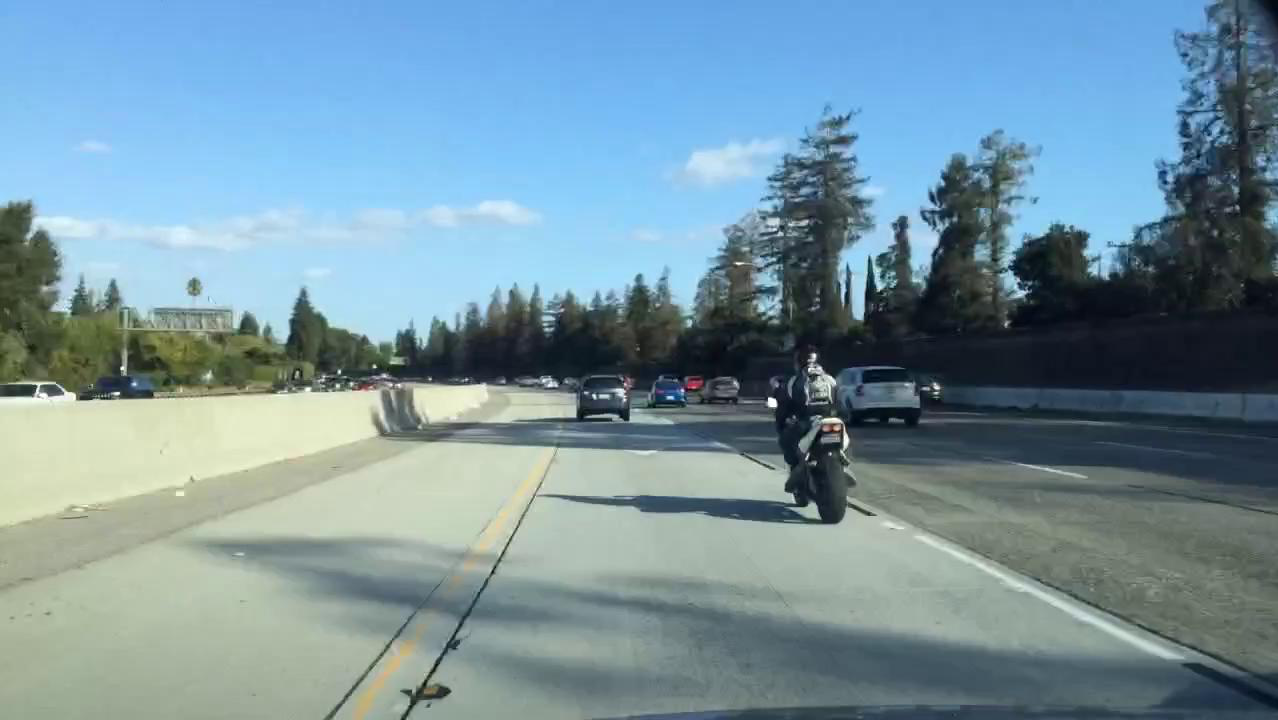}
        %\label{mnistacc}
        \end{subfigure}\;
        \begin{subfigure}[b]{0.16\linewidth}
        \caption*{\textbf{MCP}}
        \includegraphics[width=\textwidth]{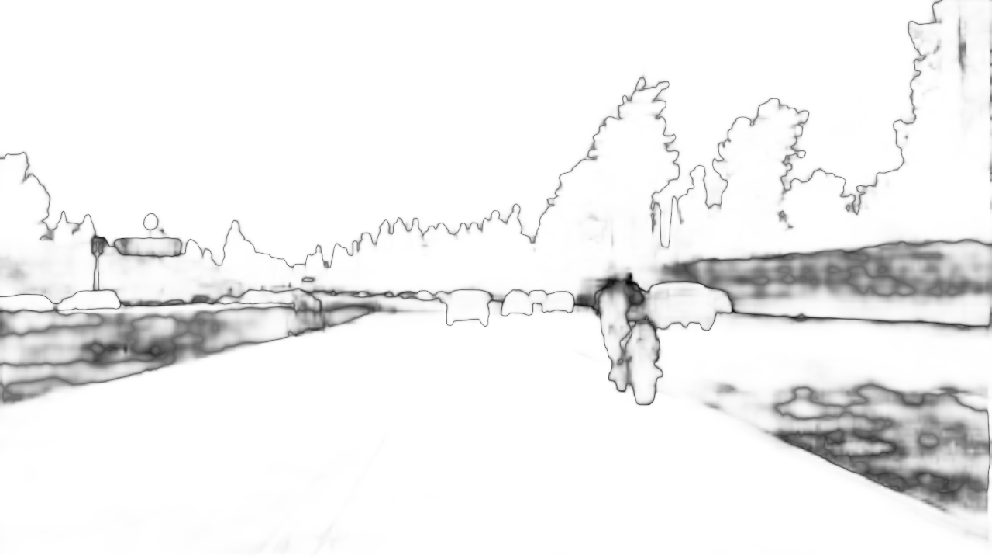}
        %\label{camvidacc}
        \end{subfigure}\;
        \begin{subfigure}[b]{0.16\linewidth}
        \caption*{\textbf{BE}}
        \includegraphics[width=\textwidth]{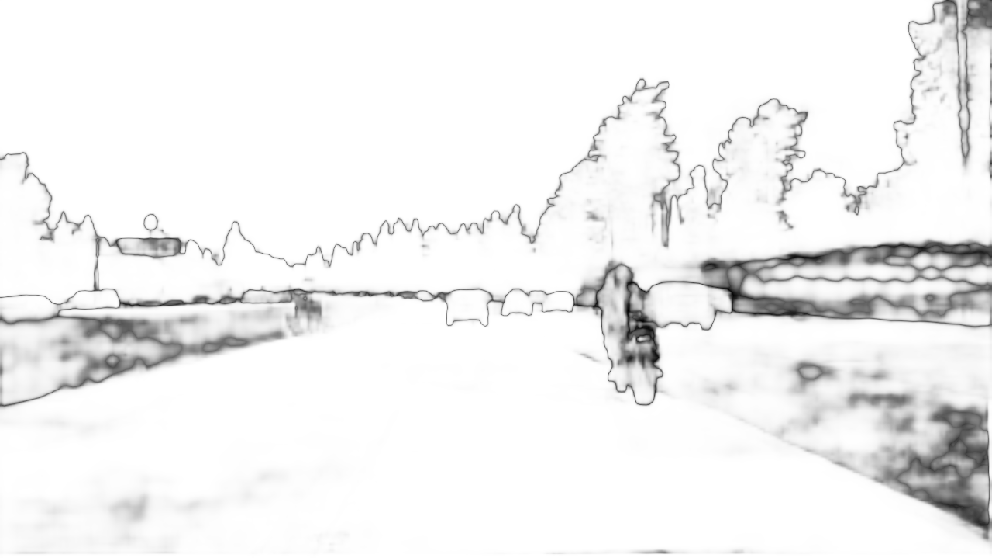}
        %\label{camvidprec}
        \end{subfigure}\;
         \begin{subfigure}[b]{0.16\linewidth}
        \caption*{\textbf{LP-BNN}}
        \includegraphics[width=\textwidth]{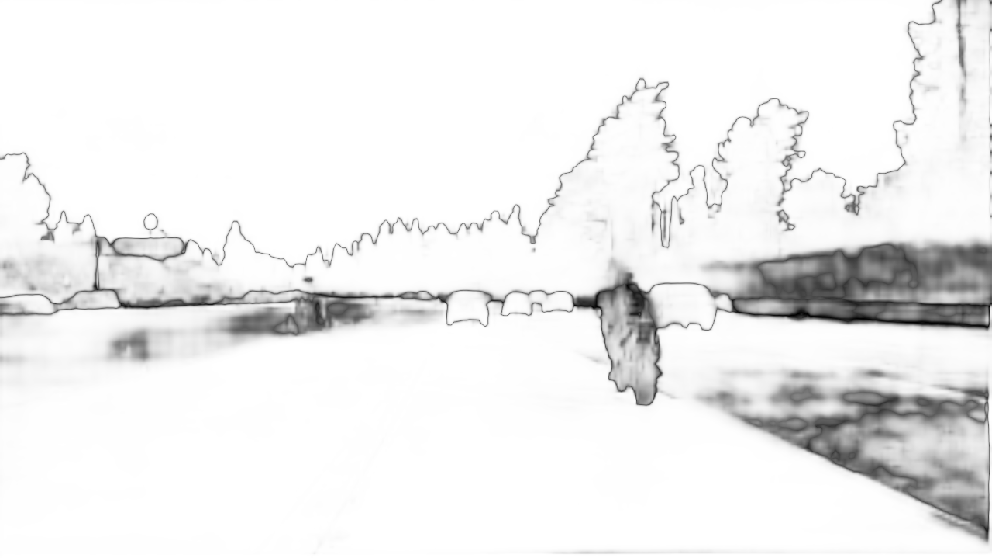}
        %\label{camvidprec}
        \end{subfigure}\;

        \begin{subfigure}[b]{0.16\linewidth}
        %\caption*{\large{\textbf{Input image}}}
        \includegraphics[width=\textwidth]{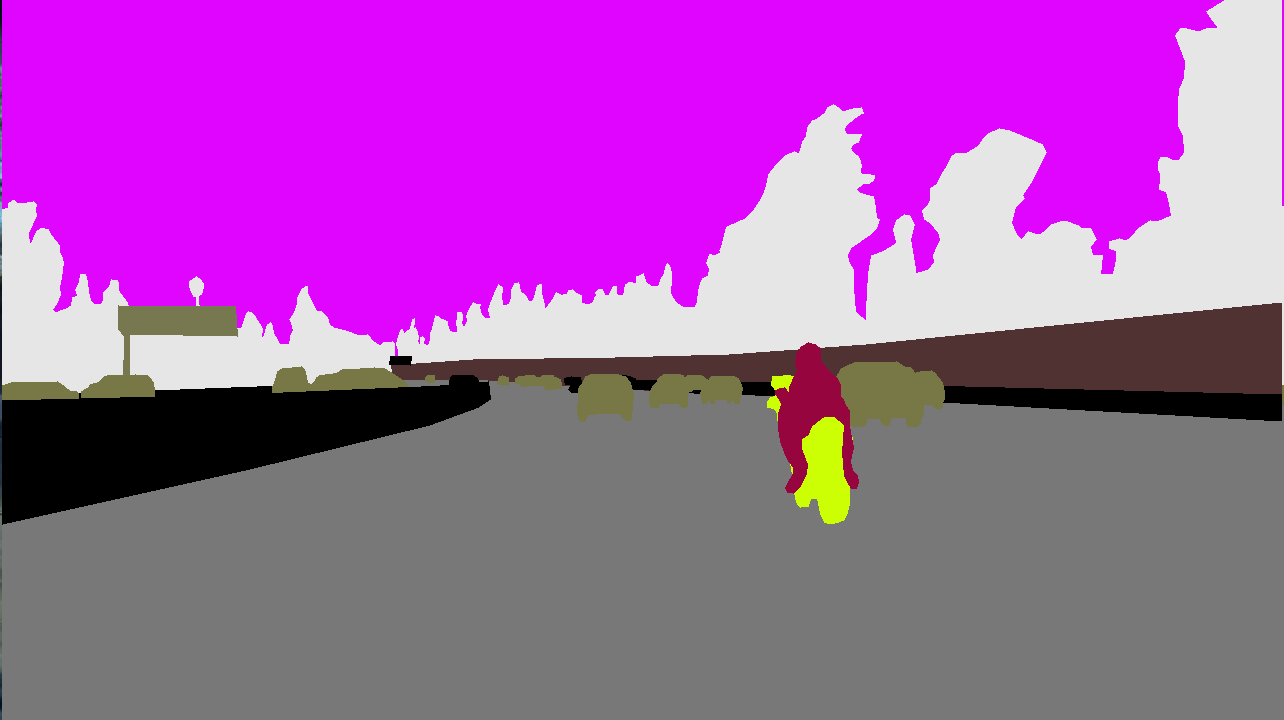}
        %\label{mnistacc}
        \end{subfigure}\;
        \begin{subfigure}[b]{0.16\linewidth}
        %\caption*{\large{\textbf{MCP}}}
        \includegraphics[width=\textwidth]{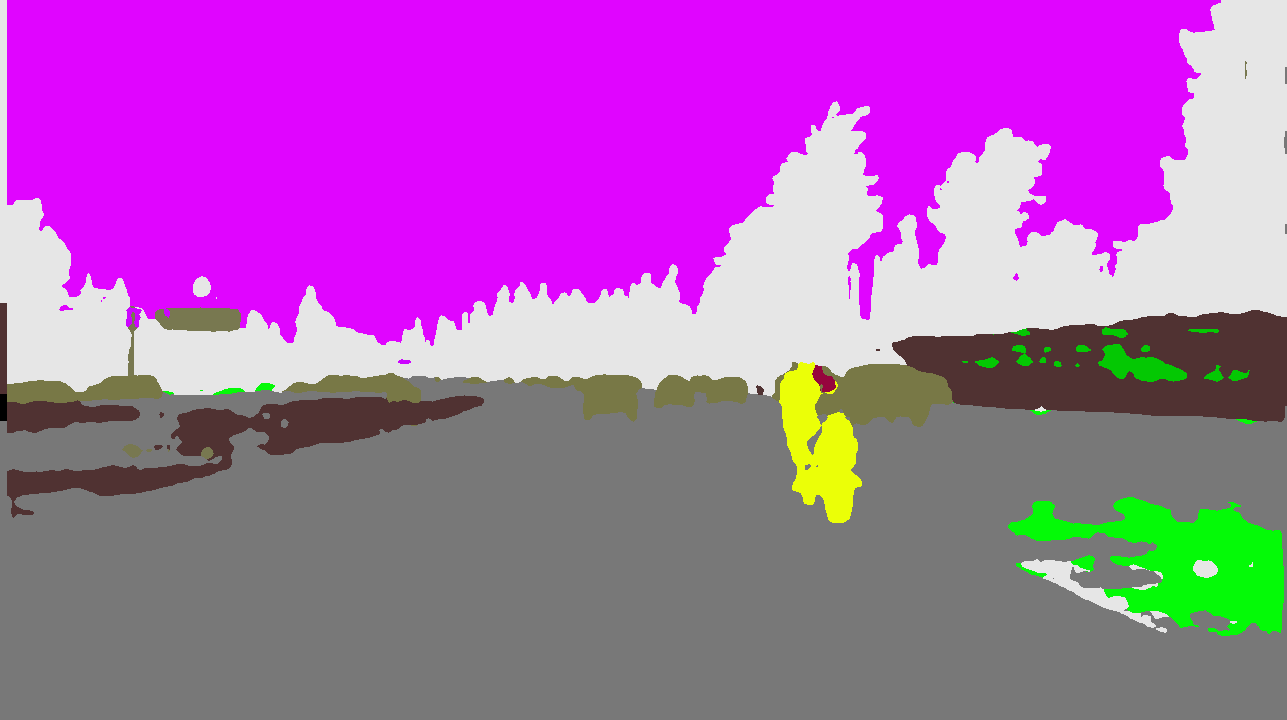}
        %\label{camvidacc}
        \end{subfigure}\;
        \begin{subfigure}[b]{0.16\linewidth}
        %\caption*{\large{\textbf{BE}}}
        \includegraphics[width=\textwidth]{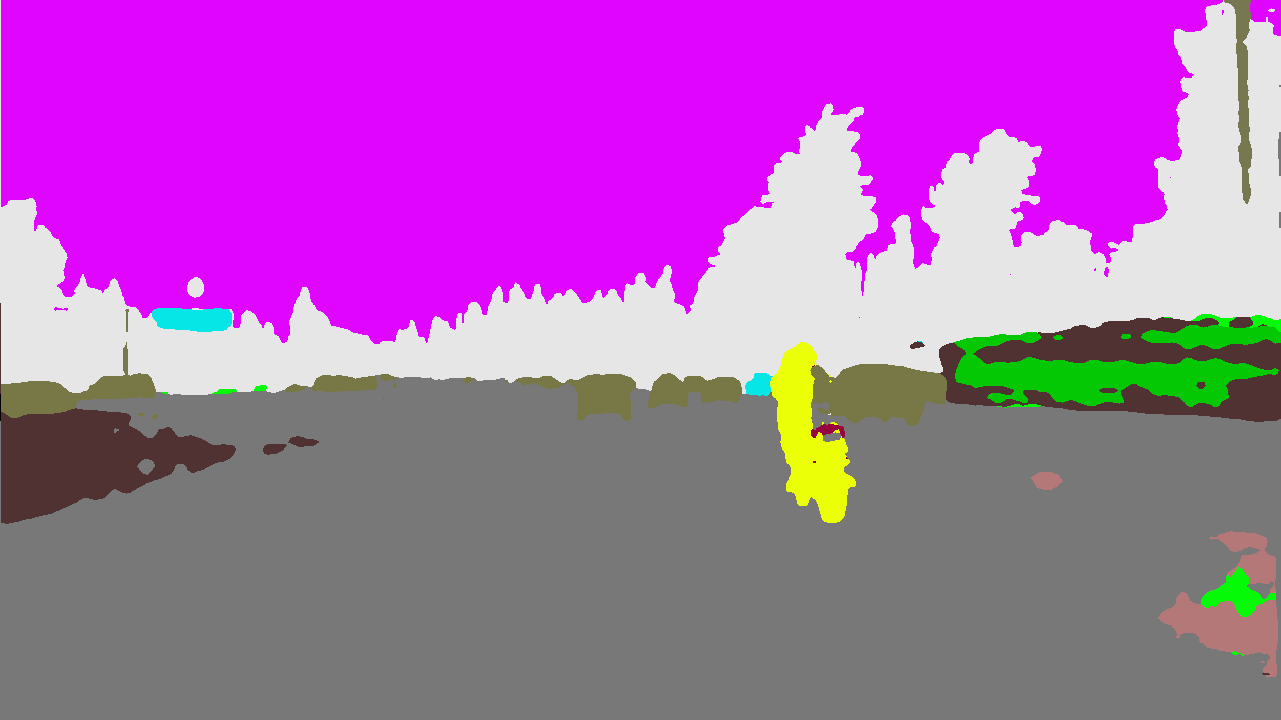}
        %\label{camvidprec}
        \end{subfigure}\;
         \begin{subfigure}[b]{0.16\linewidth}
        %\caption*{\large{\textbf{LP-BNN }}}
        \includegraphics[width=\textwidth]{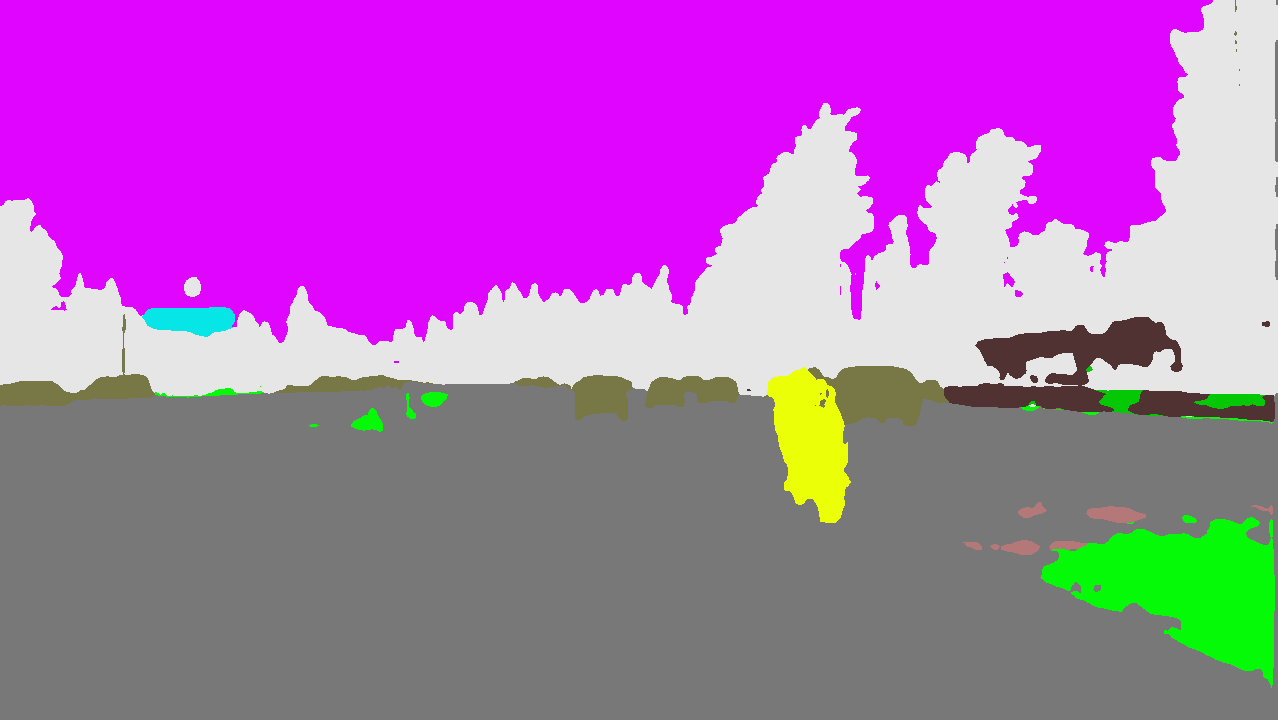}
        %\label{camvidprec}
        \end{subfigure}\;
   
        \begin{subfigure}[b]{0.16\linewidth}
        %\caption*{\large{\textbf{Input image}}}
        \includegraphics[width=\textwidth]{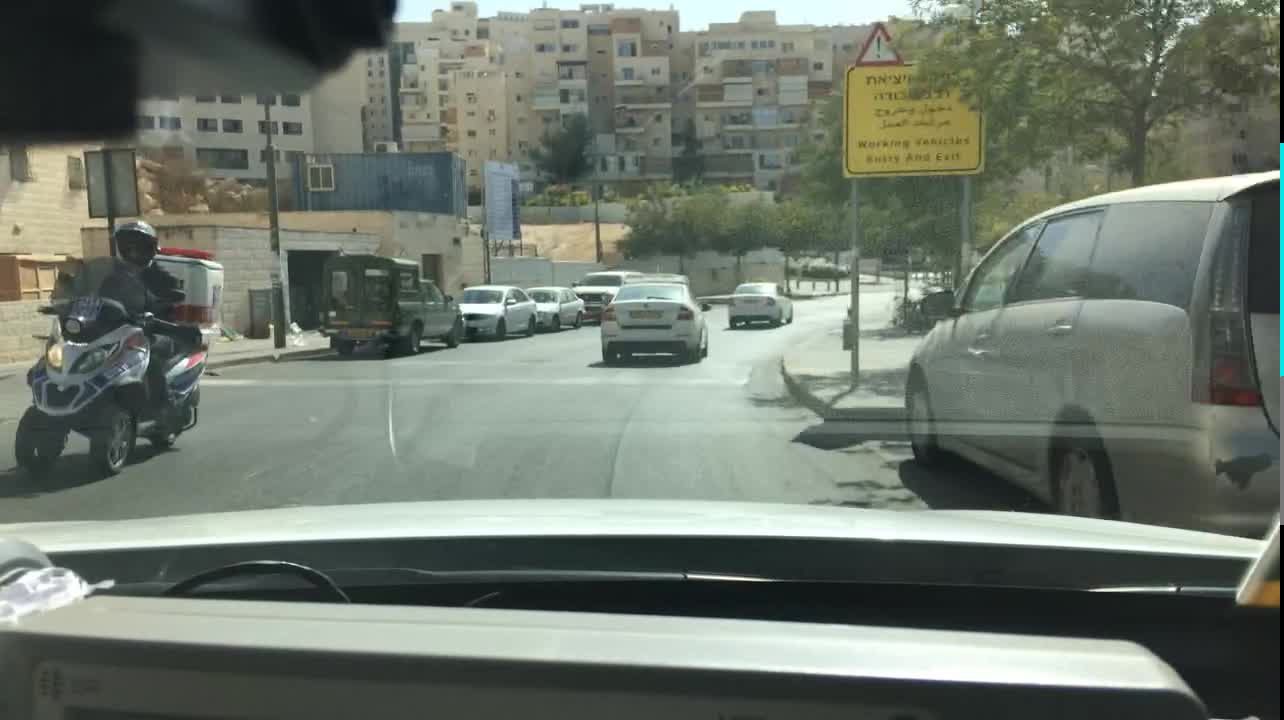}
        %\label{mnistacc}
        \end{subfigure}\;
        \begin{subfigure}[b]{0.16\linewidth}
        %\caption*{\large{\textbf{MCP}}}
        \includegraphics[width=\textwidth]{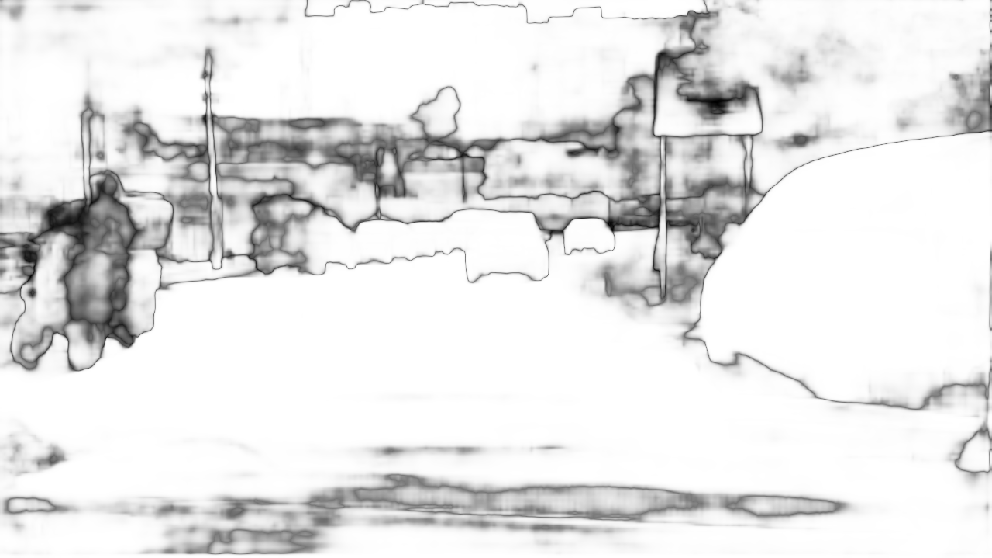}
        %\label{camvidacc}
        \end{subfigure}\;
        \begin{subfigure}[b]{0.16\linewidth}
        %\caption*{\large{\textbf{BE}}}
        \includegraphics[width=\textwidth]{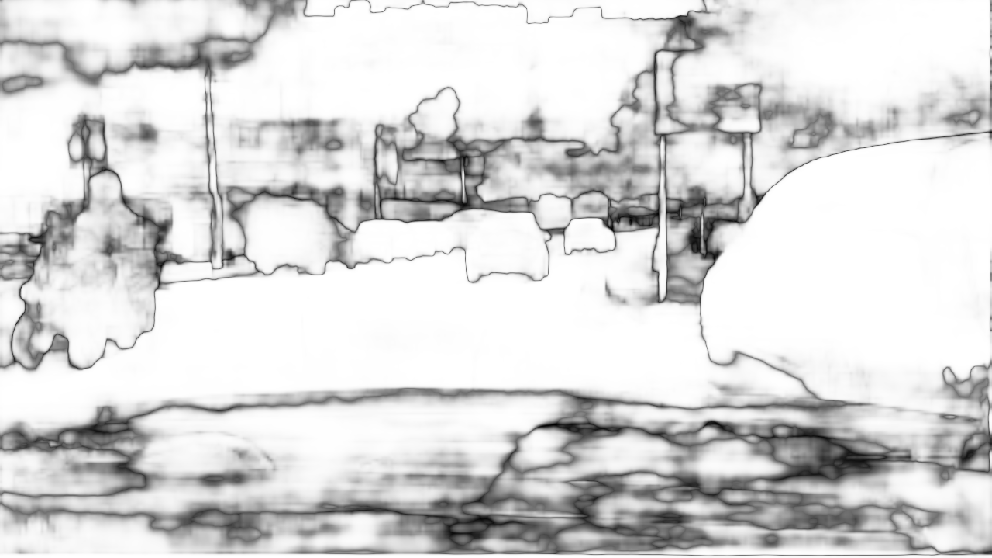}
        %\label{camvidprec}
        \end{subfigure}\;
         \begin{subfigure}[b]{0.16\linewidth}
        %\caption*{\large{\textbf{LP-BNN }}}
        \includegraphics[width=\textwidth]{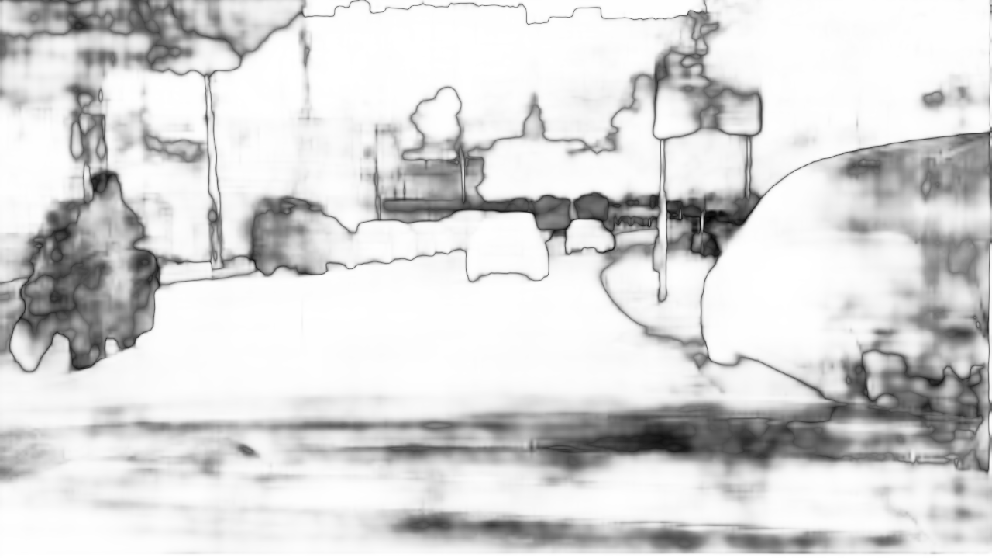}
        %\label{camvidprec}
        \end{subfigure}\;

        \begin{subfigure}[b]{0.16\linewidth}
        %\caption*{\large{\textbf{Input image}}}
        \includegraphics[width=\textwidth]{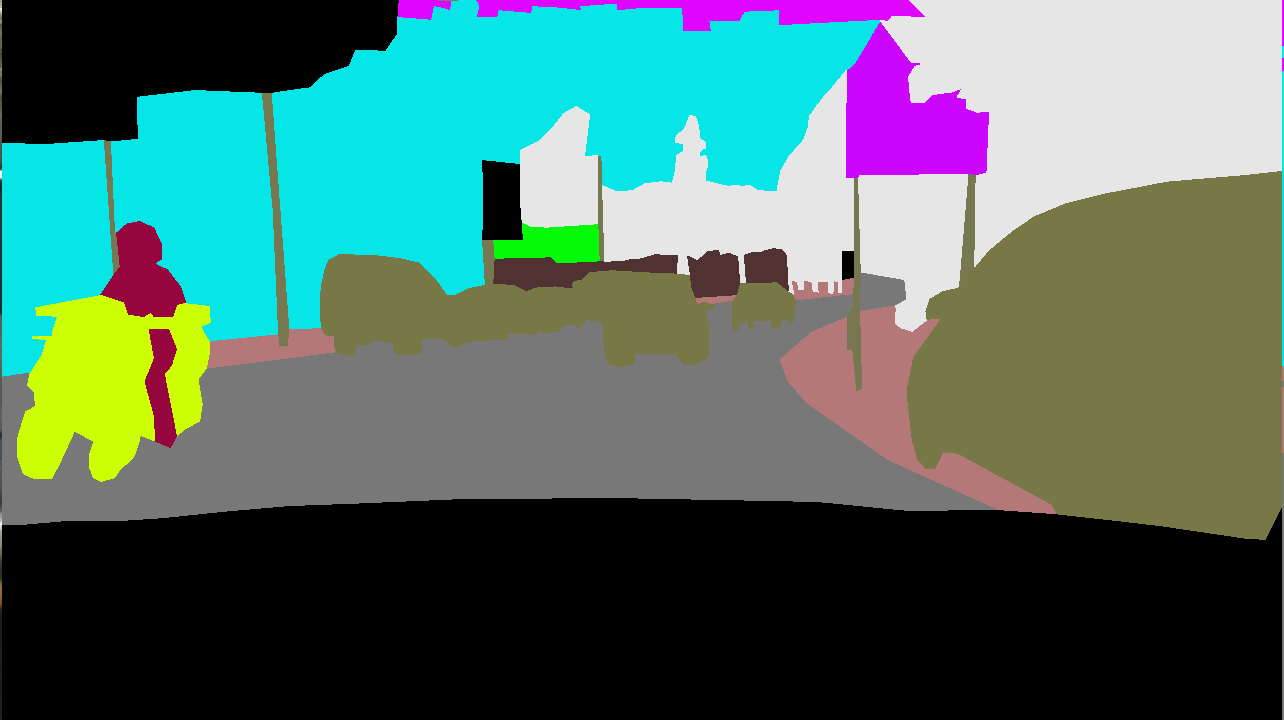}
        %\label{mnistacc}
        \end{subfigure}\;
        \begin{subfigure}[b]{0.16\linewidth}
        %\caption*{\large{\textbf{MCP}}}
        \includegraphics[width=\textwidth]{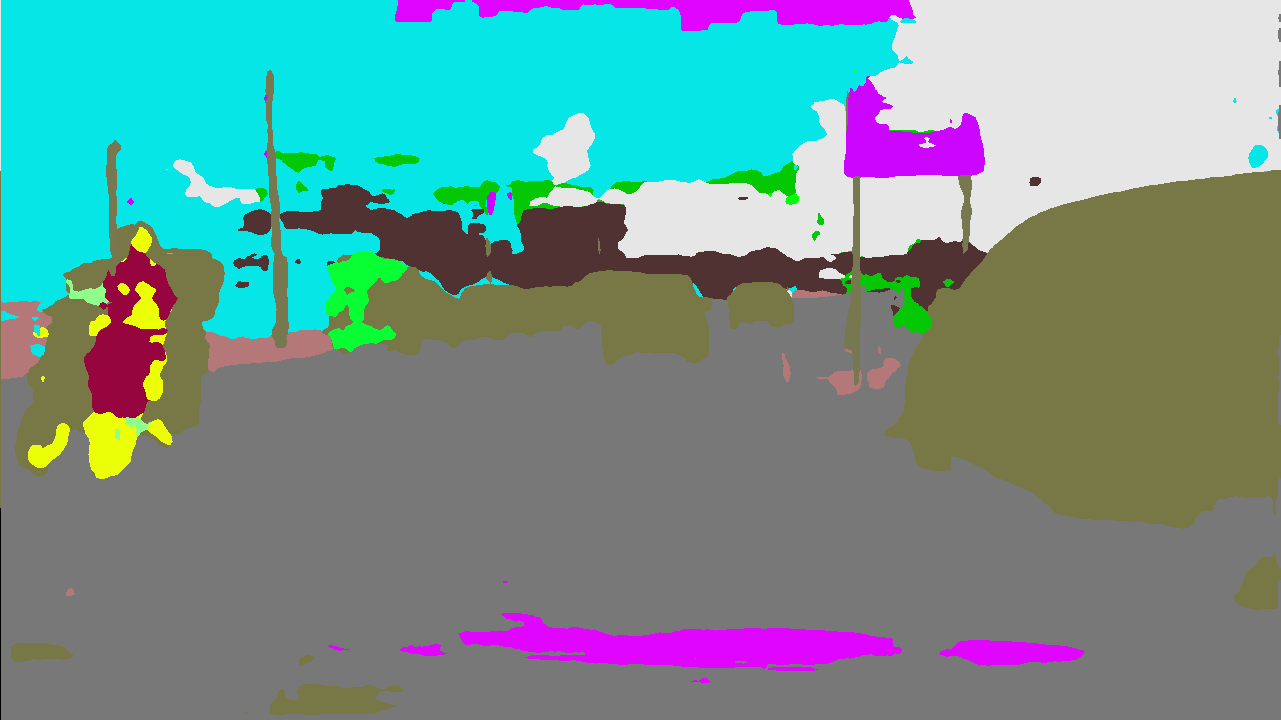}
        %\label{camvidacc}
        \end{subfigure}\;
        \begin{subfigure}[b]{0.16\linewidth}
        %\caption*{\large{\textbf{BE}}}
        \includegraphics[width=\textwidth]{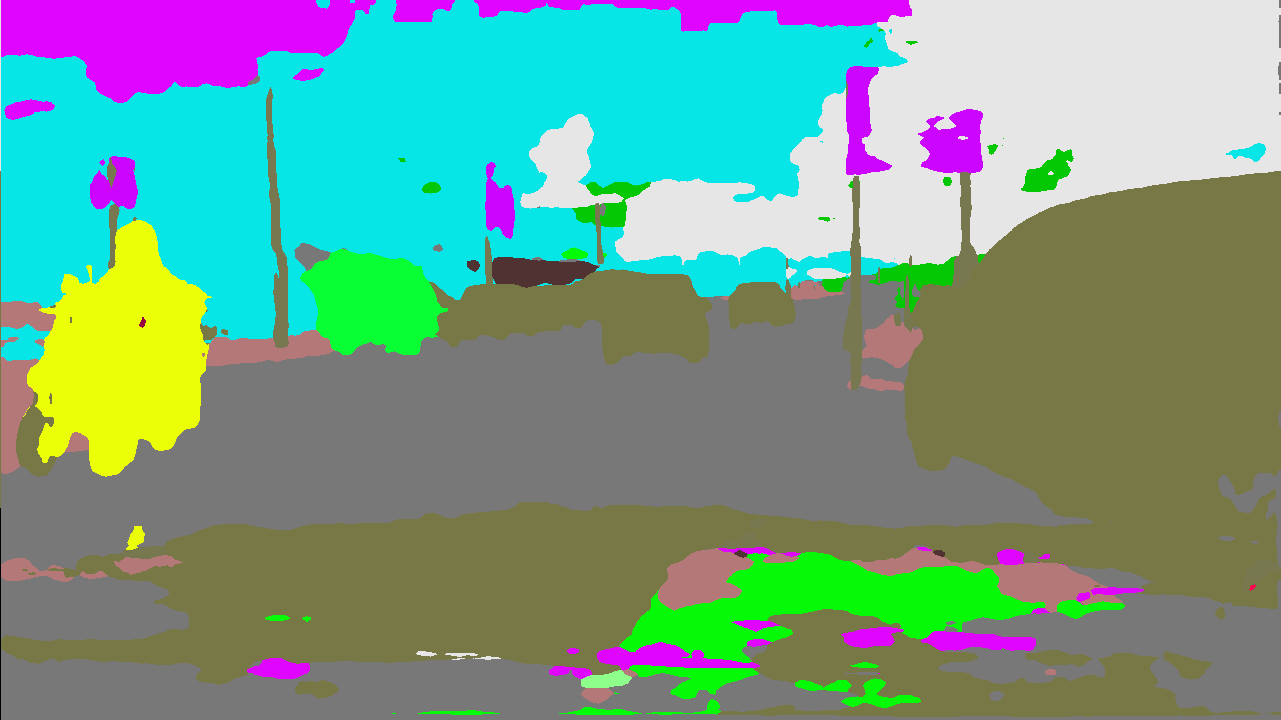}
        %\label{camvidprec}
        \end{subfigure}\;
         \begin{subfigure}[b]{0.16\linewidth}
        %\caption*{\large{\textbf{LP-BNN }}}
        \includegraphics[width=\textwidth]{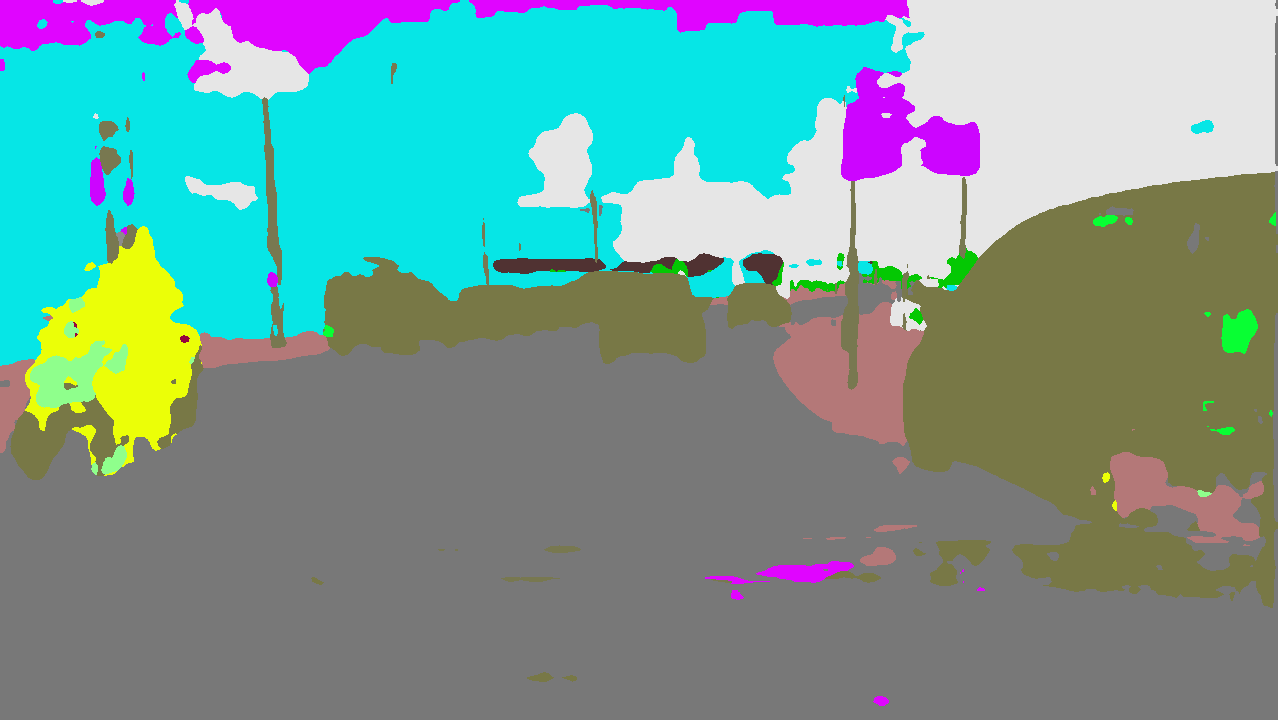}
        %\label{camvidprec}
        \end{subfigure}\;
 \caption{
%  Visual assessment on two images of BDD-Anomaly in which a motorcycle (OOD class) is present. For each image:  on the first row, input image and confidence maps for MCP, BE and LP-BNN (ours); on the second row, GT segmentation and segmentation maps for MCP, BE and LP-BNN (ours). LP-BNN is less confident on the OOD objects.
 \ab{\textbf{Visual assessment on two BDD-Anomaly test images containing a motorcycle (OOD class).} For each image: \emph{on the first row}, input image and confidence maps from MCP~\cite{hendrycks2016baseline}, BE~\cite{wen2020batchensemble}, and LP-BNN; \emph{on the second row}, ground-truth segmentation and segmentation maps from MCP, BE, and LP-BNN. LP-BNN is less confident on the OOD objects.}
 }
  \label{fig:accuconfidance}
\vspace{-5mm}
\end{figure*}

\noindent\textbf{BDD-Anomaly~\cite{hendrycks2019anomalyseg}.}
BDD-Anomaly is a subset of the BDD100K dataset~\cite{yu2020bdd100k}, composed of $6,688$ street scenes for training and $361$ for the test set. The training set contains pixel-level annotations for $17$ classes, and the test dataset is composed of the $17$ training classes and $2$ OOD classes: motor-cycle and train.
For this experiment, we use DeepLabv3+~\cite{chen2018encoder} with the experimental protocol from~\cite{hendrycks2019anomalyseg}. As previously we use ResNet50 encoder~\cite{he2016deep}. For this experiment, we use  the LP-BNN and LP-BNN + GN variants.

\noindent\textbf{Discussion.} 
We emphasize that the semantic segmentation is more challenging than the CIFAR classification since {images are bigger} and their content is more complex. % and the output space is structured. 
The larger input size constrains to use smaller \ab{mini-}batches.  This is crucial since the fast weights of the ensemble layers are trained just on one \ab{mini-}batch slice. In this experiment, we could use \ab{mini-}batches of size $4$ and train the fast weights on slices of size $1$. Yet, despite these computational difficulties, with our technique, we achieve state-of-the-art results for most metrics. We can see in Table~\ref{table:outofditribution} that our strategies {achieve} state-of-the-art performance in detecting OOD data and are well calibrated. We can also see in Figure~\ref{fig:accuconfidance}, where the OOD class is the motorcycle, that our DNN is less confident 
% in 
\ab{on} this class. Hence LP-BNN allows us to have a more reliable DNN which is essential for real-world %computer vision 
applications.

\Gianni{Table~\ref{table:time} shows the computational cost of LP-BNN \ab{and related methods}. \ab{For training,} LP-BNN
% \Isa{requires a computation time increased by a factor 1.33 only, in constrast} 
\ab{takes only $\times1.33$ more time than a vanilla approach, in contrast} to} 
% Deep Ensembles 
\ab{DE} 
% \Isa{that have a much higher computation cost}, 
\ab{that take much longer, }
% and its 
\ab{while their} performances are equivalent in most cases. 
\ab{In the same time, LP-BNN enables implicit modeling of weight correlations at every layer with limited overhead as it does not explicitly computes the covariances.}
% Also, LP-BNN allows one to model every layer covariance independently with no extra cost because it does not explicitly model the covariance. 
\ab{To the best of our knowledge, LP-BNN is the first approach with the posterior distribution computed with variational inference successfully trained and applied for semantic segmentation.}

\section{Conclusion}
We propose a new BNN framework able to quantify uncertainty in the context of deep learning. Owing to each layer of the network being tied to and regularized by a VAE, LP-BNNs are stable, efficient, and therefore easy to train compared to existing BNN models. The extensive empirical comparisons on multiple tasks show that LP-BNNs reach state-of-the-art levels with substantially lower computational cost. We hope that our work will open new research paths on effective training of BNNs. In the future we intend to explore new strategies for plugging more sophisticated VAEs in our models along with more in-depth theoretical studies.

\clearpage
{\small
\bibliographystyle{ieee_fullname}
\bibliography{egbib}
}

\clearpage

%%%%%%%%%% Merge with supplemental materials %%%%%%%%%%
\clearpage
\begin{widetext}
\begin{center}
\textbf{\large Encoding the latent posterior of Bayesian Neural Networks for uncertainty quantification (Supplementary material)}
\end{center}
\end{widetext}

\setcounter{section}{0}
\setcounter{equation}{0}
\setcounter{figure}{0}
\setcounter{table}{0}

\renewcommand{\theequation}{S\arabic{equation}}
\renewcommand{\thefigure}{S\arabic{figure}}
\renewcommand{\thetable}{S\arabic{table}}
\renewcommand{\thesection}{\Alph{section}}

\section{Discussion}

\subsection{Covariance Priors of Bayesian Neural Network}
\label{subsection1appendix}

We consider a data sample $(\vx, y)$, with $\vx \in \real^{d}$ and $y \in \real$.
We process the input data $\vx$ with a MLP network $f_{\Theta}(\cdot)$ with parameters $\Theta$ composed of one hidden layer of $h$ neurons. We detail the composing operations of the function $f_{\Theta}(\cdot)$ associated to this network: $f_{\Theta}(x) = W_2^{\top}\sigma(W_1^{\top} \vx)$, where $\sigma(\cdot)$ is an element-wise activation function. For simplicity, we ignore the biases in this example. $W_1 \in  \real^{d \times h}$ is the weight matrix associated with the first fully connected layer and $W_2 \in  \real^{h \times 1}$ the weights of the second layer. 
In BNNs, $W_1$ and $W_2$ represent random variables, 
while for classic DNNs, they are simply singular realizations of the distribution sought by BNN inference.
Most works exploiting in some way the statistics of the network parameters assume, for tractability reasons, that all weights of  $W_1$ and $W_2$ follow independent Gaussian distributions.  Hence this will lead to the following covariance matrix for $W_1$: 
% Let us consider a simple NN %\emi{(maybe say just "a simple network" because if it has a couple of layers you can not state it is deep)}
% composed of one hidden layer with $N$ neurons. For the sake of simplicity, we assume that the input is a vector  $x$ of size $D$ and  the output layer consists in a scalar $y$. %\emi{(  (Emi: to be more clear about why you say one layer in the beginning and then you talk about two) \Gianni{1 hidden layer = 2 layers}}.
% Let us write $f_{\Theta}(\cdot)$ the function associated with such a DNN  with parameters $\Theta$.
%  Let us write $W_1 \in  \real^{D \times N}$  the weight matrix associated with the first fully connected layer and $W_2 \in \real^{N \times 1}$ the one with the second one.% \emi{above you said one layer)}.
%  Hence $f_{\Theta}(x) = a(x \times W_1) \times W_2$ in which $a(\cdot)$ represents an activation function that is applied element wise.
% %\emir{For classical DNN, $W_1$ and $W_2$  are just matrices of one realization, while for BNN, these matrices represent random variables.} 
% \emi{For BNN, these matrices represent random variables, whilst for classical DNN, $W_1$ and $W_2$  are simply singular realizations of the distribution sought by BNN inference.} %\emir{In most cases, we}
% \emi{Most works exploiting in some way the statistics of the network parameters} assume \emi{for tractability reasons} that every weight of  $W_1$ and $W_2$ follow\emi{s} independent Gaussian distribution\emi{s}.  Hence this will lead to the following covariance matrix for $W_1$:

$$\begin{pmatrix}
\mbox{var}(W_1[1,1]) & 0 & 0& \ldots & 0\\
0 & \mbox{var}(W_1[2,1]) & 0& \ldots & 0\\
\vdots &  \ddots& \ddots& \ddots & \vdots\\
0 & 0 & 0& \ldots & \mbox{var}(W_1[\ab{d,h}])\\
\end{pmatrix}$$
where $\mbox{var}(W_1[i,j])$ is the variance of the coefficient 
% $i,j$
\ab{$[i,j]$} of matrix $W_1$. Similarly, for $W_2$ we will have a diagonal matrix.

Now, let us assume that $W_1$ and   $W_2$ have a latent representation 
\ab{$Z_1 {=} \big[ Z_1[1], Z_1[2], Z_1[3] \big] \in \real^{3}$}
and  
\ab{$Z_2 {=} \big[ Z_2[1], Z_2[2], Z_3[3] \big] \in \real^{3}$}, respectively, such that  for every coefficient 
% $i,j$ 
\ab{$[i,j]$}  of $W_1$ and $W_2$ there exist real weights 
\ab{$\{ \alpha^{[i,j]}_1[k]\}_{k=1}^3$ and $\{ \alpha^{[i,j]}_2[k]\}_{k=1}^3$ such that: $W_1[i,j] {=} \sum_{k=1}^3 \alpha_1^{[i,j]}[k]Z_1[k]  $ and $ W_2 {=} \sum_{k=1}^3 \alpha^{[i,j]}_2[k]Z_2[k]$, respectively.}
%\abc{I've replaced $l$ with $k$: better clarity and more standard notations.} -> ok
%\emi{In the case of} LP-BNN, we consider that $Z_1$  and $Z_2$ represent independent random variables. 
\Gianni{In the case of LP-BNN, we consider that each coefficient of $Z_1$  and $Z_2$ represent an independent random variable.} 
% Hence contrary to others,
\ab{Thus, in contrast to approaches based on the mean-field approximation  \Gianni{directly on the weights of the DNN},} we can have for each layer a non-diagonal covariance matrix with the following variance and covariance term\emi{s} for $W_1$:
% $$ \mbox{var}(W_1[i,j]) = \sum_{l}\alpha_1[l]^2\mbox{var}(Z_1[l])$$
% $$ \mbox{cov}(W_1[i,j],W_1[i',j']) = \sum_{i,j} \alpha_1[l]\alpha_1'[l] \mbox{var}(Z_1)[l]$$
% with $W_1[i',j']\sum_l \alpha_1'[l]Z_1[l]$.

\begin{equation}
\mbox{var}(W_1[i,j]) = \sum_{k=1}^3(\alpha^{[i,j]}_1[k])^2\mbox{var}(Z_1[k]) 
\end{equation}
\begin{equation}
 \mbox{cov}(W_1[i,j],W_1[i',j']) = \sum_{k=1}^3 \alpha^{[i,j]}_1[k]\alpha^{[i',j']}_1[k] \mbox{var}(Z_1[k]) 
\end{equation}
% \abc{Shouldn't $\mbox{var}(Z_1[k])$ be in fact $\mbox{var}(Z_1)$?}
%\Isa{ and the sum should be over k instead of i,j (which does not make sense since i,j are fixed)?} \ab{with $W_1[i',j']{=}\sum_{k=1}^3 \alpha_1'[k]Z_1[k]$.} -> corrected
This allows us to leverage the lower-dimensional parameters of the distributions of $Z_1$ and $Z_2$ for estimating the higher-dimensional distributions of $W_1$ and $W_2$. In this manner, in LP-BNN we model an \textbf{implicit covariance} of weights at each layer.

We note that several approaches for modeling correlation between weights have been proposed under certain settings and assumptions. For instance, Karaletesos and Bui~\cite{karaletsos2020hierarchical} model correlations between weights within a layer and across layers thanks to a Gaussian process-based approach working in the function space via hierarchical priors instead of directly on the weights. Albeit elegant, this approach is still limited to relatively shallow MLPs (e.g., one hidden layer with 100 units~\cite{karaletsos2020hierarchical}) and cannot scale up yet to deep architectures considered in this work (e.g., ResNet-50).
% Sun et al.~\cite{sun2017learning} 
Other approaches~\cite{louizos2016structured, sun2017learning} 
model layer-level weight correlations through Matrix Variate Gaussian (MVG) prior distributions, increasing the expressiveness of the inferred posterior distribution at the cost of further increasing the computational complexity w.r.t. mean-field approximated BNNs~\cite{blundell2015weight, graves2011practical}.
In contrast, LP-BNN does not \textbf{explicitly model the covariance} by conveniently leveraging fully connected layers to project weights in a low-dimensional latent space and performing the inference of the posterior distribution there. This strategy leads to a lighter BNN that is competitive in terms of computation and performance for complex computer vision tasks.

% \emi{Karaltesos and Bui~}\cite{karaletsos2020hierarchical} model correlations between weights within a layer and across layers thanks to a Gaussian Process-based approach working in the function space instead of directly on the weights. 
% However they cannot scale up from the largest MLP they evaluated (one hidden layer with 100 units) to the deep architectures we consider here (ResNet-50). 
% \emi{Sun \etal~}\cite{sun2017learning} model weight correlations through Matrix Variate Gaussian prior distributions, increasing the computational complexity w.r.t. \cite{blundell2015weight}.
%  In contrast, LP-BNN does not \textbf{explicitly model the covariance}. This strategy allows us to have a lighter BNN that can be used for computer vision tasks.
 
 %\emi{Emi remarks : y is mentioned only once and is not related to the other variables} \Sev{Sev: en fait dans le tableau des notations, $y_i$ est present}
 
\subsection{The utility of Rank-1 perturbations }
\label{subsection2appendix}

\ab{One could ask why using the Rank-1 perturbation formalism from BE~\cite{wen2020batchensemble}, instead of simply feeding the weights of a layer to the VAE to infer the latent distribution. Rank-1 perturbations significantly reduce the number of weights upon which we train the VAE, due to the decomposition of the fast weights into $\vr$ and $\s$. This further allows us to consider multiple such weights, $J$, at each forward pass enabling faster training of the VAE as its training samples are more numerous and more diverse.} 

Next, we establish connections between the cardinality $J$ of the ensemble and the posterior covariance matrix.
% Our prior distribution allows for the introduction of correlations between weights, which is a desirable property due to its superior expressiveness~\cite{dusenberry2020efficient} but which can be otherwise difficult to approximate. 
\ab{The mechanism of placing a prior distribution over the latent space enables an implicit modeling of correlations between weights in their original space. This is a desirable property due to its superior expressiveness~\cite{foong2020expressiveness, karaletsos2020hierarchical} but which can be otherwise computationally intractable or difficult to approximate.} 
The covariance matrix of our prior \ab{in the original weight space} is a Rank-1 matrix. Thanks to the Eckart-Young theorem (Theorem 5.1 in~\cite{wang2012geometric}), we can quantify the error of approximating the covariance by a Rank-1 matrix, based on the second up to the last singular values.

Let us denote by $\Theta_1, \ldots,\Theta_J$ the $J$ weights trained by our algorithm,   $\Theta_{\text{avg}}=\frac{1}{J}\sum_{j=1}^{J}\Theta_j$ and $\Delta_j =\Theta_j - \Theta_{\text{avg}}$. The differences and the sum in the previous equations are calculated element-wise on %\Isa{for ? je ne suis pas sure de bien comprendre...} 
all the weights of the DNNs. 
Then, for each new data sample {$\vx$}, the prediction of the DNN $f_{\Theta_{\text{avg}}}(\cdot)$  is equivalent to the average of the DNNs $f_{\Theta_j}(\cdot)$  applied on  $\vx$ :
\begin{equation}
\frac{1}{J}\sum_{j=1}^{J}  f_{\Theta_j}(\vx) = f_{\Theta_{\text{avg}}}(\vx)+\mathcal{O}\left( \| \Delta \|^2 \right) 
\end{equation}
with $ \| \Delta \|  = \max_j  \| \Delta_j \|$.
% where the $L_2$ is computed over all weights.
\ab{The $L_2$ norm is computed over all weights.}
% The proof can be found in Section 3.5 of \cite{izmailov2018averaging}.
\ab{We refer the reader to the proof in \S3.5 of~\cite{izmailov2018averaging}.}
It follows that in fact we do not learn a Rank-1 matrix, but an up to Rank-$J$ covariance matrix, if all the $\s_j$ $\vr_j$ are independent. 
Hence the choice of $J$ acts as an approximation factor of the covariance matrix. 
\ab{Wen \etal\cite{wen2020batchensemble} tested 
different values of $J$ and found} that $J=4$ was the best compromise, which we also use \ab{here}.

\subsection{Computational complexity}
% \subsection{\ab{Memory costs of BNNs}}
\label{subsection3appendix}

Recent 
\ab{works}~\cite{fort2019deep,wilson2020bayesian} 
studied the weight modes 
computed by Deep Ensembles \ab{under a BNN lens}, yet these \ab{approaches are} computationally \ab{prohibitive} at the scale required for \ab{practical} computer vision tasks. % Dusenberry \etal~\cite{dusenberry2020efficient} propose a \ab{more} scalable solution for image classification, which is nonetheless prone to high instabilities due to the important number of parameters and to the fact that $\vr_j$ and $\s_j$ are the latent variables \abc{Latent variable is potentially very confusing here} \Isa{and the message is not clear...} of the variational distribution.
\ab{Recently, Dusenberry et al.~\cite{dusenberry2020efficient} proposed a more scalable approach for BNNs, which can still be subject to high instabilities as the ELBO loss is applied over a high-dimensional parameter space, all BatchEnsemble parameters. Increased stability can be achieved by leveraging large mini-batches that bring more robust feature and gradient statistics, at significantly higher computational cost (large virtual mini-batches are obtained through distributed training over multiple TPUs).}
% \Gianni{Dusenberry et al.~\cite{dusenberry2020efficient} propose a more scalable solution for image classification, which is nonetheless prone to high instabilities due to the important number of parameters and the fact that they do not rely on extra latent variables and directly use the ELBO loss on all the BatchEnsemble parameters.  } %\Sev{it seems to be clearer to me}
In comparison, our 
approach 
% requires less memory resources 
\ab{has a smaller memory overhead} since we encode $\vr_j$ in a lower dimensional space (we found empirically that a latent space of size {only} $32$ provides {an appealing} compromise between accuracy and compactness). \ab{The ELBO loss here is applied over this lower-dimensional space which is easier to optimize.} The only additional cost in terms of parameters and memory used 
\ab{w.r.t.} BE is related to the compact VAEs associated with each layer. 

% Besides 
\ab{In addition to} the lower number of parameters, LP-BNN training is more stable than for Rank-1 BNN~\cite{dusenberry2020efficient} due to the reconstruction term $\|\vr_j -\hat{\vr}_j \|^2_2$  which regularizes the $\mathcal{L}_{\scriptscriptstyle \text{LP-BNN}}$
loss in Eq.~(7) \Isa{of the main paper} %~\eqref{eq:loss-lpbnn}
by controlling the variances of the sampled weights.  
% BNNs usually need a range of  heuristics
\ab{In practice, to train BNNs successfully, a range of carefully crafted heuristics are necessary}, e.g., clipping, initialization from truncated Normal \Isa{distributions}, extra weight regularization to stabilize training~\cite{dusenberry2020efficient}. For \method, training is overall straightforward even on complex and deep models, e.g., DeepLabV3+, thanks to the VAE module that is stable \ab{and trains faster}.

\begin{table}[t!]
\renewcommand{\figurename}{Table}
\renewcommand{\captionfont}{\small}
\begin{center}
\scalebox{0.8}
{
\begin{tabular}{l|c|c|c|c|c}
\toprule
\textbf{Learning Rate }                                              & 0.2   & 0.1   & 0.05  & 0.01  & 0.005 \\
\midrule
\begin{tabular}[c]{@{}l@{}}\textbf{BNN }\\ accuracy\end{tabular}         &22.48 & 44.60 & 49.83 & 48.70 & 56.69 \\
\midrule
\begin{tabular}[c]{@{}l@{}}\textbf{BNN }\\ epoch div\end{tabular}       &3     & 25    & 65    & None  & None  \\
 \midrule
\begin{tabular}[c]{@{}l@{}}\textbf{LP-BNN }\\ accuracy\end{tabular}        & 20.02 & 55.04 & 59.68 & 63.73 & 64.41 \\
 \midrule
\begin{tabular}[c]{@{}l@{}}\textbf{LP-BNN }\\ epoch div\end{tabular}      &3     & None  & None  & None  & None \\ 

\bottomrule
\end{tabular}
}
\end{center}
\caption{\Gianni{\textbf{Stability analysis of BNNs.}} Stability experiment with LeNet 5 architecture and 80 epoch\ab{s} on CIFAR-10. 
% Please note that 
\ab{On the epoch divergence row, \emph{None}} means that the DNN does not diverge. }\label{table:stability}
\vspace{-3pt}
\end{table}

\subsection{Stability of Bayesian Neural Networks}
In this section, we experiment on CIFAR-10 to evaluate the stability of LP-BNN versus a 
% classical
\ab{classic} BNN. \emi{For} this experiment, we use \emi{the} LeNet-5 architecture and choose a weight decay of $1e-4$  \emi{along with} a \ab{mini-}batch size of $128$. Our goal is to see \emi{whether} both techniques are stable when we vary the learning rate.  Both DNNs were trained under the \Gianni{exact} same conditions for $80$ epochs.
In Table~\ref{table:stability}, we present two metrics for both DNNs. The first metric is the accuracy.  \Gianni{The second metric is the epoch \emi{during which} the training loss of the DNN \ab{explodes, i.e.,} is equal to infinity. This phenomenon may occur if the DNN is highly unstable to train. 
% BNN happens to be highly unstable.
}
\ab{We argue that LP-BNN is visibly more stable and standard BNNs during training.}
We can see from Table~\ref{table:stability} that LP-BNN is more stable than the 
% classical 
\ab{standard} BNN \emi{as it does not diverge for a wide range of learning rates}. \emi{Moreover, its accuracy is higher than that of a standard BNN implemented on the same architecture, a property that we attribute to the VAE regularization.}

\subsection{LP-BNN diversity}

\begin{table}[t!]
\renewcommand{\figurename}{Table}
\renewcommand{\captionfont}{\small}
\begin{center}
\scalebox{0.75}
{
\begin{tabular}{l|c|c|c}
\toprule
 &   \textbf{ratio errors $\uparrow$ }  & \textbf{Q-statistic $\downarrow$ }   & \textbf{correlation coefficient $\downarrow$ } \\ 
\midrule
\textbf{DE}        &\textbf{0.9825 }     & \textbf{0.9877 }   & \textbf{0.6583} \\ 
\midrule
\textbf{BE}         &0.5915     & 0.9946     & 0.7634  \\ 
 \midrule
\textbf{LP-BNN }      & 0.8390     & \textbf{0.9842}    & \textbf{0.6601 }      \\ 
\bottomrule
\end{tabular}
}
\end{center}
\caption{\Gianni{\textbf{Comparative results of diversity \ab{scores} for image classification 
% tasks
\ab{on the CIFAR-10 dataset.}
}}
% Quantitative evaluation of the diversity on CIFAR-10  dataset. 
}
\label{table:diversity}
\vspace{-3pt}
\end{table}

\begin{table}[t!]
\renewcommand{\figurename}{Table}
\renewcommand{\captionfont}{\small}
\begin{center}
\scalebox{0.75}
{
\begin{tabular}{l|c|c|c}
\toprule
 &   \textbf{ratio errors $\uparrow$ }  & \textbf{Q-statistic $\downarrow$ }   & \textbf{correlation coefficient $\downarrow$ } \\  
\midrule
\textbf{DE}        &0.4193      & 0.9690    & 0.7568 \\ 
\midrule
\textbf{BE}         &0.2722     & 0.9874    & 0.8352  \\ 
 \midrule
\textbf{LP-BNN }      & \textbf{0.4476 }    & \textbf{0.9595}    & \textbf{0.7332}       \\ 
\bottomrule
\end{tabular}
}
\end{center}
\caption{\Gianni{\textbf{Comparative results of diversity \ab{scores} for image classification 
% tasks 
\ab{on the CIFAR-10-C dataset}.}} 
% Quantitative evaluation of the diversity on CIFAR-10-C dataset.
% \ab{The evaluation is conducted on the CIFAR-10-C dataset.}
}\label{table:diversity_c}
\vspace{-3pt}
\end{table}

\ab{At test-time, }
\Gianni{
% During inference, 
BNNs and DE aggregate the different predictions. For BNNs, these predictions come from the different  realizations of the posterior distribution, while, for the DE, these predictions \emi{are provided by} several DNNs trained in parallel.
As proved in \cite{fort2019deep,rame2021dice}, the diversity \emi{among} these different predictions is 
% the
key to quantify the uncertainty of a DNN. Indeed, we want the different DNN predictions to have a high variance when the 
% DNN 
\ab{model} is not accurate. 
\emi{Figure~\ref{fig:diversityall} highlights this desirable property of high variance on out-of-distribution samples exhibited by LP-BNN.} Also, as in~\cite{rame2021dice}, we evaluate the ratio-error introduced in~\cite{aksela2003comparison}. The ratio-error between two classifiers is the  number of data \ab{samples} on which 
% just 
\ab{only} one classifier is wrong divided by the number of 
% data 
\ab{samples} on \emi{which} they are both wrong. A higher value means that the two classifiers are less likely to make the same errors. 
We also evaluated the Q-statistics \cite{aksela2003comparison}, which measures the diversity between two classifiers. The value of the Q-statistics \emi{is} between \ab{$-1$ and $1$} and is defined 
% by:
\ab{as:}
\begin{equation}
    Q = \frac{N_{11}N_{00} - N_{10}N_{01}}{N_{11}N_{00} + N_{10}N_{01}}
\end{equation}
where $N_{11}$ and $N_{00}$ are the numbers of data \emi{on which} both classifiers are correct and incorrect\emi{, respectively}.  $N_{10}$ and  $N_{01}$ are the number of data where just one of the two classifier\emi{s} is wrong. 
If the two classifiers are always wrong or right for all data\emi{, then} $N_{10}{=}N_{01}{=}0$ and $Q{=}1$, \emi{while} if both classifiers always make errors on different inputs, then $Q{=}-1$. The maximum diversity comes when $Q$ is minimum.} 

\Gianni{Finally, we evaluated the correlation coefficient~\cite{aksela2003comparison}, which assesses %the correlation of each classifier vector errors 
\Isa{the correlation between the error vectors of the two classifiers}. %The error vector is a binary vector whose coefficient equals one if the data corresponding to this coefficient is a wrong prediction and zeros everywhere else.
Tables \ref{table:diversity} and \ref{table:diversity_c} illustrate that, for the normal case (CIFAR-10), LP-BNN 
% brings the same quantity of diversity as DE,
\ab{displays similar diversity with DE,} while in the corrupted case (CIFAR-10-C) LP-BNN 
% brings more diversity
\ab{achieves better diversity scores.} 
% Hence 
\emi{We conclude that} 
% with respect to 
\ab{in terms of diversity metrics,} LP-BNN has \emi{indeed} the behavior that one \emi{would} expect \emi{for uncertainty quantification purposes}.
}

\label{subsection5appendix}
\begin{figure*}[!t]
\renewcommand{\captionfont}{\small}

    \centering
    \scalebox{0.85}
{
\begin{tabular}{cccccccc}
\raisebox{7mm}{\large{Input}}
& 
\includegraphics[width=0.10\linewidth]{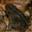}\label{image1} &
\includegraphics[width=0.10\linewidth]{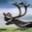}\label{image2} &
\includegraphics[width=0.10\linewidth]{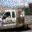}\label{image4} &
\includegraphics[width=0.10\linewidth]{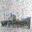}\label{image4} &
\includegraphics[width=0.10\linewidth]{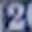}\label{image4} &
\includegraphics[width=0.10\linewidth]{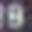}\label{image4} &
 \\

\raisebox{7mm}{\large{LP-BNN}} &
 \includegraphics[width=0.15\linewidth]{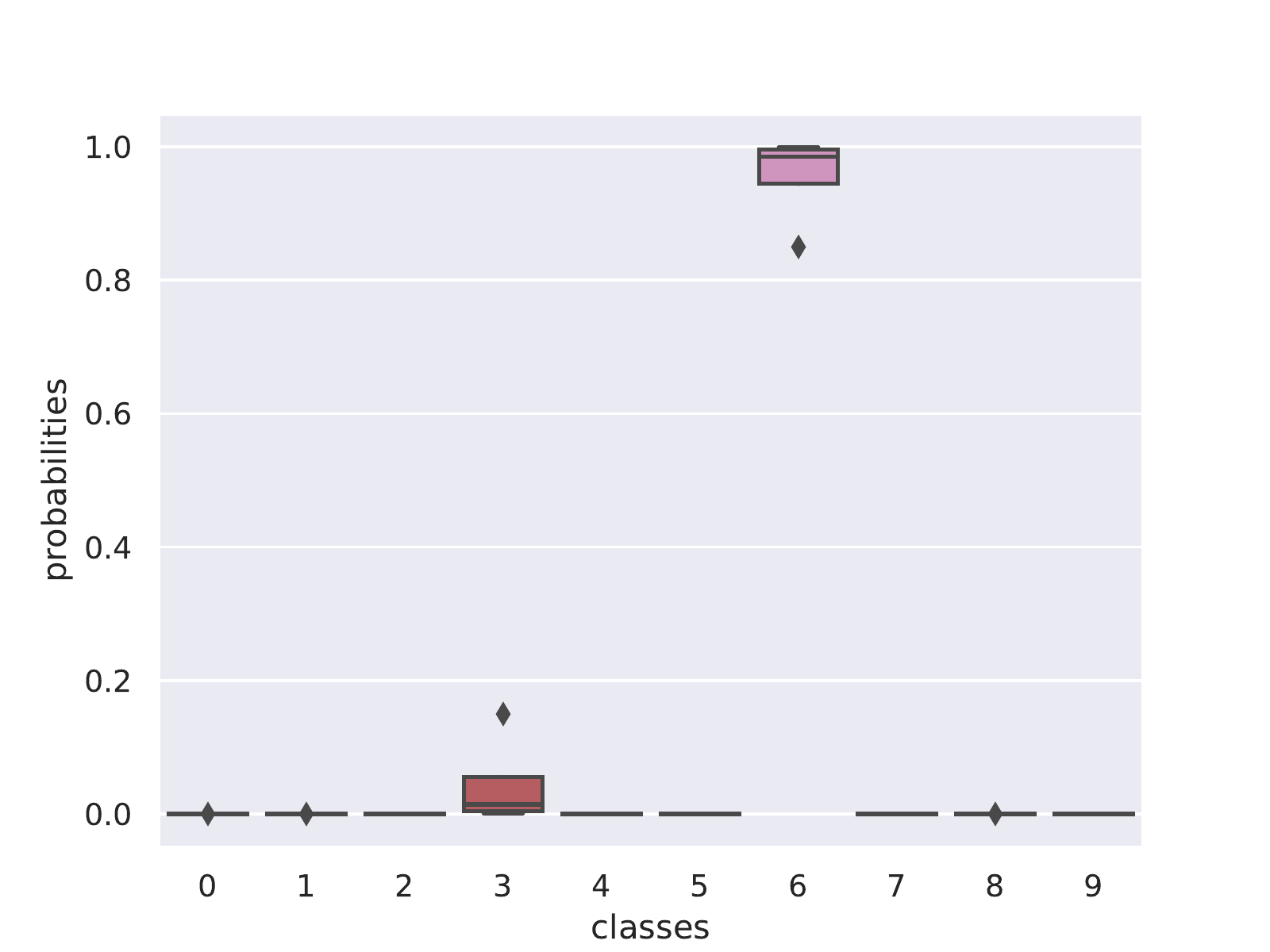}\label{dist1} &
 \includegraphics[width=0.15\linewidth]{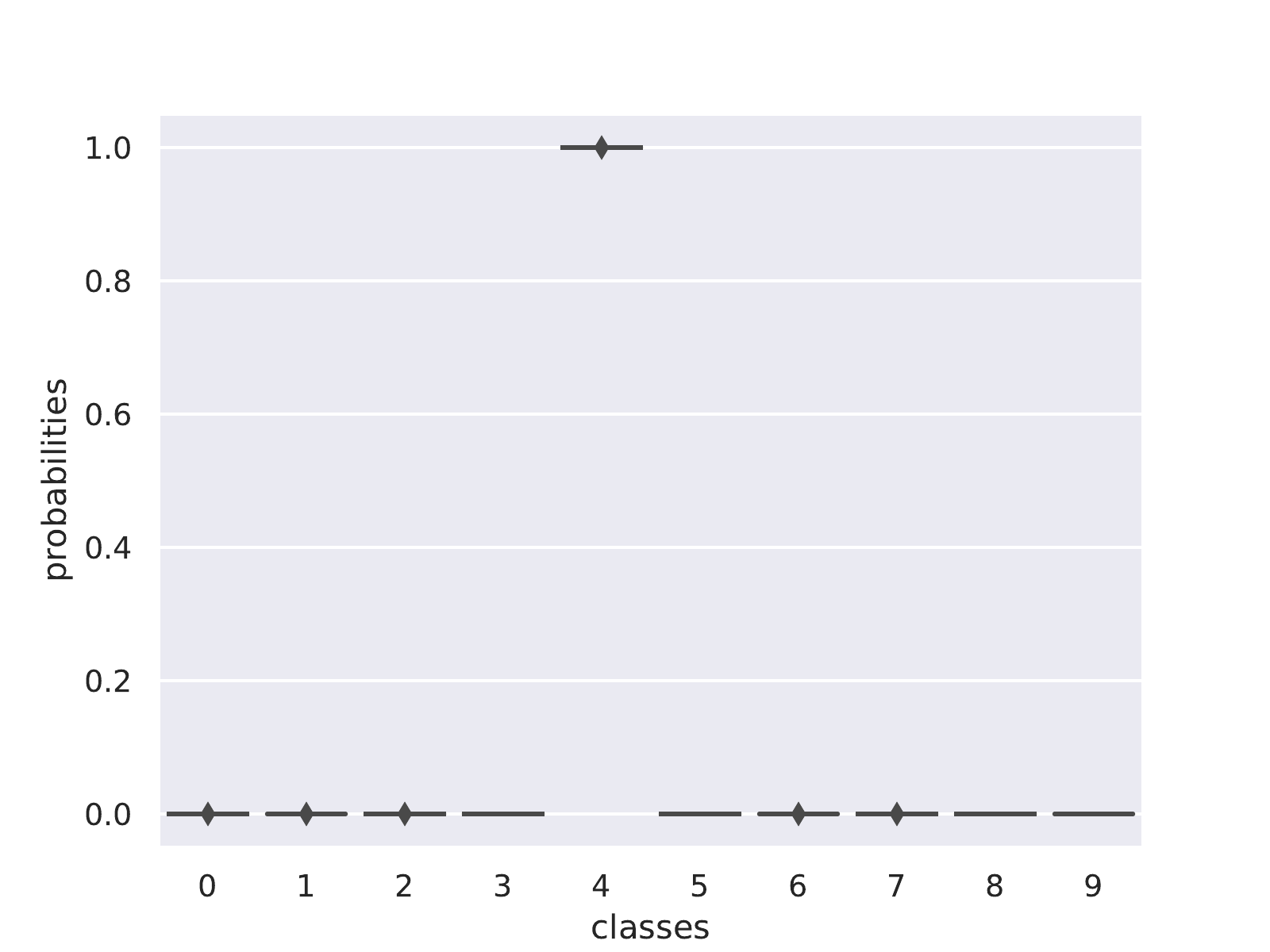}\label{dist2} &
 \includegraphics[width=0.15\linewidth]{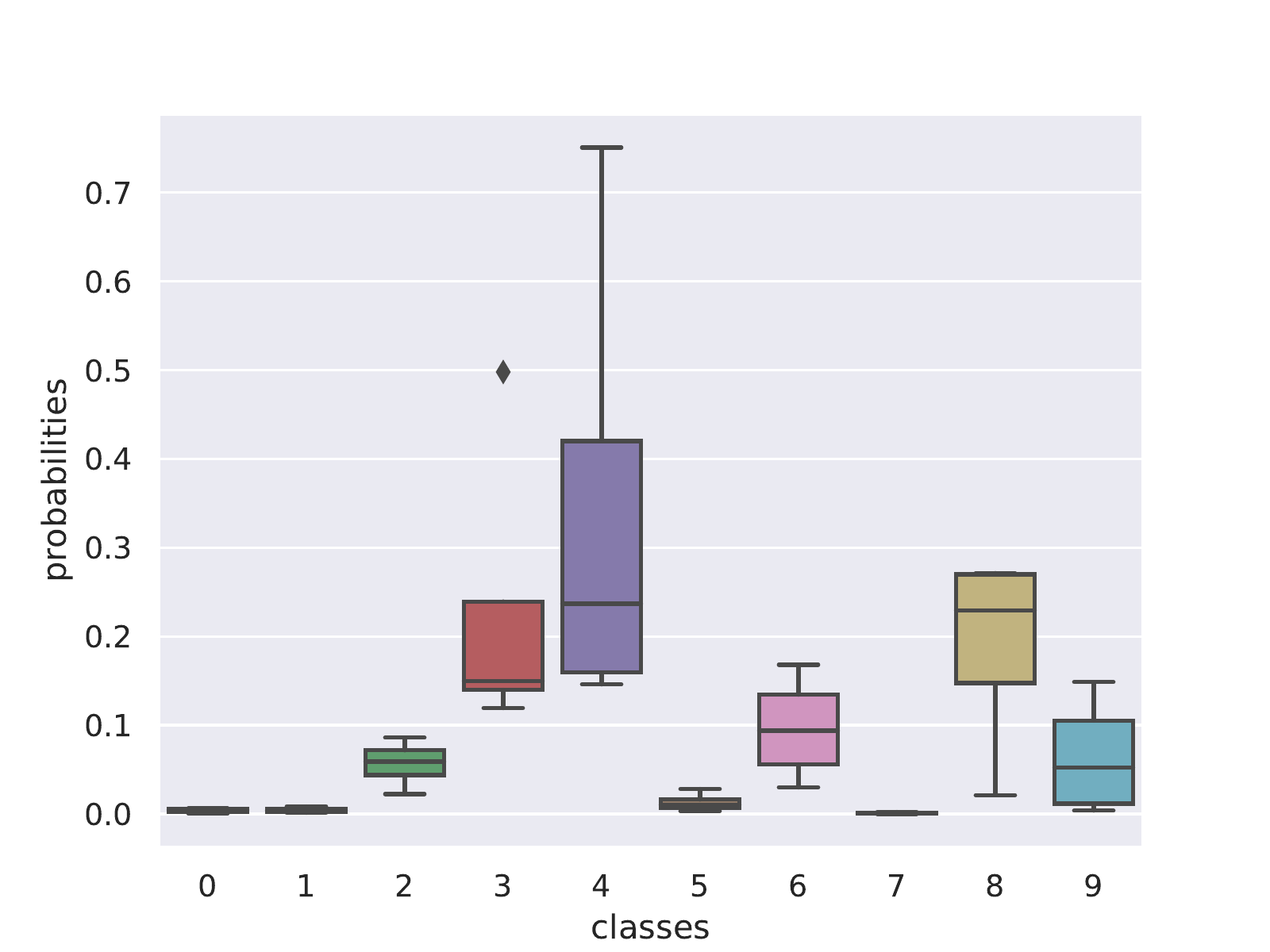}\label{dist4} &
 \includegraphics[width=0.15\linewidth]{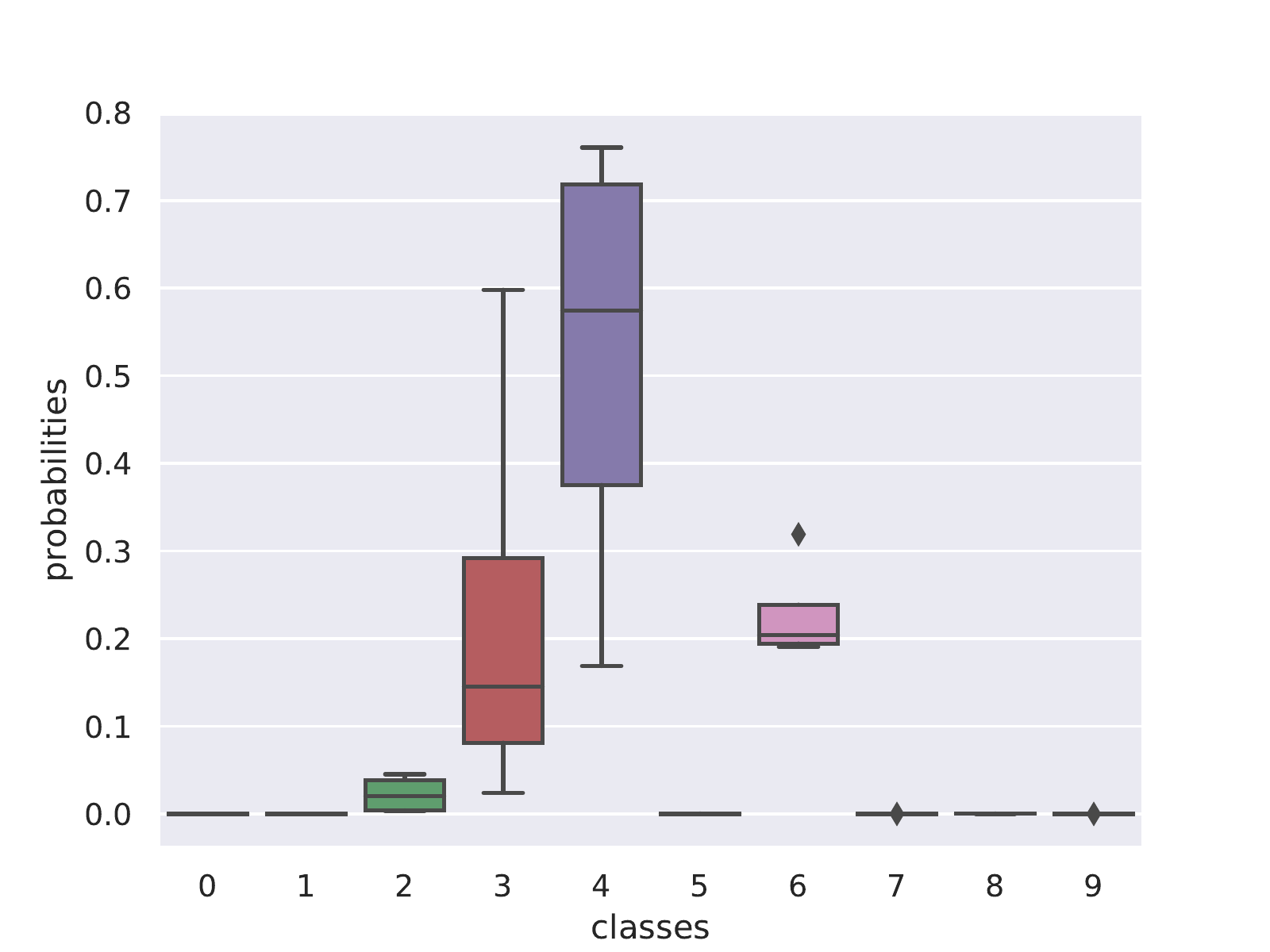}\label{dist4} &
 \includegraphics[width=0.15\linewidth]{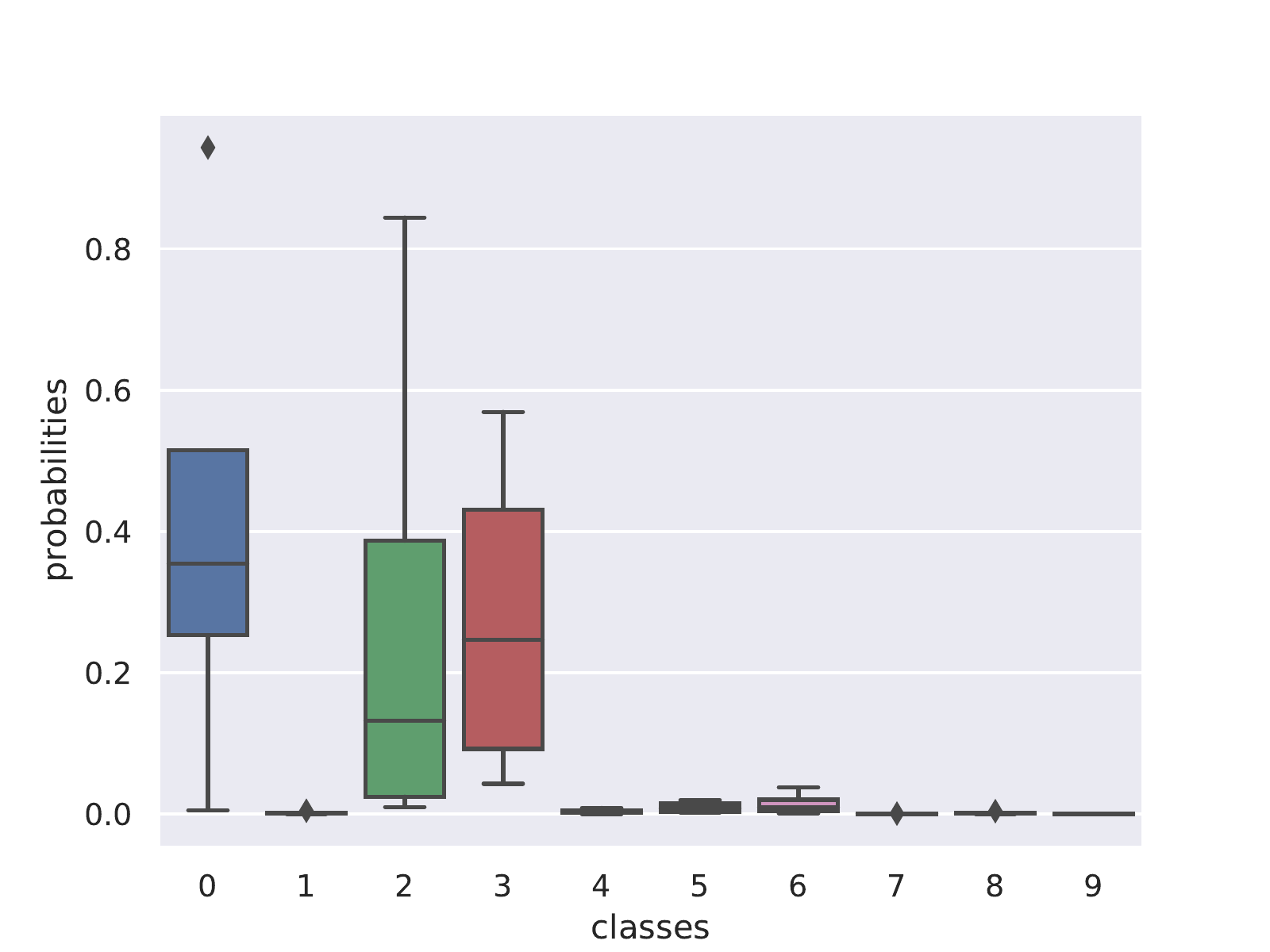}\label{dist4} &
 \includegraphics[width=0.15\linewidth]{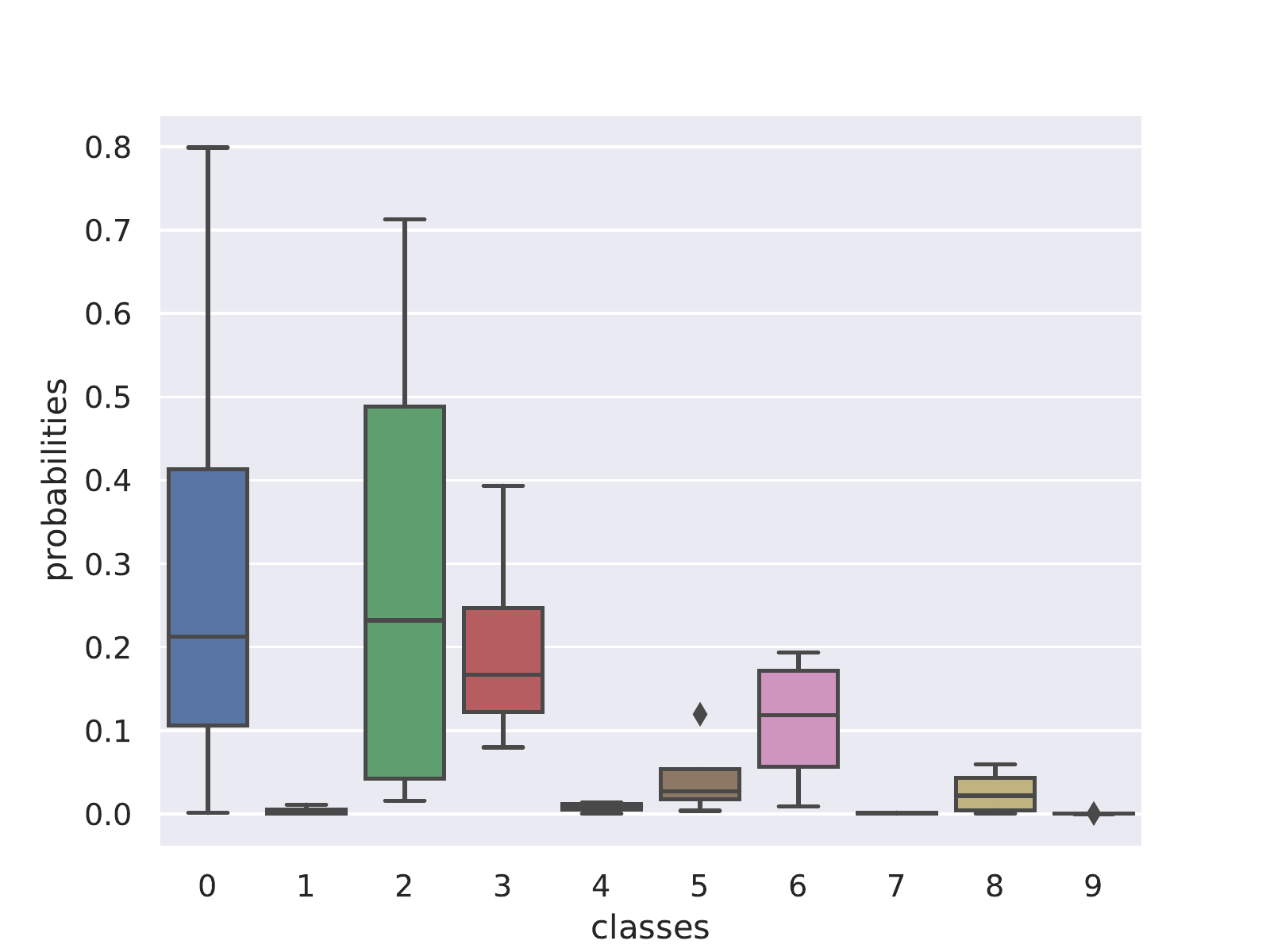}\label{dist4} &
 \\
\raisebox{7mm}{\large{BE}} &
 \includegraphics[width=0.15\linewidth]{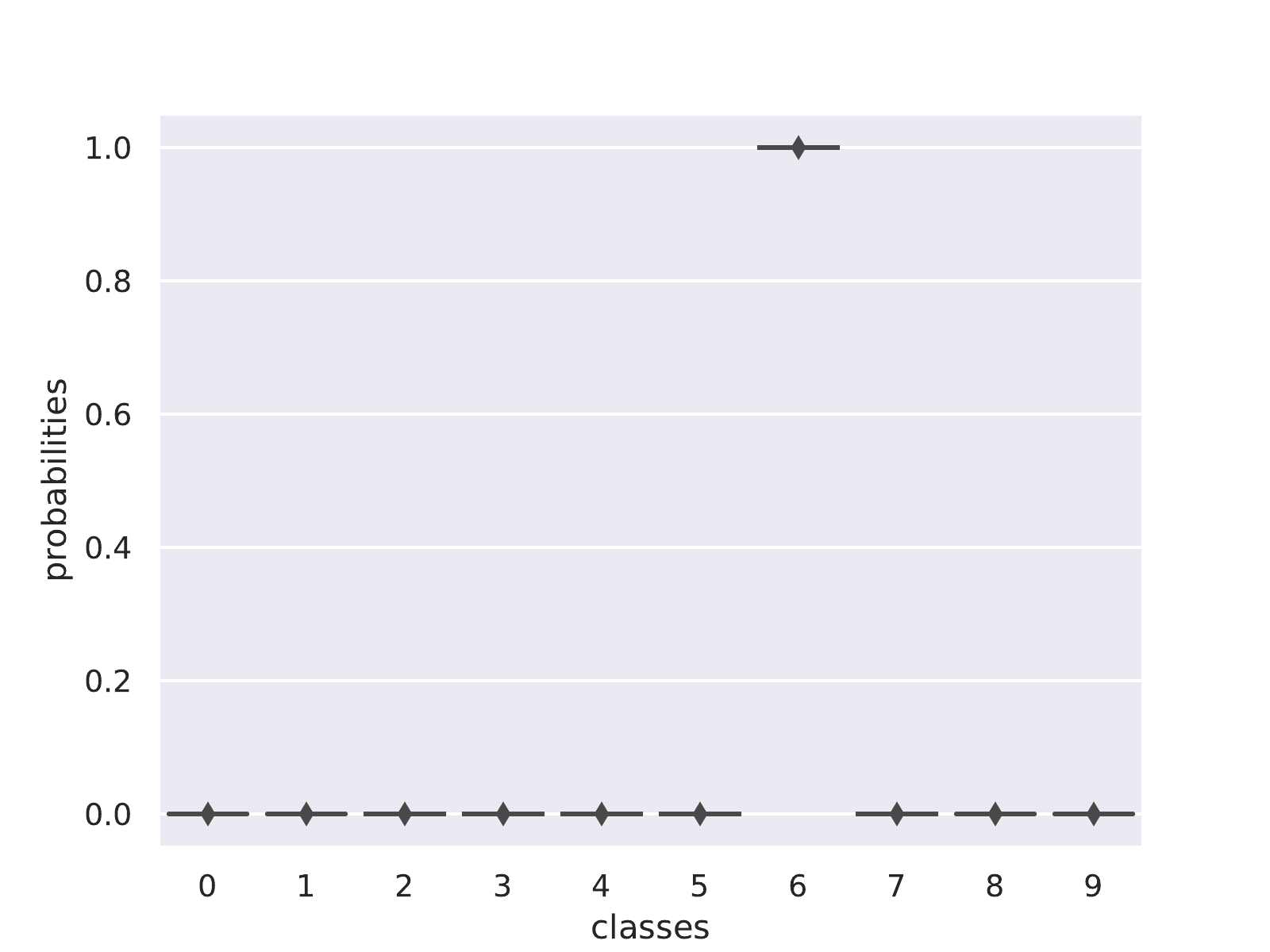}\label{dist1} &
 \includegraphics[width=0.15\linewidth]{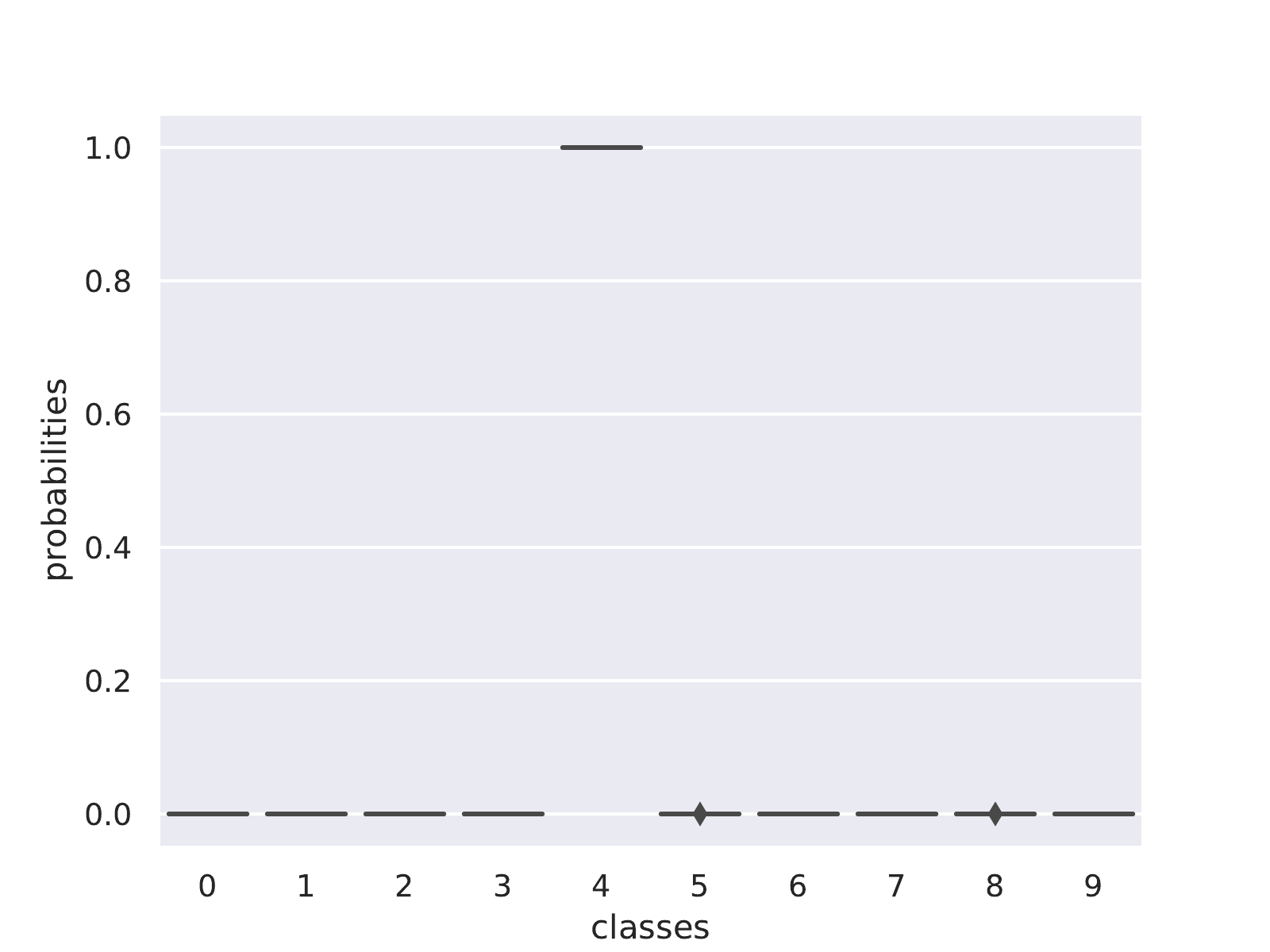}\label{dist2} &
 \includegraphics[width=0.15\linewidth]{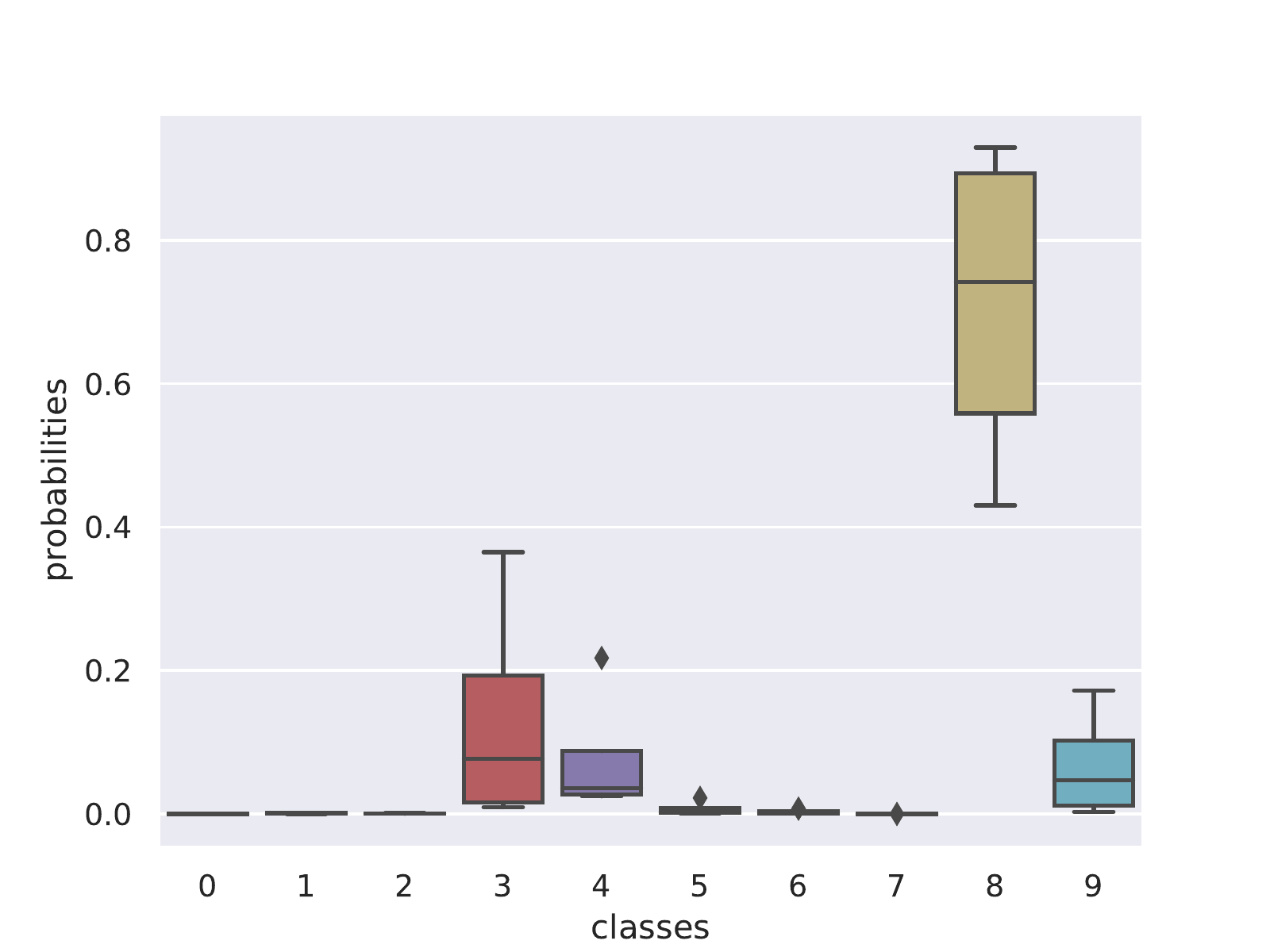}\label{dist4} &
 \includegraphics[width=0.15\linewidth]{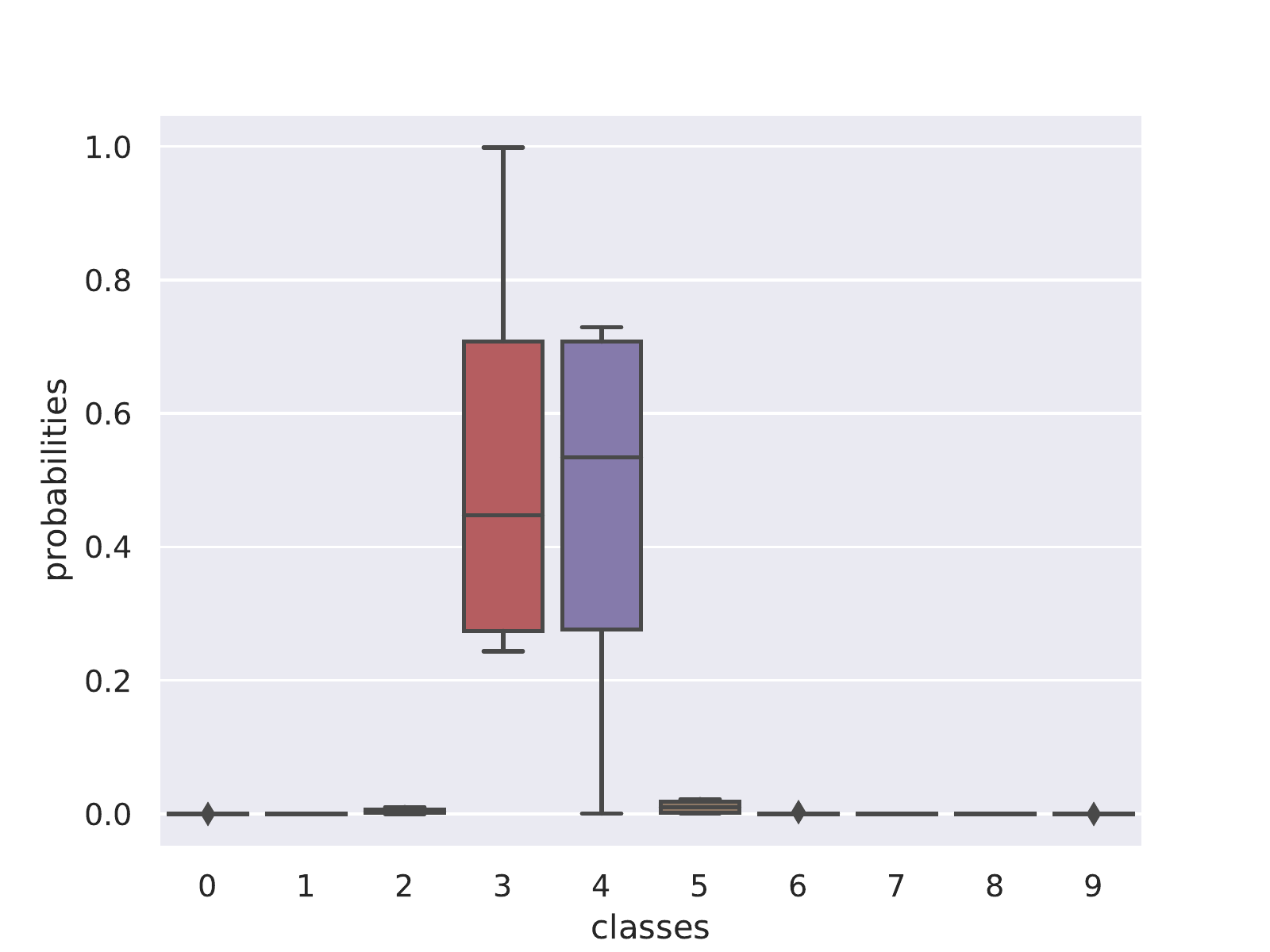}\label{dist4} &
 \includegraphics[width=0.15\linewidth]{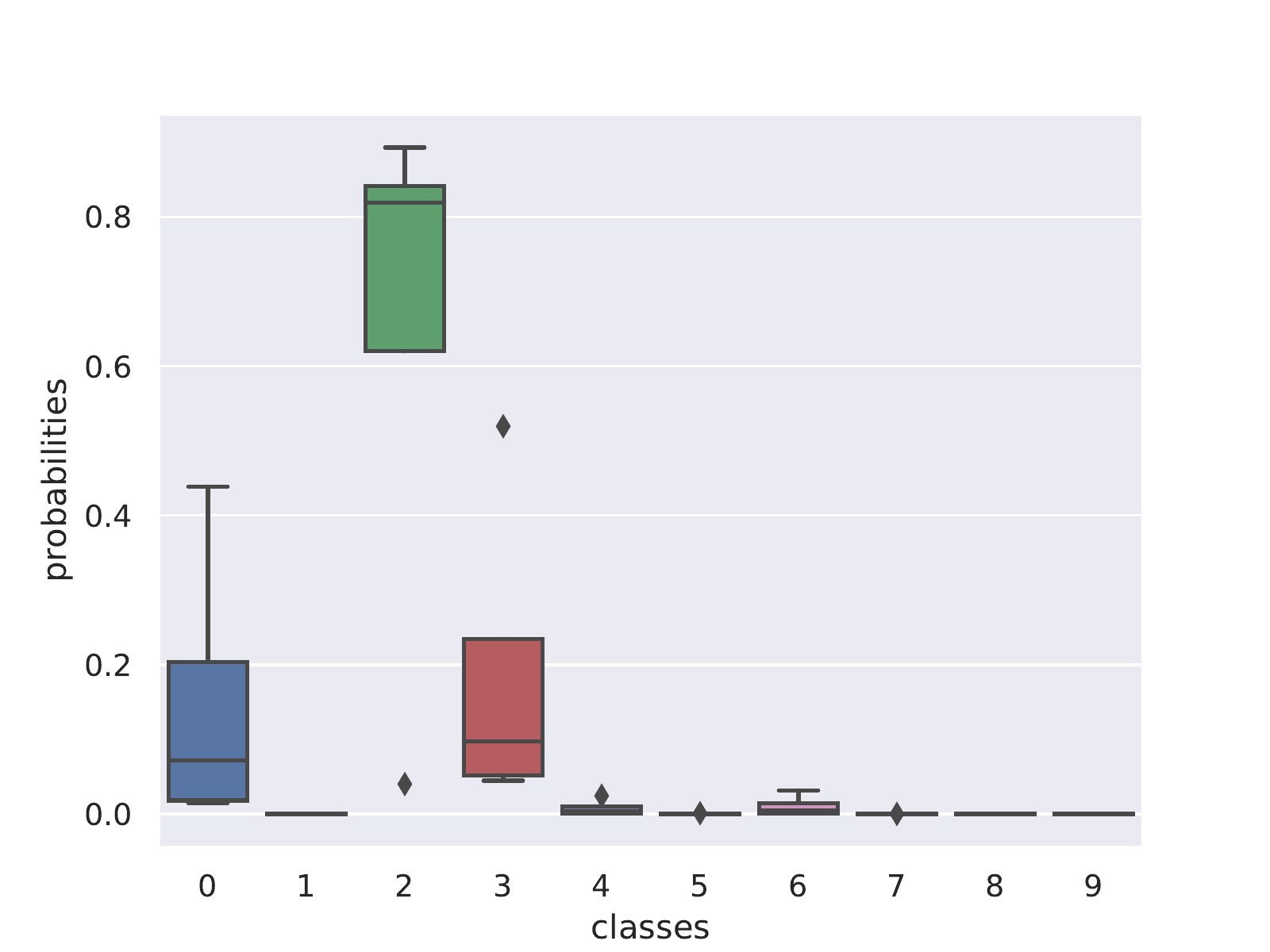}\label{dist4} &
 \includegraphics[width=0.15\linewidth]{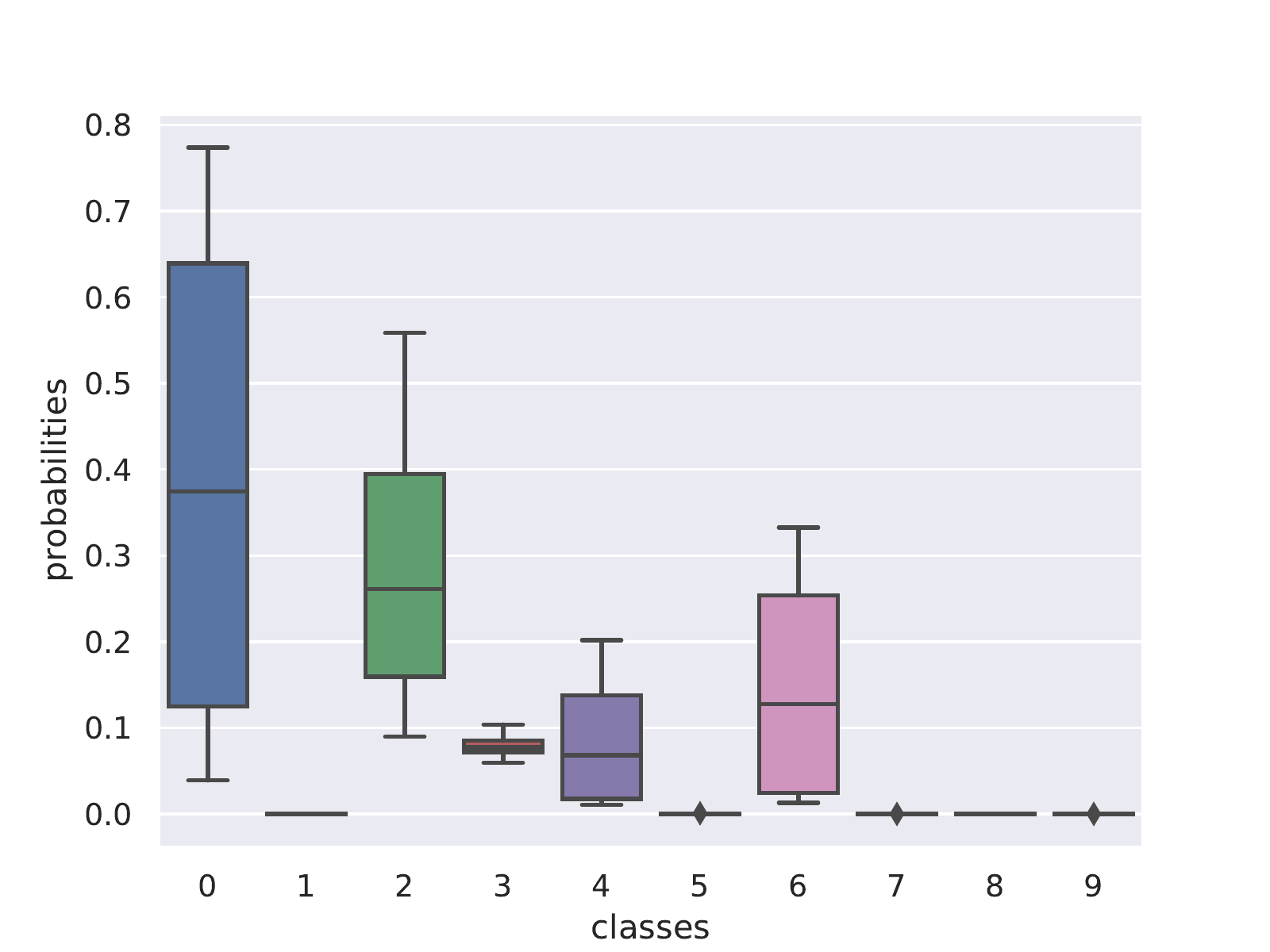}\label{dist4} &
 \\
\raisebox{7mm}{\large{DE}} & 
 \includegraphics[width=0.15\linewidth]{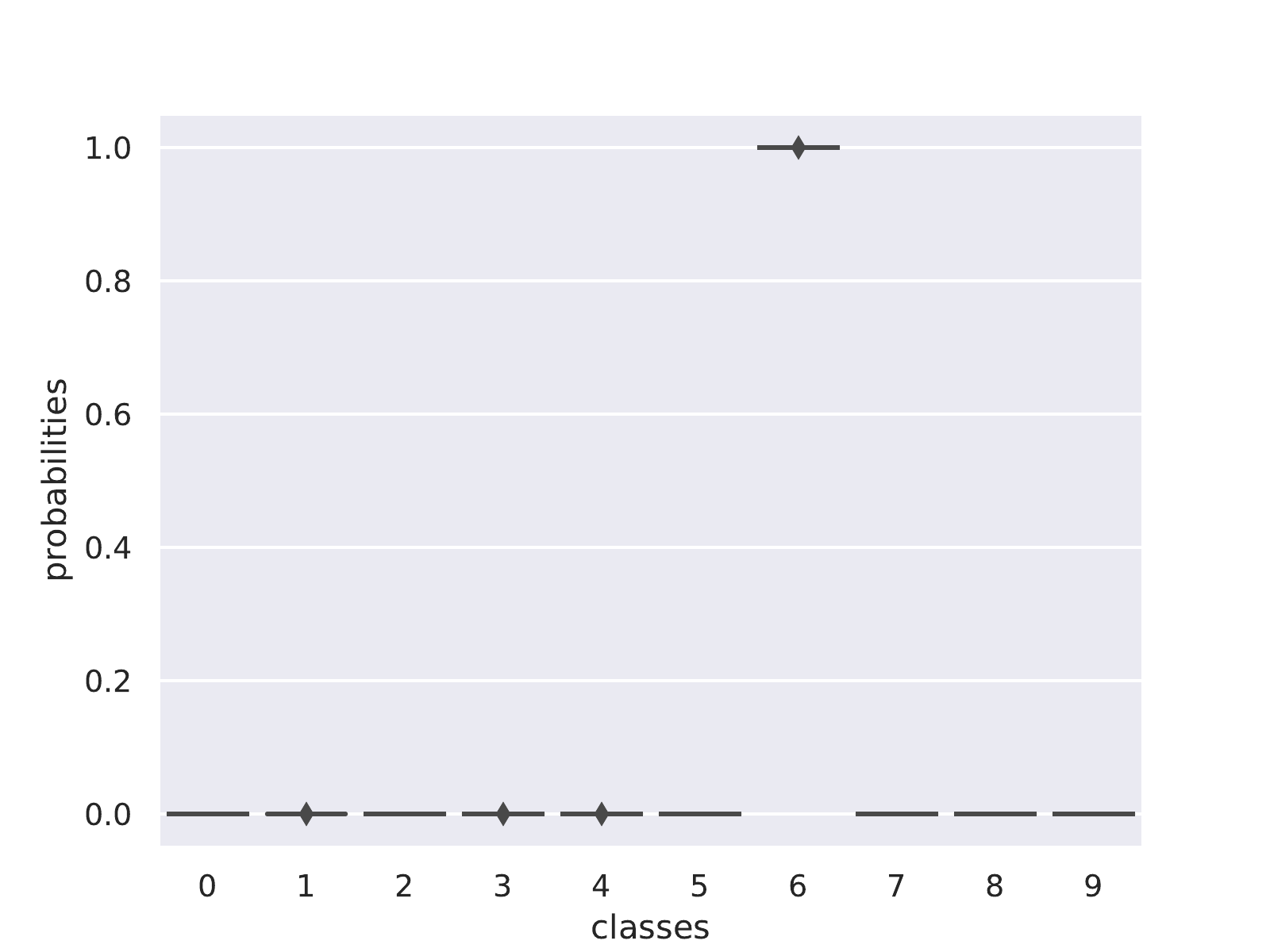}\label{dist1} &
 \includegraphics[width=0.15\linewidth]{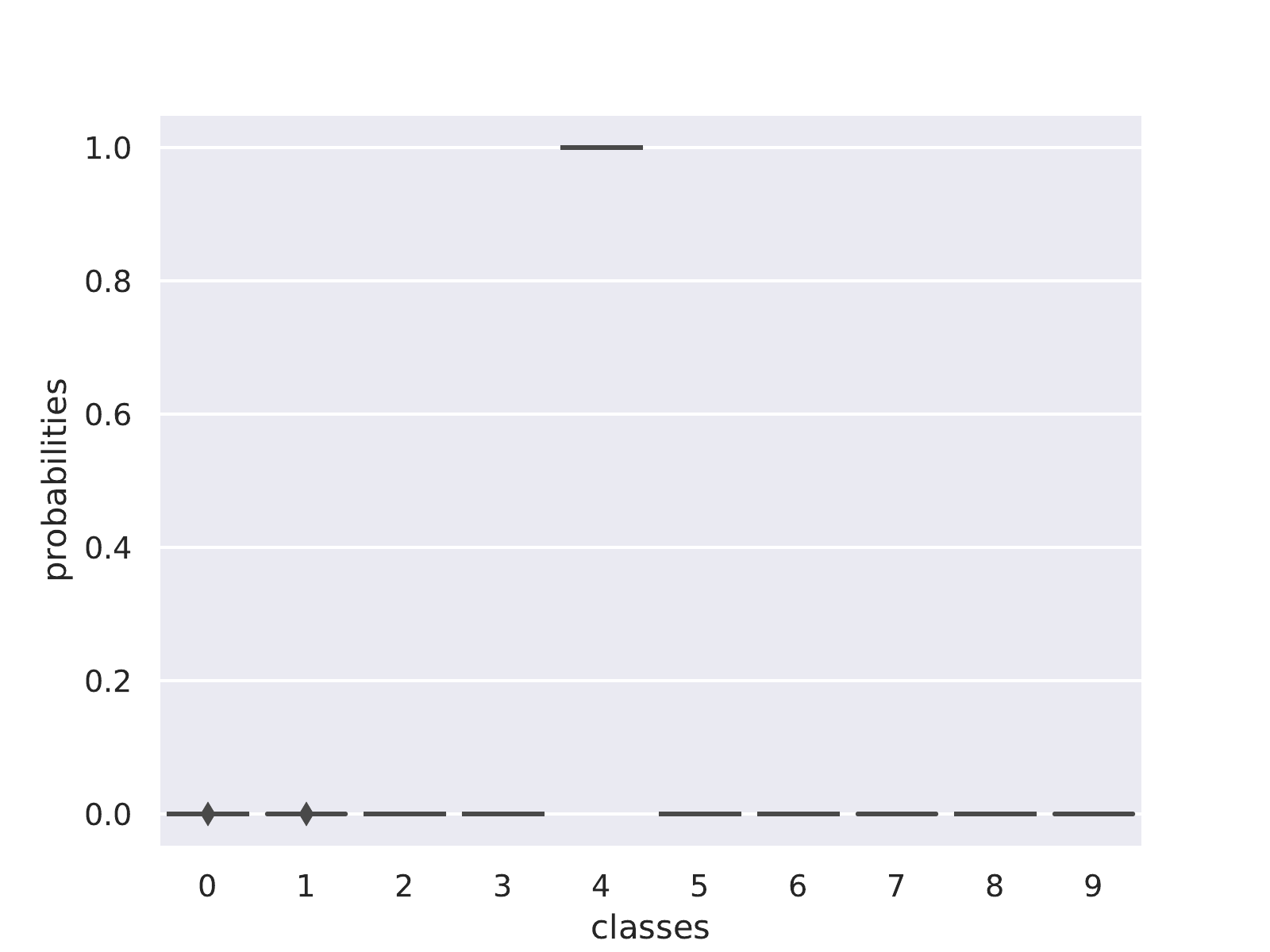}\label{dist2} &
 \includegraphics[width=0.15\linewidth]{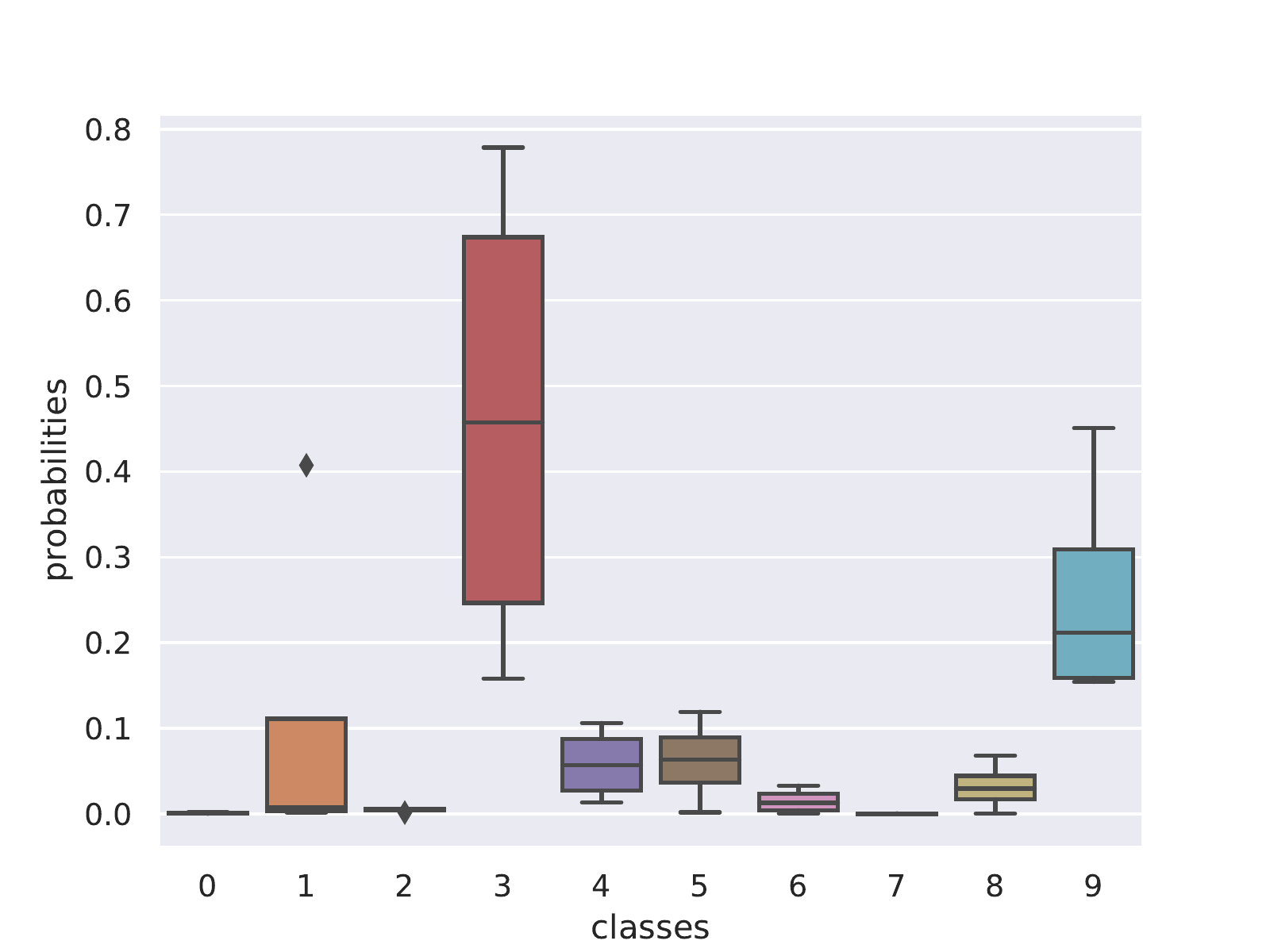}\label{dist4} &
 \includegraphics[width=0.15\linewidth]{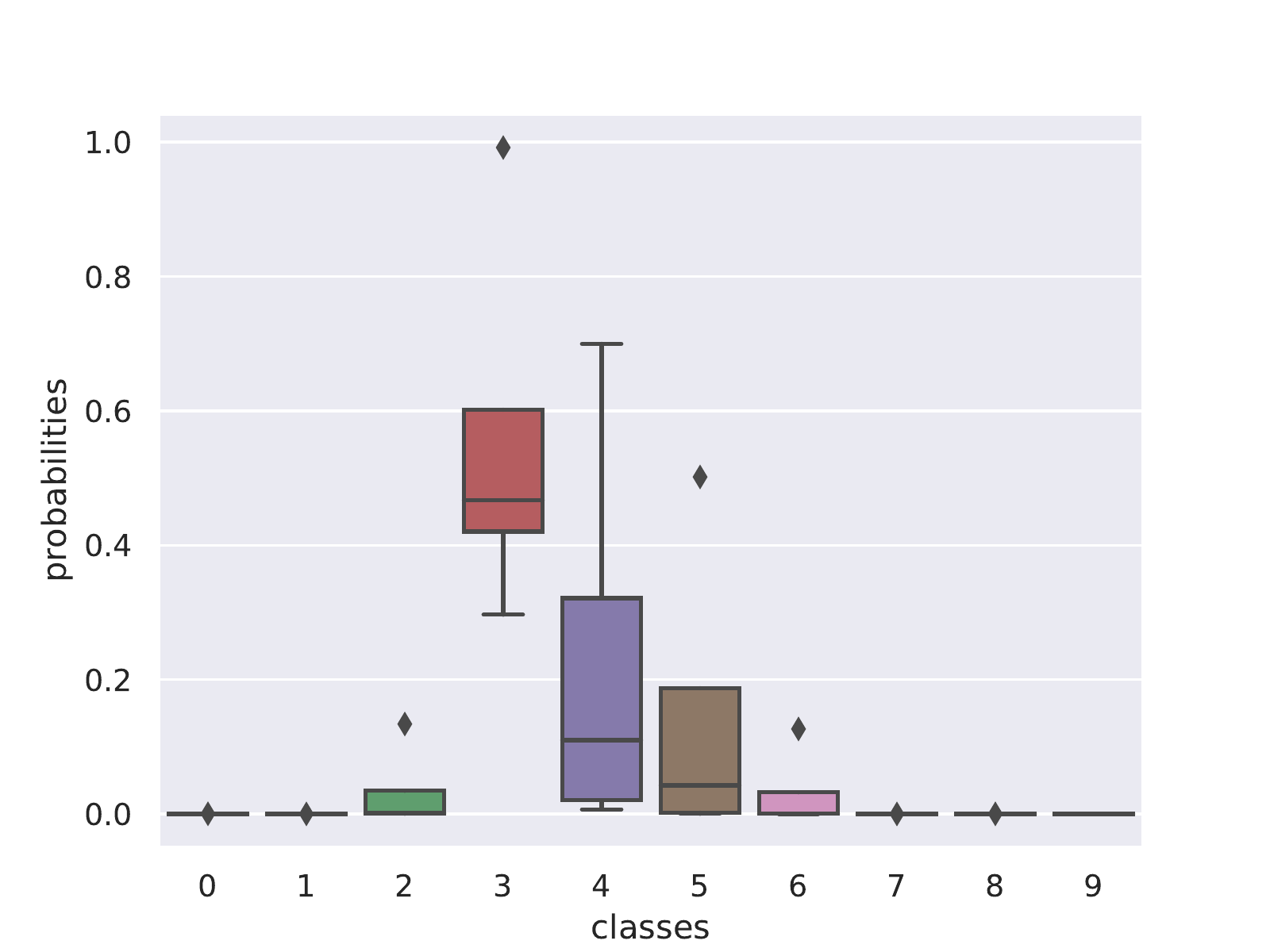}\label{dist4} &
 \includegraphics[width=0.15\linewidth]{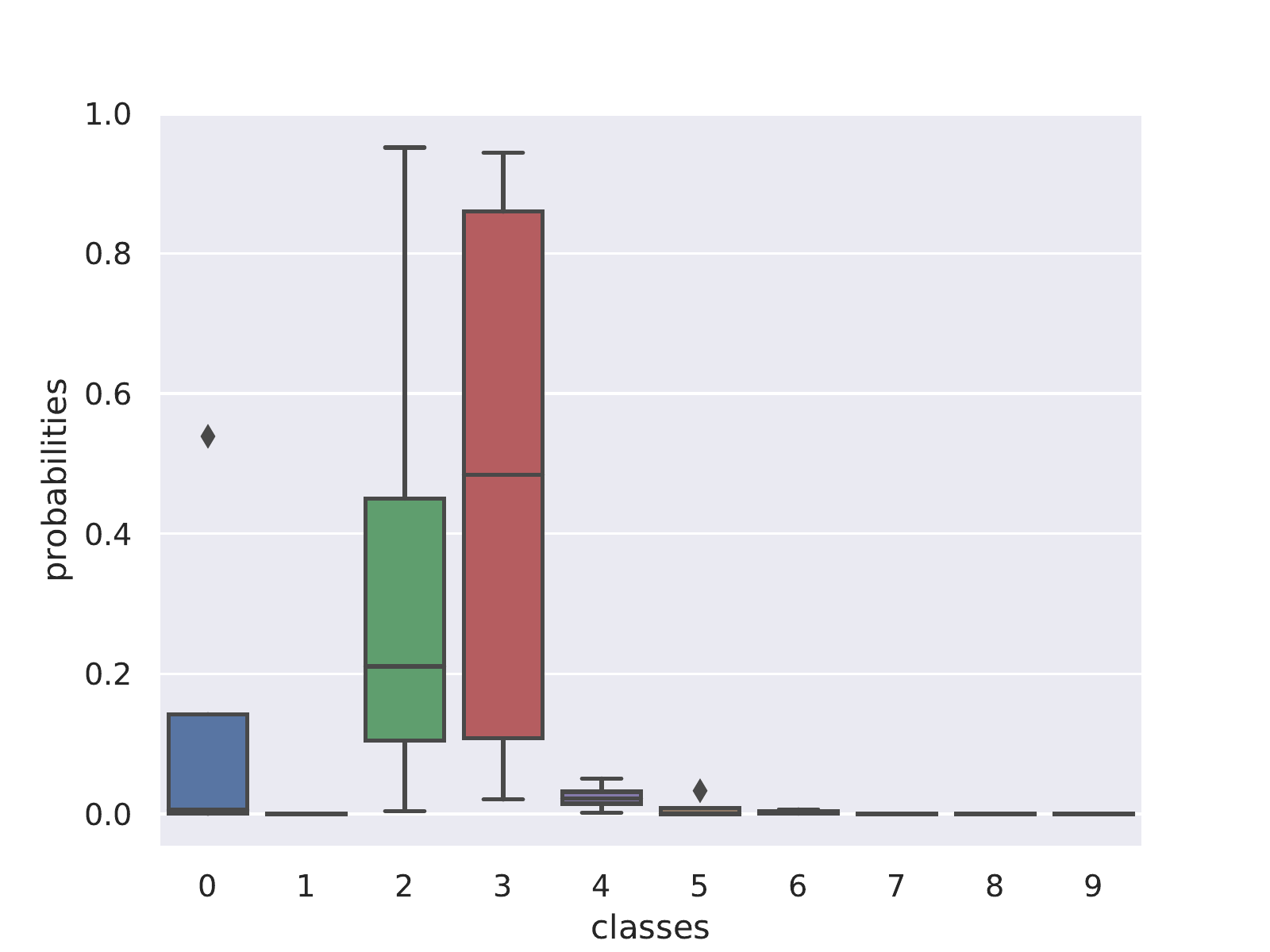}\label{dist4} &
 \includegraphics[width=0.15\linewidth]{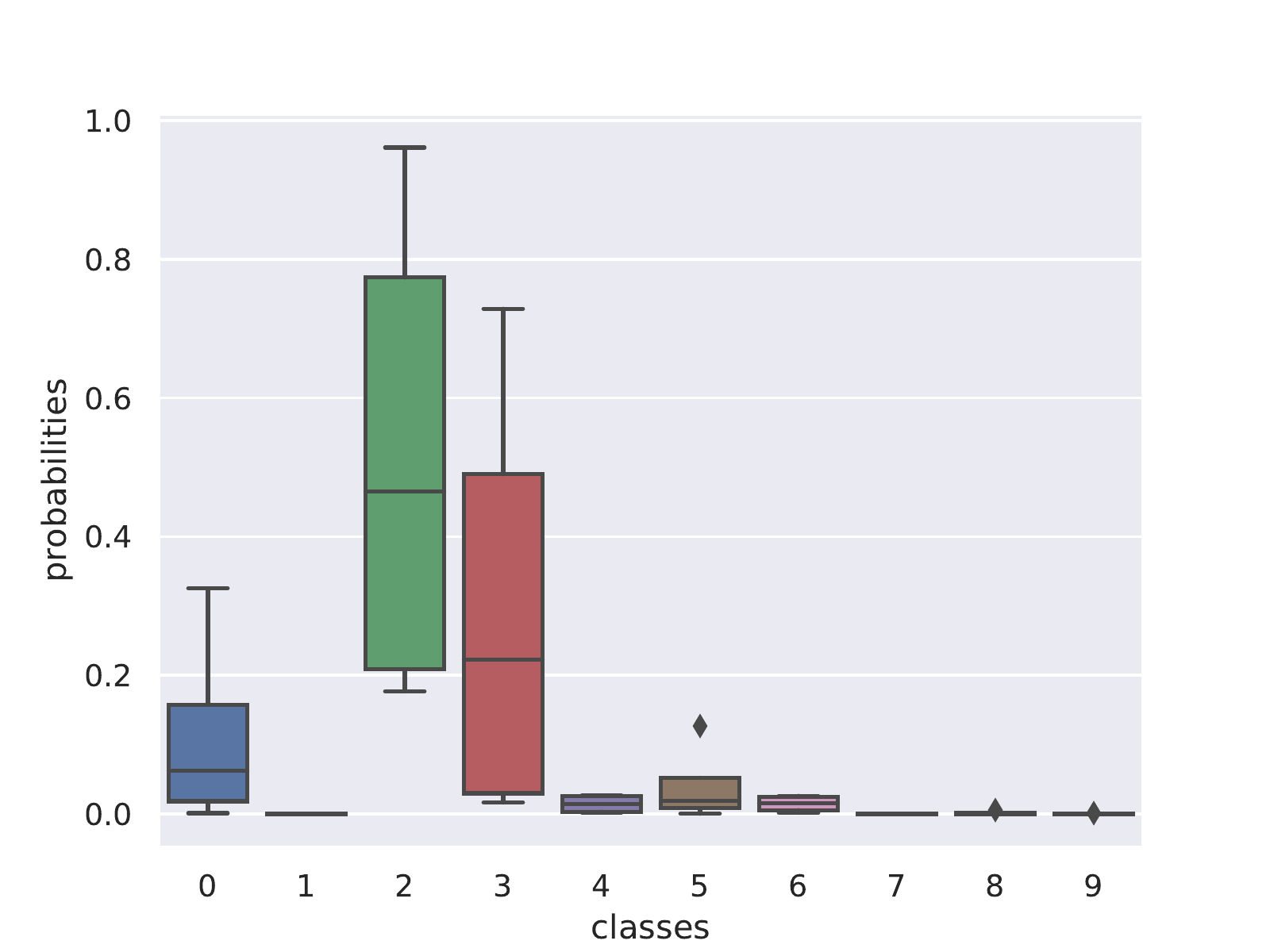}\label{dist4} &
 \end{tabular}
 }
\caption{\textbf{Diversity of predictions of different ensemble methods.}  The first row contains in order {two} images from the test set of CIFAR-10, of  CIFAR-10-C  and of SVHN, respectively. The next three rows represent the corresponding outputs of the different sub models for the three ensembling algorithms being considered: \method, BatchEnsemble and Deep Ensembles. }\label{fig:diversityall}
\end{figure*}

\section{Implementation details}

%We intend to release the codes on GitHub to share with the community our new layers that could be applied to any computer vision tasks. Our code is implemented in Pytorch \cite{NEURIPS2019_9015}. 
%\abc{We could mention here some info about the implementation, \eg pytorch, and the fact that we intend to release the code to facilitate reproducibility and progress in this area.}

\ab{For our implementation, we use PyTorch~\cite{paszke2019pytorch} and will release the code after the review in order to facilitate reproducibility and further progress. In the following we share the hyper-parameters for our experiments on image classification and semantic segmentation.}

% \subsection{StreetHazards~\cite{hendrycks2019anomalyseg} or BDD-Anomaly~\cite{hendrycks2019anomalyseg} experiments}
\subsection{Semantic segmentation experiments}
{In Table \ref{table:tab3}, we summarize the \ab{hyper-parameters} used in the StreetHazards~\cite{hendrycks2019anomalyseg} and BDD-Anomaly~\cite{hendrycks2019anomalyseg} experiments. }

\begin{table}[t!]
\renewcommand{\figurename}{Table}
\renewcommand{\captionfont}{\small}
\begin{center}
\scalebox{0.85}
{
\begin{tabular}{l|c|c}
\toprule
  \ab{\textbf{Hyper-parameter}} &   \textbf{StreetHazards} & \textbf{BDD-Anomaly}        \\ 
%   \ab{Hyper-parameter} &   StreetHazards & BDD-Anomaly        \\ 
\midrule
Ensemble size $J$          &4 & 4 \\ 
\midrule
learning rate         &0.1 &0.1 \\ 
 \midrule
batch size        &4 & 4 \\ 
 \midrule
number of train epochs  & 25 & 25 \\ 
\midrule
 weight decay for    $\theta^{\text{slow}} $  weights  &0.0001 & 0.0001 \\ 
 \midrule
  weight decay for $\theta^{\text{fast}} $    weights   &0.0 & 0.0 \\ 
 \midrule
 cutout         & True& True \\ 
 \midrule
 SyncEnsemble BN        & False & False \\ 
  \midrule
Group Normalisation      & True & True \\ 
  \midrule
 Size of the latent space $\vz$        & 32 & 32 \\ 
\bottomrule
\end{tabular}
}
\end{center}
\caption{
% Values of the hyper-parameters used in the semantic segmentation experiments (Section~5.3).
\ab{\textbf{Hyper-parameter configuration used in the semantic segmentation experiments (\S5.3).
}}
}\label{table:tab3}
\end{table}

\begin{table}[t!]
\renewcommand{\figurename}{Table}
\renewcommand{\captionlabelfont}{\bf}
\renewcommand{\captionfont}{\small}
\begin{center}
\scalebox{0.85}
{
\begin{tabular}{l|c|c}
\toprule
 \ab{\textbf{Hyper-parameter}} &   \textbf{CIFAR-10}  & \textbf{CIFAR-100} \\ 
%  \ab{Hyper-parameter} &   CIFAR-10  & CIFAR-100   \\  
\midrule
Ensemble size $J$          &4 & 4 \\ 
\midrule
 initial learning rate         &0.1 &0.1 \\ 
 \midrule
batch size        &128 & 128 \\ 
 \midrule
lr decay ratio     &0.1 & 0.1 \\ 
 \midrule
 lr decay epochs         &[80, 160, 200] & [80, 160, 200] \\ 
 \midrule
number of train epochs  &250 & 250 \\ 
\midrule
 weight decay for    $\theta^{\text{slow}} $  weights  &0.0001 & 0.0001 \\ 
 \midrule
  weight decay for $\theta^{\text{fast}} $    weights   &0.0 & 0.0 \\ 
 \midrule
 cutout         & True& True \\ 
 \midrule
 SyncEnsemble BN        & False & False \\ 
  \midrule
 Size of the latent space $\vz$        & 32 & 32 \\ 
\bottomrule
\end{tabular}
}
\end{center}
\caption{
% Values of the hyper-parameters used in the classification experiments (Section~5.2). 
\ab{\textbf{Hyper-parameter configuration used in the classification experiments (\S5.2).}}
} 
\label{table:tab2}
\vspace{-3pt}
\end{table}

% \subsection{CIFAR-10~\cite{krizhevsky2009learning} or CIFAR-100~\cite{krizhevsky2009learning} experiments}
\subsection{Image classification experiments}

In Table \ref{table:tab2}, we summarize the \ab{hyper-parameters} used in the CIFAR-10~\cite{krizhevsky2009learning} and CIFAR-100~\cite{krizhevsky2009learning} experiments.

% \vspace{10pt}

\section{Notations}

In Table \ref{table:tab1}, we summarize the {main} notations used in the paper. Table \ref{table:tab1} should facilitate the understanding of Section~2 (the preliminaries) and Section~3 (the presentation of our approach) of the main paper.

%\abc{Shouldn't we put the hyper-params for semantic segmentation experiments, too?} \Isa{yes, it would probably be more consistent to have the same type of presentation for all experiments}

\begin{table*}[!t]
\renewcommand{\figurename}{Table}
\renewcommand{\captionfont}{\small}
\centering
\scalebox{0.90}
{
\begin{tabular}{ll}
\toprule
\textbf{Notations}                      & \textbf{Meaning}  \\ 
\midrule
% $n$                            & number of training samples                                    \\ 
% \midrule
$\mathcal{D}=\{ (\vx_i,y_i )\}_{i=1}^{n}$ & the set of $n$ data samples and the corresponding labels \\ 
\midrule
% $f_{\Theta}(\cdot)$                            & DNN with parameters $\Theta$  \\ 
% \midrule
$\Theta$                      & the set of weights of a DNN            \\ \midrule
$P(\Theta)$& the prior distribution over the weights of a DNN    \\ 
\midrule
\Gianni{$Q_{\nu}(\Theta)$}& \Gianni{the variational prior distribution over the weights of a DNN used in standard BNNs}\ab{~\cite{blundell2015weight}}   
\\ 
\midrule
\Gianni{$\nu$}& \ab{the parameters of the variational prior distribution over the weights of a DNN used in standard BNNs~\cite{blundell2015weight}
}   \\ 
\midrule
$P(y_i \mid \vx_i,\Theta)$     & the likelihood that DNN outputs $y_i$ \emi{following a} prediction over input image $\vx_i$   \\ 
\midrule
$J$   & the number of ensembling DNNs  \\ 
\midrule
%$\theta^{\text{slow}} = \{ \vomega^{f} \}$   & the common/slow weights of the network   \\ \midrule
%$\theta^{\text{fast}} = \{ (\vr_j, \s_j) \}_{j=1}^{J}$   & the set of fast weights for ensembling of $J$ networks  \\ \midrule
\ab{$\theta^{\text{slow}}=\{ \mathrm{W}_{\scriptstyle \text{share}}\} $}   & \ab{the shared ``slow''} weights of the network   \\ 
\midrule
% \Gianni{$\theta^{\text{fast}} $ }  & \ab{the set of individual ``fast''} weights for ensembling of $J$ networks  \\ 
% \midrule
% $\theta^{\text{slow}} = \mathrm{W}_{\scriptstyle \text{share}}$   & the common/slow weights of the network in the case we just have one fully connected layer  \\ 
% \midrule
$\theta^{\text{fast}} = \{ \mathrm{W}_j \}_{j=1}^{J} = \{ (\vr_j, \s_j) \}_{j=1}^{J}$  & \ab{the set of individual ``fast''} weights of BatchEnsemble for ensembling of $J$ networks \\
%   &  in the case we just have one fully connected layer \\ 
\midrule
  $\theta^{\text{fast}} =  \{ (\hat{\vr}_j , \s_j) \}_{j=1}^{J}$   & \ab{the  set of fast weights of \method~ for ensembling of $J$ networks.} \\ &  \ab{$\hat{\vr}_j$ are sampled from the latent weight space of weights $\vr_j$.} \\
%   &  in the case we have \emi{only} one fully connected layer \\ 
\midrule
$\theta^{\text{variational}} = \{ (\phi_j, \vpsi_j) \}_{j=1}^{J}$   & the parameters of the VAE for \ab{computing the low dimensional latent distribution} of $\vr_j$  \\ \midrule
\ab{$\genc(\cdot)$} & the encoder of the VAE applied on $\vr$  %\Isa{then $\vr_j$ should appear in the notation}
\\ \midrule
\ab{$\gdec(\cdot)$} &  \ab{the decoder of the VAE for reconstructing $\hat{\vr}$ from latent code of $\vr$} \\ \midrule 
$Q_{\phi}(\vz \mid \vr)$ 
 & the variational distribution over the weights $\vr$ to approximate the intractable posterior $P_{\psi}(\vz \mid \vr)$   \\ \midrule
$ (\vmu_j, \vsigma_j) =\genc(\vr_j)$  & \ab{encoder output that parameterize a multivariate Gaussian with diagonal covariance} \\ \midrule
$\vz_j \sim Q_{\phi}(\vz \mid \vr) = \mathcal{N}(\vz; \vmu_j, \vsigma^{2}_{j} \mathbf{I})$  &  sampling a latent code $\vz$ from the latent distribution \\
\midrule
$ P_{\psi}(\vz_j)= \mathcal{N}(\vz;0, \mathbf{I})$ with $j \in [1,J]$ & the prior distribution on  $\vz_j$ \\ 
\midrule
$\hat{\vr}_j = \gdec(\vz_j)$ &  the reconstruction of $\vr_j$ from its latent distribution \ab{, \ie the variational fast weights}  %\emi{(I think the equality is strict here, the approx would be with $\vr_j$} 
\\ 
\midrule 
%\Gianni{$W_{j} = ({\vr}_j,  \s_j^t$) with $j \in [1,J]$} & the weight in common to one subset of the batch out of $J$ subsets in BatchEnsemble \\ &\Isa{expliquer quelque part la difference entre $\vr_j$ et $\s_j$ -- $t$ ?}\\ \midrule
%$( \hat{\vr}_j , \s_j^t)$ with $j \in [1,J]$ & the weight in common to one subset of the batch out of $J$ subsets of VAEBNN  \\ \midrule
%$\overline{\vomega}_{j} = \vomega^f \odot \vomega_j  $  & the weight computed from slow and fast weights, specific to each subnet $j$  \\ \midrule
\Gianni{$\overline{\mathrm{W}}_{j} = \mathrm{W}_{\scriptstyle \text{share}} \odot \ab{(\vr_j  \s_j^{\top} )}   $} & \ab{the weight of a BatchEnsemble network $j$ computed from slow and fast weights}  \\
%   &  subset $j$  in the case we have \emi{only} one fully connected layer \\ 
 & where $\odot$ is the Hadamard product and  \ab{$(\vr_j  \s_j^{\top} ) $} the inner product between these two vectors.\\
\midrule
\Gianni{$\overline{\mathrm{W}}_{j} = \mathrm{W}_{\scriptstyle \text{share}} \odot \ab{(\hat{\vr}_j  \s_j^{\top} )} $}&  \ab{the weight  of \method~ network network $j$ computed from slow and variational fast weights} \\
%   & subset $j$  in the case we have \emi{only} one fully connected layer %\\ & \Isa{$\hat{\vr}_j  \s_j^t$ ou   $(\hat{\vr}_j , \s_j^t ) $ ?}
\bottomrule
\end{tabular}
}
\caption{\textbf{Summary of the main notations of the paper.}}
\label{table:tab1}
\end{table*}

%\clearpage

% \addtolength{\textheight}{-3.2cm}

\end{document}